\documentclass[10pt,twocolumn,letterpaper]{article}
\usepackage{3dv}
\usepackage{times}
\usepackage{epsfig}
\usepackage{amsmath}
\usepackage{amssymb}

\usepackage{comment}
\usepackage{url}            %
\usepackage{booktabs}       %
\usepackage{amsfonts}       %
\usepackage{nicefrac}       %
\usepackage{microtype}      %
\usepackage{xcolor}         %
\usepackage{xspace}
\usepackage{tabularx}
\usepackage{graphicx}
\usepackage{color}
\usepackage{multirow}
\usepackage[numbers,sort,compress]{natbib}
\usepackage{adjustbox}

\usepackage[pagebackref=true,breaklinks=true,colorlinks,bookmarks=false]{hyperref}
\usepackage[capitalize,noabbrev]{cleveref}

\newcommand{\internet}{\textsc{3DADN}\xspace}
\newcommand{\lasr}{\textsc{LASR}\xspace}
\newcommand{\viser}{\textsc{ViSER}\xspace}
\newcommand{\ditto}{\textsc{Ditto}\xspace}
\newcommand{\cuboidopt}{\textsc{CubeOPT}\xspace}
\newcommand{\cuboidrand}{\textsc{CubeRand}\xspace}
\newcommand{\dhoi}{\textsc{D3DHOI}\xspace}

\newcommand{\accr}{\textsc{AccR}\xspace}
\newcommand{\accp}{\textsc{AccP}\xspace}
\newcommand{\accm}{\textsc{AccM}\xspace}
\newcommand{\accrp}{\textsc{AccRP}\xspace}
\newcommand{\accrpm}{\textsc{AccRPM}\xspace}

\newcommand{\R}{\ensuremath{\mathbb{R}}}

\newcommand{\unit}[1]{\ensuremath{\,\mathrm{#1}}}
\newcommand{\denselist}{\itemsep 0pt\parsep=0pt\partopsep 0pt\vspace{-\topsep}}
\newcommand{\mypara}[1]{\vspace{2pt}\noindent\textbf{#1}}

\newcommand\best[1]{\textbf{#1}}

\newcommand\gtobjmark{$^*$}
\newcommand\ourtitle{Articulated 3D Human-Object Interactions From RGB Videos:\\An Empirical Analysis of Approaches and Challenges}

\newcolumntype{Y}{>{\centering\arraybackslash}X}
\newcommand\imgclip[2]{\adjincludegraphics[Clip={#1\width} {#1\height} {#1\width} {#1\height}]{#2}}

\threedvfinalcopy %

\title{\ourtitle}
\author{Sanjay Haresh, Xiaohao Sun, Hanxiao Jiang, Angel X. Chang, Manolis Savva\\
Simon Fraser University\\\href{https://3dlg-hcvc.github.io/3dhoi/}{3dlg-hcvc.github.io/3dhoi}
}

\begin{document}
\maketitle

\begin{abstract}
Human-object interactions with articulated objects are common in everyday life.
Despite much progress in single-view 3D reconstruction, it is still challenging to infer an articulated 3D object model from an RGB video showing a person manipulating the object.
We canonicalize the task of articulated 3D human-object interaction reconstruction from RGB video, and carry out a systematic benchmark of five families of methods for this task: 3D plane estimation, 3D cuboid estimation, CAD model fitting, implicit field fitting, and free-form mesh fitting.
Our experiments show that all methods struggle to obtain high accuracy results even when provided ground truth information about the observed objects.
We identify key factors which make the task challenging and suggest directions for future work on this challenging 3D computer vision task.

\end{abstract}

\section{Introduction}

Indoor environments contain many articulated objects with which we interact on a daily basis.
Doors, kitchen cabinetry, fridges, and drawers are but a few examples.
Thus, a comprehensive 3D understanding of the world requires modeling how people interact with articulated objects.

\begin{figure}
\includegraphics[width=\linewidth]{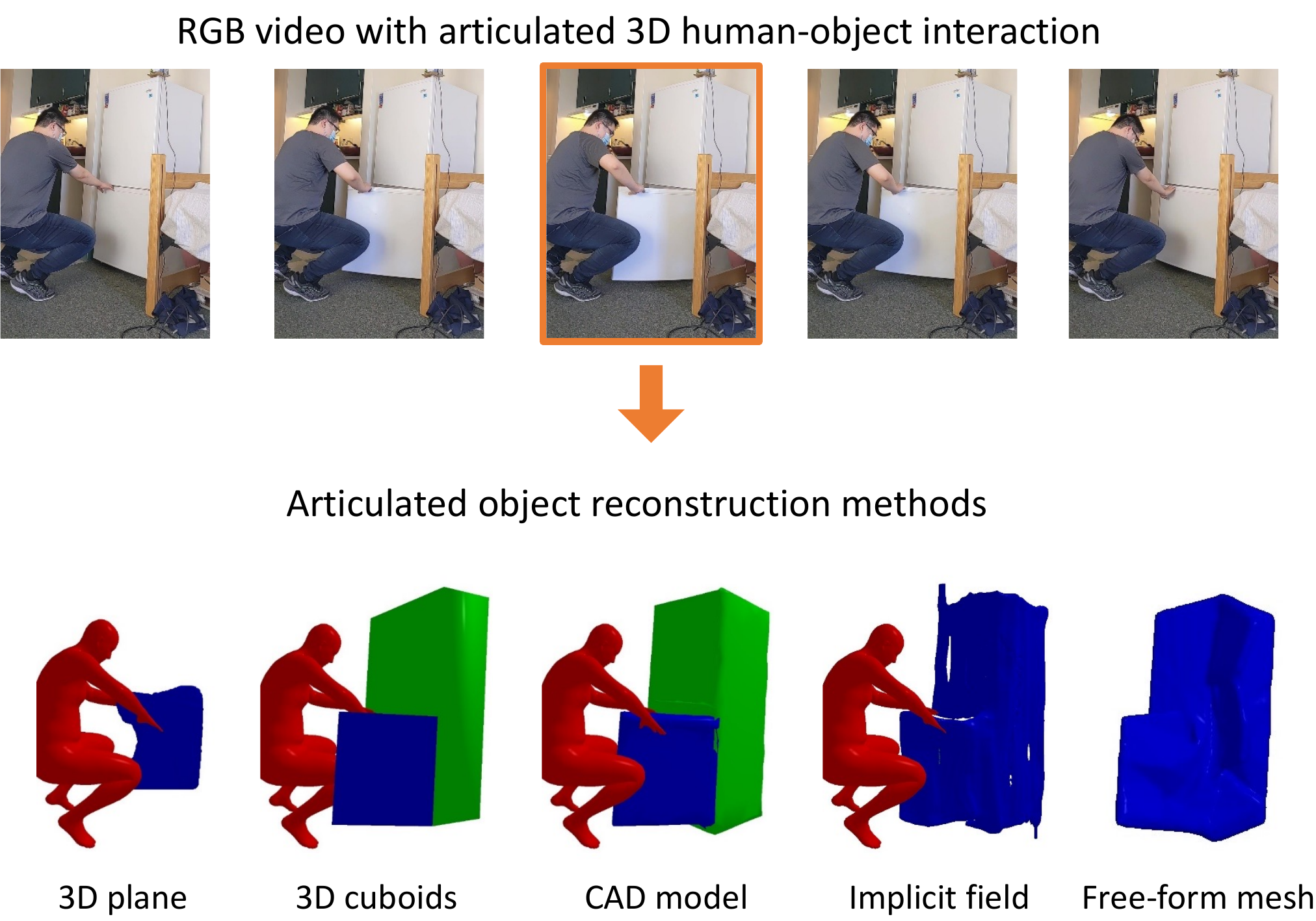}
\caption{
We address the articulated 3D human-object interaction task.
Given an input RGB video, methods tackling this task output a 3D representation of an articulated object manipulated by a human.
This is a challenging task requiring 3D object reconstruction under severe occlusion, as well as part motion estimation.
We systematically benchmark methods using a spectrum of representations for the articulated object ranging from simple planes and cuboids to CAD models, implicit fields, and free-form meshes.
}
\label{fig:overview}
\end{figure}

There has been much recent progress in dynamic 3D reconstruction targeting humans~\cite{loper2015smpl,bogo2016keep,pavlakos2019expressive,iqbal2021kama}, animals~\cite{zuffi20173d,zuffi2018lions,zuffi2019three}, and objects~\cite{yang2021lasr,yang2021viser,yang2021banmo,wu2021dove}.
There is also increasing interest in whole-body human-object interactions (HOI) in 3D~\cite{zhang2020perceiving,xie2022chore,bhatnagar2022behave}.
However, this line of work assumes static scenarios~\cite{zhang2020perceiving,xie2022chore}, or assumes the object is not articulated~\cite{bhatnagar2022behave}.

There is much less work on modeling 3D HOI with articulated household objects (e.g. `opening a microwave').
Such scenarios exhibit many challenges.
Firstly, both the person and articulating object are moving and changing in appearance.
Moreover, even though the object may articulate many household objects such as furniture and appliances, are themselves not easily movable.
This coupled with typically limited viewpoints for such interaction scenarios leads to significant occlusions between the object and the human, and in particular very partial observations of the object (e.g., the back of a fridge is rarely observed, see \Cref{fig:overview}).
Thus inference of 3D HOI with articulated objects from monocular RGB video is a challenging problem, especially as there is limited ground truth data for supervision.

While there are challenges in modeling articulated 3D HOI, there are also opportunities.
Many common objects with which people interact are composed of rigid parts, and can be abstracted to simple primitive shapes.
For instance, 3DADN~\cite{qian2022understanding} recently simplified the articulated 3D HOI problem to a single 3D plane estimation for a moving part such as a microwave oven door.
\citet{xu2021d3d} introduced a dataset of RGB videos of 3D HOI with articulated objects, and demonstrated that the human pose can be used to refine the articulated object reconstruction.
Additional physical constraints and priors connecting the static and moving parts of objects and the moving human body can likely enable even better articulated 3D HOI reconstruction. %

In this paper, we define and canonicalize the ``RGB video to articulated 3D HOI'' task.
Our goal is to analyze the performance of recent techniques for direct reconstruction of 3D articulated objects from RGB video.
To this end, we systematically benchmark five families of techniques: plane estimation, cuboid abstraction, CAD model fitting, implicit fields, and freeform mesh fitting.
We find that even with access to ground truth information all methods struggle to obtain high quality results especially when considering motion parameter estimation for articulated parts.
Direct optimization methods using cuboidal abstractions or CAD models perform relatively well compared to more complex approaches, but in the latter case are limited by needing a dataset of CAD models matching the input videos.

\section{Related Work}

\mypara{3D articulated object reconstruction.}
There has been growing interest in reconstructing 3D articulated objects i.e. objects with moving parts such as drawers and doors.
Earlier work focused on using probabilistic models for modeling articulated objects as kinematic graphs~\cite{sturm2009learning,sturm2011probabilistic}.
Other work predicted part poses~\cite{liu2020nothing} and articulation parameters~\cite{michel2015pose, li2020category,jain2021screwnet}, and reconstructed objects that can be articulated from partial 3D point clouds~\cite{mu2021sdf,jiang2022ditto}.
Recent work has employed radiance fields to model articulated objects~\cite{tseng2022cla}.
The above work typically operates on largely unoccluded views of a single object of interest.
Such views are quite dissimilar from real-world videos of human interaction with common household objects, where clearly the object does not typically articulate on its own.
Our task focuses on reconstructing a 3D articulated object from RGB videos of natural interactions involving both human and object.

\mypara{Monocular 3D articulated objects from RGB video.}
We focus on single-view RGB video input as it is ubiquitous and captures the inherent temporal nature of motion by the human causing the object articulation.
Despite this fact, there is little prior work taking advantage of the temporal nature of articulation.
Some existing work relies on sequences of depth frames~\cite{jain2021screwnet,jain2022distributional} or estimated depth frames~\cite{liu2020nothing}.
Methods that estimate 3D articulated objects from RGB video tend to assume predetermined articulation structure (e.g., human, four-legged animal)~\cite{kanazawa2019learning, biggs2018creatures}. 
More recent methods~\cite{yang2021lasr, yang2021viser,wu2021dove,yang2021banmo} have explored category-agnostic articulated object reconstruction.
There has also been work that discovers parts and joints from multiview video~\cite{noguchi2022watch}.
We focus on reconstructing a 3D model of an articulated object from a single fixed-view RGB video, during a natural interaction involving a human manipulating the object.

\mypara{3D human object interaction.}
While there is a long line of work on 2D human-object interaction (see recent survey by \citet{bergstrom2020human}), there is much less work on 3D human-object interaction.
Some work deals with hand-object interactions~\cite{hasson2019learning,cao2021reconstructing} typically focusing on estimating hand pose for manipulations of rigid objects.
There has been some recent work on hand interactions with deformable objects~\cite{tsoli2018joint} and with articulated objects~\cite{fan2022articulated}.
We focus on full-body interactions with common household furniture and appliances.
Considering full-body 3D HOI there has been work on exploiting human-object constraints to model object arrangements~\cite{jiang2015modeling,fisher2015activity}, populate humans in scenes~\cite{hassan2021populating} and to recover human object arrangements in 3D~\cite{savva2016pigraphs}.
\citet{zhang2020perceiving} presented an optimization-based method to recover static human and object arrangements in 3D from a single RGB image.
\citet{bhatnagar2022behave} infer 3D HOI from RGB videos.
More recently, \citet{xie2022chore} handle dynamic interactions but again with non-articulating objects.
None of this work on full-body 3D HOI work handles articulated objects.
In contrast, \citet{xu2021d3d} proposed to exploit human-object constraints to reconstruct 3D articulated objects from videos of 3D HOI.
Like \citet{zhang2020perceiving}, this work requires predefined category-level information in the form of CAD models for each object category.
\citet{qian2022understanding} estimate a 3D plane for the moving part of an articulating object in a 3D HOI video.
In this paper, we systematically benchmark representative methods from these recent approaches for articulated 3D HOI on a standardized task definition and dataset.

\section{Task Problem Statement}

Here, we define our task.
Given a single-view fixed-viewpoint RGB video with $N$ frames $X = \{x_1, x_2, ..., x_N\}$ showing a human interacting with an object, we reconstruct an articulated model of the object capturing the shape $S$, pose $P=\{R,T,\sigma\}$, and articulation parameters $A_t$ at each timestamp $t$.
The object shape $S$ includes both the static parts of the object and a moving part that articulates according to $A_t$.
The object pose is parameterized by a rotation $R$, translation $T$, and scale $\sigma$.
The motion parameters $A_t=\{c_t,d_t,\alpha_t\}$ are composed of the motion axis origin $c_t \in \R^3$, motion axis direction $d_t \in \R^3$ (we also evaluate a directionless motion axis $a_t$), and motion state $\alpha_t$.
For simplicity we assume that the object is stationary so there is a single motion axis (i.e. $c$, $d$, $a$ are constant across $t$).
In this work, we restrict our study to rotational motion, and define $\alpha_t$ to be the rotated angle from the closed state. 
For consistency, we define positive angles to be counterclockwise about the motion axis direction.
We measure task performance along three axes:
\begin{itemize}\denselist
    \item \textbf{Reconstruction}: the object shape $S$ should accurately represent the object in the video.
    \item \textbf{Pose}: the inferred 3D pose of the object $P = \{R,T,\sigma\}$ should match the input video.
    \item \textbf{Motion}: the moving parts of $S$ should have accurate articulation parameters $A_t$.
\end{itemize}

The task defined above is challenging because: i) no 3D supervision is assumed and therefore the problem is highly under-constrained; and ii) we do not assume any category-level information for the objects (i.e. the objects being reconstructed are not known a priori).
This significantly increases the space of possible reconstructions.
Note that even though the video input includes a human actor, and modeling the human may be beneficial to methods tackling this task, we do not evaluate human pose estimation or mesh reconstruction.

\section{Methods}

Our goal is to benchmark different approaches for reconstructing an articulated 3D object from an RGB video of a human interacting with the object.
We pick representative methods that span a spectrum of output 3D representations types: 3D planes, 3D cuboids, CAD models, implicit fields, and free-form meshes.
This spectrum reconstructs the object with 3D representations ranging from ``low-fidelity'' to ``high-fidelity''.

\subsection{3D Plane Estimation}

\citet{qian2022understanding} propose the \emph{3D Articulation Detection Network} (3DADN) to detect the articulating part and estimate a 3D plane for it given an RGB video of a human manipulating an object as input.
Articulating part detection is done with a Mask R-CNN~\cite{he2017mask} architecture that is trained on internet videos to regress a part bounding box and motion axis.
A part plane regression head and mask head are finetuned using ScanNet~\cite{dai2017scannet} scenes augmented with virtual humans.
Then, a temporal optimization refines per-frame predictions across the video to improve motion axis consistency and compute motion values (i.e. rotation angles).

We adapt 3DADN to our task by making the following modifications.
First, we approximate the moving part as a cuboid, which we obtain by extruding the estimated 3D plane along the plane normal by $5$\unit{cm}.  Since 3DADN does not estimate the static part of the object, we use the moving part as a proxy for the entire shape $S$. %
Second, the original 3DADN method assumes monotonically increasing or decreasing motion parameters in an input video (i.e. only opening or closing motion).
Thus, we split input videos into parts containing only one type of motion (using the ground truth motion annotation), apply 3DADN to each part and recombine the output into a single sequence.
Finally, since we use a pretrained 3DADN, we adjust input videos to match the field-of-view of its training dataset through central cropping and padding.
Since 3DADN provides per-frame estimates of the motion axis and origin, we select the median of the predicted axes to obtain a consistent axis for all the frames.

\subsection{3D Cuboid Estimation}

Our cuboid-based abstraction baseline \cuboidopt represents each articulated object with two cuboids, $S = \{C_b, C_m\}$, where $C_b$ represents the static part and $C_m$ represents the moving part of the object.
The two cuboids are constrained to have an axis of rotation $a_m$ which coincides with an edge of the base part $C_b$.
Under this constraint, we optimize for a rotation $R$ and translation $T$ for the entire object assembly.
Since the two parts can have different sizes, there are scale parameters $\sigma_b$ and $\sigma_m$ for the base part and the moving part respectively.
Finally, we also optimize for articulation state parameters $\alpha_{t}$ which give the degree of rotation of the moving part at each timestamp $t$.
\Cref{fig:arch} provides an overview of the approach.
See the supplement for more implementation details.

\begin{figure}
\includegraphics[width=\linewidth]{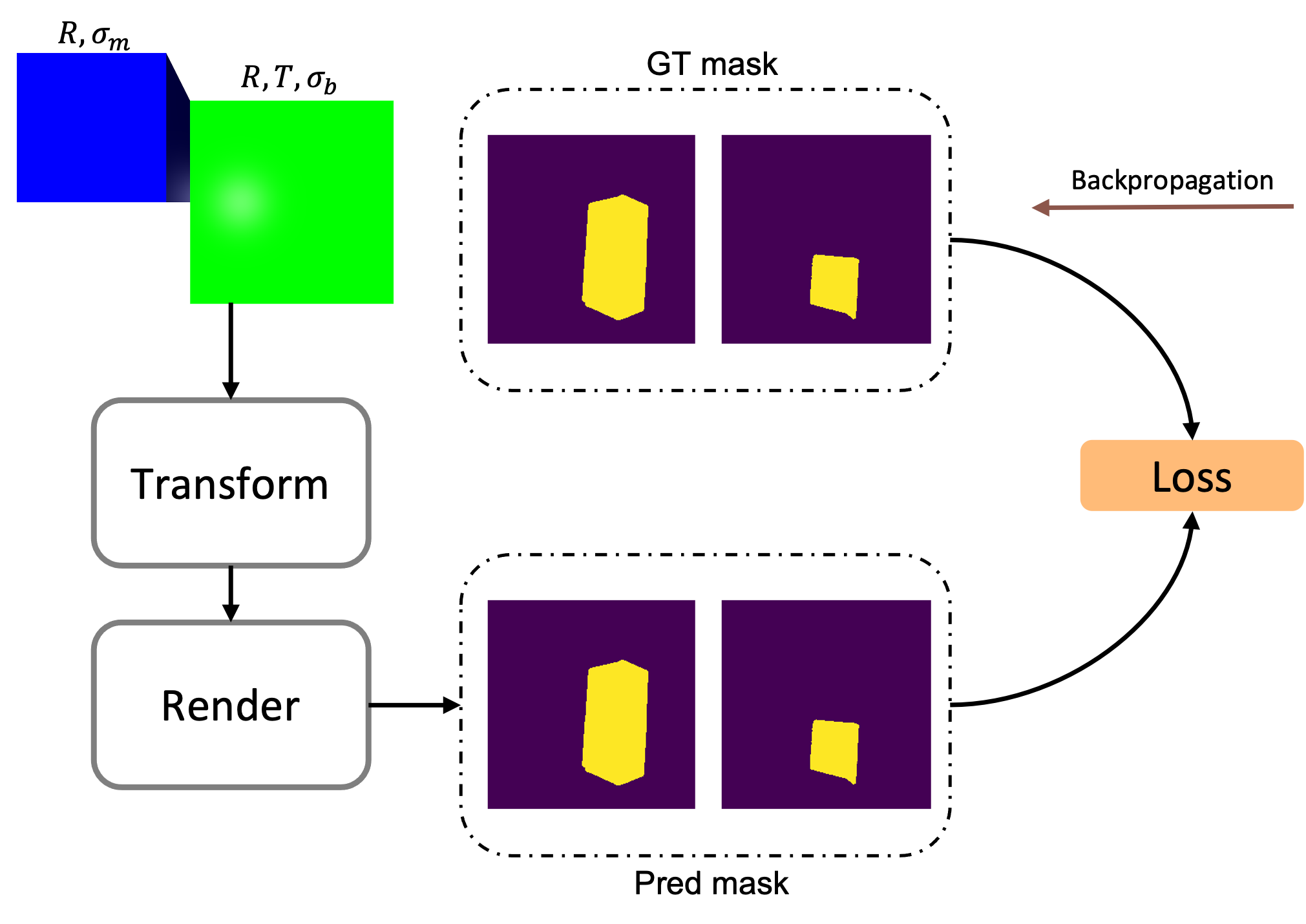}
\caption{\cuboidopt overview. We recover R, T, $\sigma_b$ and $\sigma_m$ estimates through gradient based optimization by minimizing the discrepancy between predicted object masks and ground truth object masks.
We use additional loss terms but do not show them here for simplicity.
}
\label{fig:arch}
\end{figure}

\subsection{CAD Model Fitting}

\citet{xu2021d3d} proposed an optimization based approach that reconstructs articulated objects and humans from RGB video.
The method adopts an optimization approach similar to \cuboidopt but approximates the object shape $S$ using a CAD model.
The pose $P$ and articulation parameters $A_t$ are optimized by minimizing the silhouette discrepancy with object segmentation masks and using human-object interaction terms to constrain the relative position of the human and object in 3D.

We evaluate three variations of the D3DHOI method.
The original method uses ground truth CAD models to fit the data which leads to very good results for object shape and articulation reconstruction.
However, this is not realistic as we typically cannot assume we have ground truth CAD models for an object observed in the input video at test time.
Therefore, we also evaluate two more variations of the method with less ground truth knowledge (and add the suffix ``GT-CAD'' to the original method name to communicate this privileged information is provided as input):
\begin{itemize}\denselist
    \item \textbf{\dhoi-Rand-Cat-CAD: } Instead of optimizing with the ground truth CAD model we randomly select a model from the same category.
    Note that while this variant assumes less information than the ground truth CAD model, it is still a strong baseline as it is given the category label of the object in the input video.
    \item \textbf{\dhoi-Rand-CAD: } We randomly select a CAD model from all available CAD models.
    This variant is more general as it does not rely on a priori knowledge of the object category.
    However it is still restricted to the 24 CAD models in the dataset from \citet{xu2021d3d}.
\end{itemize}

\subsection{Implicit Fields for Articulated Objects}

Implicit field have recently been used for modeling articulated objects~\cite{mu2021sdf,jiang2022ditto}.
We benchmark \ditto~\cite{jiang2022ditto}, a category-agnostic method for reconstructing articulated objects using implicit fields.
\ditto takes as input a pair of point clouds (PCs) of the object before and after interaction (in two different articulation states).
The method assumes that the two PCs are in the same pose.
To adapt \ditto to our task we project the GT CAD model into each frame using GT pose and camera parameters.
We then select one frame that is in the closed state (based on GT states) as the reference PC and iterate through the other frames using the PC obtained from the other frames treat other frames as the second PC.
This effectively gives us partial (single-view) GT PCs for each frame.
We use \ditto pretrained on the shape2motion~\cite{wang2019shape2motion} dataset and apply it to our dataset.
For each frame we obtain a predicted mesh reconstruction, segmentation, and joint parameters.
To obtain a consistent estimate of the joint axis and origin we take the median of the frame-wise predictions, similarly to 3DADN and LASR).
We then compute the mean errors over all the frames. 
There are some additional differences between how we use \ditto and how it was used in the original paper, which we discuss in detail in the supplement.

\subsection{Free-Form Mesh Fitting}

As representatives of free-form 3D mesh fitting we use the LASR~\cite{yang2021lasr} and \viser~\cite{yang2021viser} methods.
LASR is based on the classical ``analysis by synthesis'' approach i.e. instead of learning-based methods that learn a data-driven model using a large dataset, LASR reconstructs objects by overfitting to a single video.
It takes a mesh sphere and ``morphs'' it into the object shape $S$ by optimizing for silhouette, optical flow and texture losses.
This approach can work well with video data because videos can provide strong 3D multi-view constraints for a single object (relative to single-view images).
Using videos also enables the use of optical flow which provides dense correspondence of features between two images.

LASR also learns the camera parameters jointly with the object reconstruction.
We use the predicted camera parameters to compute the object pose parameters.
However, we note that the predicted mesh often does not articulate and the motion in the video is mostly explained by rigid transformations of a largely static object mesh.
Moreover, the predicted linear blend skinning weights do not carry semantic meaning making it difficult to create discrete ``moving'' and ``static'' parts, and compute the corresponding articulation motion parameters (e.g., axis of rotation).
\viser builds on LASR by learning surface embeddings to establish long range correspondences across video frames.
Long-range video pixel correspondences are forced to be consistent with an underlying canonical 3D mesh through embeddings that capture the appearance of each surface point.

While these methods provide good results on reconstruction of moving articulated objects such as animals and people from monocular videos, they struggle to reconstruct unobserved parts of the object.
We will see that these approaches fail to perform in our setting of indoor human-object interactions where many of the objects are partially observed.

\begin{table*}[t]
\caption{Error metrics for \internet, \cuboidrand, \cuboidopt (all losses), \dhoi and \lasr.  
\gtobjmark indicates that the method has access to the GT object mask (all methods except \internet and \cuboidrand).  In addition, \cuboidopt has access to GT part mask, \dhoi requires CAD models and \ditto has access to ground-truth depth maps.
We see that all methods exhibit fairly high motion parameter errors.
\lasr exhibits high pose translation and rotation errors.
}
\resizebox{\linewidth}{!}{
\begin{tabular}{@{} l rr rrr rr rr @{}}
\toprule
& \multicolumn{2}{c}{Reconstruction Error $\downarrow$} & \multicolumn{3}{c}{Pose Error $\downarrow$}
& \multicolumn{4}{c}{Motion Error $\downarrow$}\\
\cmidrule(l{0pt}r{2pt}){2-3} \cmidrule(l{0pt}r{2pt}){4-6}  \cmidrule(l{0pt}r{2pt}){7-10}  
Method & CD (Object) & CD (Moving) & Rotation & Translation & Scale &  Origin & Axis &  Direction & State  \\
\midrule
\internet~\cite{qian2022understanding} & $4.51 \pm 0.12$ & $0.44 \pm 0.04$ & $47.54 \pm 1.52$ & $2.10 \pm 0.06$ &$0.50 \pm 0.01$ & $0.99 \pm 0.10$ & $26.64 \pm 1.88$ & $111.88 \pm 4.48$ &$89.77 \pm 21.22$ \\
\midrule
\cuboidrand & $ 1.34 \pm 0.04 $ & $ 2.07 \pm 0.01 $ & $ 52.45 \pm 1.36 $ & $ 3.04 \pm 0.09 $ & $ 0.14 \pm 0.01 $ & $ 1.48 \pm 0.00 $ & $ 60.26 \pm 0.00 $ & $ 98.28 \pm 0.00 $ & $ 196.20 \pm 39.82 $ \\ 

\cuboidopt\gtobjmark & $1.57 \pm 0.11$ & $0.66 \pm 0.06$ & $38.94 \pm 1.74$ & $2.37 \pm 0.09$ &$0.28 \pm 0.01$ & $0.71 \pm 0.03$ & $27.42 \pm 2.06$ & $94.09 \pm 4.53$ &$126.60 \pm 5.00$ \\

\midrule
\dhoi-GT-CAD~\cite{xu2021d3d}\gtobjmark & $\textbf{0.45} \pm 0.05$ & $\textbf{0.15} \pm 0.02$ & $17.48 \pm 1.04$ & $\textbf{0.85} \pm 0.04$ &$\textbf{0.15} \pm 0.01$ & $\textbf{0.28} \pm 0.02$ & $\textbf{10.27} \pm 0.89$ & $\textbf{10.27} \pm 0.89$ &$\textbf{16.29} \pm 0.77$ \\
\dhoi-Rand-Cat-CAD\gtobjmark  & $0.66 \pm 0.06$ & $0.35 \pm 0.03$ & $24.84 \pm 1.53$ & $1.16 \pm 0.05$ &$0.18 \pm 0.01$ & $0.38 \pm 0.02$ & $19.27 \pm 1.78$ & $28.91 \pm 2.91$ &$24.28 \pm 1.07$ \\
\dhoi-Rand-CAD\gtobjmark & $1.90 \pm 0.20$ & $0.61 \pm 0.04$ & $35.26 \pm 1.66$ & $1.78 \pm 0.07$ &$0.22 \pm 0.01$ & $0.50 \pm 0.02$ & $54.30 \pm 2.43$ & $85.09 \pm 3.34$ &$29.71 \pm 0.96$ \\
\midrule
\ditto~\cite{jiang2022ditto} & $0.65 \pm 0.03$ & $1.63 \pm 0.06$ & $\textbf{9.75} \pm 0.37$ & $0.97 \pm 0.02$ &$0.06 \pm 0.00$ & $0.29 \pm 0.00$ & $67.57 \pm 1.61$ & $87.21 \pm 2.16$ &$70.10 \pm 1.59$ \\

\lasr~\cite{yang2021lasr}\gtobjmark & $1.19 \pm 0.04$ & $1.58 \pm 0.05$ & $37.60 \pm 1.13$ & $8.42 \pm 0.29$ &$0.43 \pm 0.02$ & $10.56 \pm 0.27$ & $60.07 \pm 1.63$ & $92.79 \pm 2.53$ &$161.27 \pm 4.93$ \\

\viser~\cite{yang2021viser}\gtobjmark & $1.48 \pm 0.04$ & $1.71 \pm 0.05$ & $49.66 \pm 1.51$ & $20.46 \pm 0.21$ &$0.61 \pm 0.01$ & $20.60 \pm 0.37$ & $56.00 \pm 1.43$ & $93.53 \pm 2.61$ &$188.55 \pm 5.22$ \\

\bottomrule
\end{tabular}
}
\label{tab:results-err-fullset}
\end{table*}

\begin{table*}[t]
\caption{Accuracy results for \internet, \cuboidrand, \cuboidopt (all losses), \dhoi and \lasr.
\gtobjmark indicates that the method has access to the GT object mask (all methods except \internet and \cuboidrand).  In addition, \cuboidopt has access to GT part mask, \dhoi requires CAD models and \ditto has access to ground-truth depth maps.
Motion accuracies are based on matches for the reconstruction and pose accuracies (i.e. within the error threshold for reconstruction and pose). Overall accuracies are based on matches for all the indicated combinations of parameters.
The rotation threshold is 10 degrees, the translation threshold is 0.5, and the scale threshold is 0.3.
}
\resizebox{\linewidth}{!}{
\begin{tabular}{@{} l rrr rrrr rrrr rrr @{}}
\toprule
& \multicolumn{3}{c}{Reconstruction Accuracy $\%$} & \multicolumn{4}{c}{Pose Accuracy $\%$} & \multicolumn{4}{c}{Motion Accuracy $\%$} & \multicolumn{3}{c}{Overall Accuracy $\%$} \\
\cmidrule(l{0pt}r{2pt}){2-4} \cmidrule(l{0pt}r{2pt}){5-8} \cmidrule(l{0pt}r{2pt}){9-12} \cmidrule(l{0pt}r{2pt}){13-15} 
Method & Object@0.5 & Moving@0.5 & \accr & Rot@10 & Trans@0.5 & Scale@0.3 & \accp & O@0.5 & OA@10 & OAD@10 & \accm@10 & \accrp & RPOA & \accrpm \\
\midrule
\internet~\cite{qian2022understanding} & $1.0$ & $38.6$ & $1.0$ & $2.8$ & $0.7$ &$8.5$ & $0.0$ & $15.3$ & $7.8$ & $4.4$ &$0.0$ & $0.0$ & $0.0$ & $0.0$ \\
\midrule
\cuboidrand & $ 8.7 $ & $ 10.6 $ & $ 1.0 $ & $ 0.3 $ & $ 0.8 $ & $ 43.2 $ & $ 0.0 $ & $ 0.0 $ & $ 0.0 $ & $ 0.0 $ & $ 0.0 $ & $ 0.0 $ & $ 0.0 $ & $ 0.0 $ \\ 
\cuboidopt\gtobjmark & $35.5$ & $65.5$ & $29.5$ & $19.2$ & $1.7$ &$62.9$ & $1.0$ & $43.9$ & $20.1$ & $16.3$ &$10.0$ & $1.0$ & $0.7$ & $0.5$ \\
\midrule
\dhoi-GT-CAD~\cite{xu2021d3d}\gtobjmark & $\textbf{76.0}$ & $\textbf{90.1}$ & $\textbf{74.3}$ & $44.7$ & $\textbf{41.8}$ &$86.8$ & $\textbf{33.9}$ & $84.9$ & $\textbf{59.8}$ & $\textbf{59.8}$ &$\textbf{38.7}$ & $\textbf{33.4}$ & $\textbf{33.3}$ & $\textbf{25.2}$ \\ 
\dhoi-Rand-Cat-CAD\gtobjmark & $64.5$ & $75.9$ & $51.9$ & $35.1$ & $23.5$ &$83.1$ & $18.4$ & $73.6$ & $42.7$ & $41.8$ &$22.4$ & $16.1$ & $15.1$ & $9.9$ \\
\dhoi-Rand-CAD\gtobjmark & $38.8$ & $60.7$ & $26.2$ & $20.8$ & $8.2$ &$77.2$ & $5.9$ & $59.8$ & $18.8$ & $13.4$ &$5.7$ & $4.4$ & $2.7$ & $1.8$ \\
\midrule
\ditto~\cite{jiang2022ditto} & $45.3$ & $15.6$ & $6.4$ & $\textbf{60.7}$ & $17.1$ &$\textbf{97.8}$ & $10.5$ & $\textbf{91.1}$ & $3.5$ & $0.4$ &$0.0$ & $0.7$ & $0.0$ & $0.0$ \\

\lasr~\cite{yang2021lasr}\gtobjmark & $14.1$ & $5.6$ & $1.3$ & $4.1$ & $0.0$ &$38.7$ & $0.0$ & $0.0$ & $0.0$ & $0.0$ &$0.0$ & $0.0$ & $0.0$ & $0.0$ \\

\viser~\cite{yang2021viser}\gtobjmark & $7.9$ & $7.3$ & $0.1$ & $5.4$ & $0.0$ &$0.5$ & $0.0$ & $0.0$ & $0.0$ & $0.0$ &$0.0$ & $0.0$ & $0.0$ & $0.0$ \\

\bottomrule
\end{tabular}
}
\label{tab:results-acc-fullset}
\end{table*}

\section{Experiments}

\subsection{Data}
We conduct our experiments on the D3D-HOI dataset~\cite{xu2021d3d} which contains 256 RGB videos of a person interacting with nine types of common household objects.
We choose this dataset because it provides challenging scenarios of humans interacting with common household objects.
It is also the only human-object interaction dataset with fine-grained 3D annotations for the object shape and motion parameters.
Each video is annotated with an articulated 3D CAD object which is aligned with the RGB image.
In addition, \citet{xu2021d3d} provide an estimated SMPL~\cite{loper2015smpl} mesh of the person predicted by running EFT~\cite{joo2021exemplar}.
Using the annotated CAD object we generate object and moving part segmentation masks which we treat as the ground truth segmentation masks in our experiments.
We conduct our experiments on eight object categories with revolute motion which amount to a total of 239 videos.

\subsection{Metrics}

We evaluate the five types of methods by measuring their performance at: 1) object shape reconstruction; 2) object pose estimation; and 3) motion parameter estimation (for the articulating part).
For each, we define error metrics and accuracy metrics based on thresholding the errors.  All metrics are computed per frame, and then averaged over each video, and then across videos.
We adopt Chamfer Distance (CD) for our reconstruction metric~\cite{fan2017point}, and translational, rotational and scale errors as object pose metrics~\cite{xiang2017posecnn,wang2019normalized}.
The motion parameter estimation metrics are based on computing the error for individual motion parameters.

\mypara{Reconstruction.}
To evaluate the quality of the reconstructed articulated object and its parts, we compute the Chamfer Distance (CD) between $10000$ sampled surface points from the estimated 3D representation and the ground truth.
Since our goal for this metric is to measure reconstruction quality, we factor out pose by normalizing the scale of both the predicted mesh and the ground truth mesh by their maximum dimension.
We then apply ICP~\cite{arun1987least} to align the two meshes.
This allows us to use the CD metric to capture the reconstruction error. %
We also measure the CD for both the entire object and the moving part in isolation.
For the accuracy metric, we compute a combined accuracy (\accr) that combines the thresholded reconstruction error accuracy for both the whole object and the moving part.

\mypara{Pose.}
The scale normalization and ICP alignment from above are also used to evaluate the object pose.
We measure the rotation and translation error based on the prediction-to-ground-truth mesh alignment transformation matrix calculated by ICP.
For the rotation error, we calculate the error in degrees between two rotation matrices.
For the translation error, we calculate the Euclidean distance.
For the scale error, we compute the ratio between the maximum dimension of the ground truth object and the predicted object.
If the ratio is larger than $1$ we take its inverse.
Then we subtract this value from $1$ to obtain a relative scale error value in $[0,1]$.
For the accuracy metric, we define a combined rotation, translation and scale match accuracy (\accp).

\mypara{Motion.}
We evaluate the error of the predicted motion origin point, motion axis (ignoring direction since parametrizations based on both directions are possible), motion axis with direction, and finally motion state (rotation angle).
We compute the angle error for the axis and direction, and the distance to the axis for the origin.  
Note that the maximum angle error is $90^\circ$ for the motion axis and $180^\circ$ for the motion axis direction.
We report accuracy metrics for these motion parameters in staggered combinations, starting with motion origin (O), origin+axis (OA), OA+Direction (OAD), and overall motion parameter accuracy including all motion parameters: axis, direction, state error (\accm).

\mypara{Overall.}
Finally, we report combined accuracy for reconstruction and pose (\accrp),
and combined accuracy for reconstruction, pose, and motion state (\accrpm).

\subsection{Implementation Details}

For 3DADN~\cite{qian2022understanding}, D3D-HOI~\cite{xu2021d3d}, LASR~\cite{yang2021lasr}, \viser~\cite{yang2021viser} and \ditto~\cite{jiang2022ditto} we use the open-sourced implementations provided by the authors to evaluate these methods on our benchmark.
For \ditto, we use the pretrained network released by the authors to run inference on our data.
We implement the CuboidOPT method in PyTorch3D~\cite{ravi2020pytorch3d} and minimize the final objective function using gradient based optimization.
We use PyTorch3D's differentiable renderer to render the object and part masks used to compute the loss functions.
We do not assume any ground truth information about the object shape and human-object interaction contact point so the optimization is over all templates and left/right hand combinations for human-object interaction terms.
We select the optimization run achieving minimum loss.
We use Adam~\cite{kingma2014adam} as our optimizer with a learning rate of $0.05$ and a decay factor of $10$ in the last $25\%$ of the iterations.
We optimize each model for a total of $500$ iterations. 

\begin{figure*}
\includegraphics[width=\linewidth]{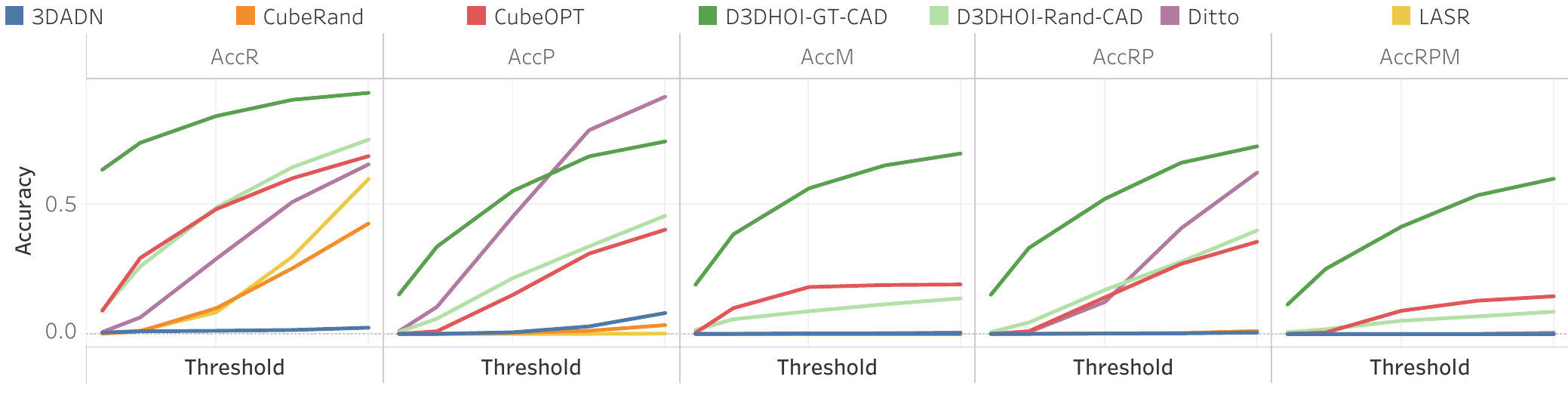}
\caption{Accuracy of reconstruction (\accr), pose (\accp), motion (\accm), reconstruction+pose (\accrp), reconstruction+pose+motion (\accrpm) at varying thresholds: $<0.25,0.5,1.0,1.5,2.0$ for reconstruction (CD) and axis origin error, $<0.2,0.3,0.4,0.5,0.6$ for scale error, $<5,10,15,20,25^\circ$ for angle error (pose rotation, motion parameters).
As the threshold is increased, the strictness of the condition for a prediction to be correct is relaxed, and the accuracy increases.
\dhoi-GT-CAD which relies on having the GT CAD model in addition to the GT object mask has the best overall performance.
The cuboidal abstraction method \cuboidopt has access to the GT object and part image mask, and the free-form mesh fitting \lasr has access to the GT object mask.
The latter is able to reconstruct the object somewhat but ends up underperforming the randomly initialized \cuboidopt, showing that having a cuboidal assumption is helpful for many articulated objects.
\ditto has access to GT point clouds of the objects and therefore outperforms all other baselines in pose estimation.
\internet is a 2D method that does not have access to GT information and leverages pretraining on large amounts of internet videos.
Since it models only the moving part, it cannot reconstruct the static part and has poor overall reconstruction performance.
All methods other than \cuboidopt and \dhoi are unable to predict the combined motion parameters, and overall pose, indicating the signifcant challenges of this task.
}
\label{fig:accuracy-plots}
\end{figure*}

\subsection{Results}
\begin{figure*}
\setkeys{Gin}{width=\linewidth}
\footnotesize
\begin{tabularx}{\textwidth}{@{} Y Y Y Y Y Y @{}}
GT & \internet & \cuboidopt & \dhoi-GT-CAD & \ditto & \lasr \\

\imgclip{0}{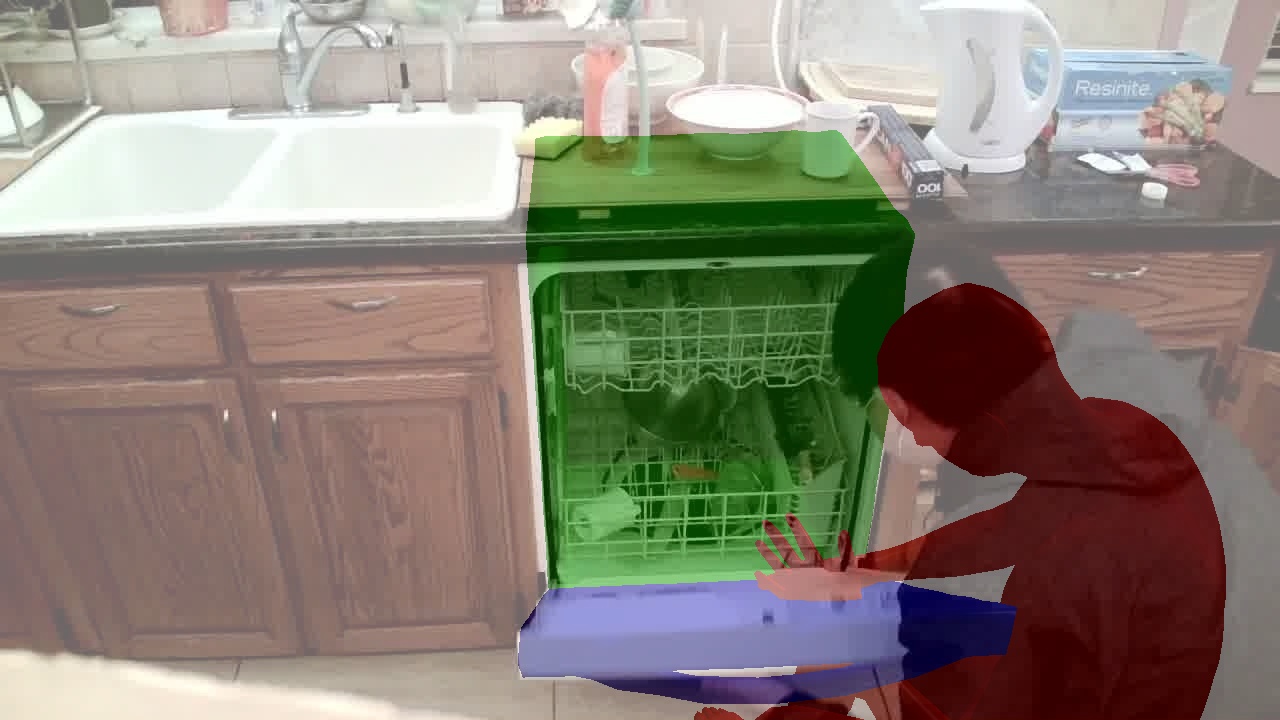} & 
\imgclip{0}{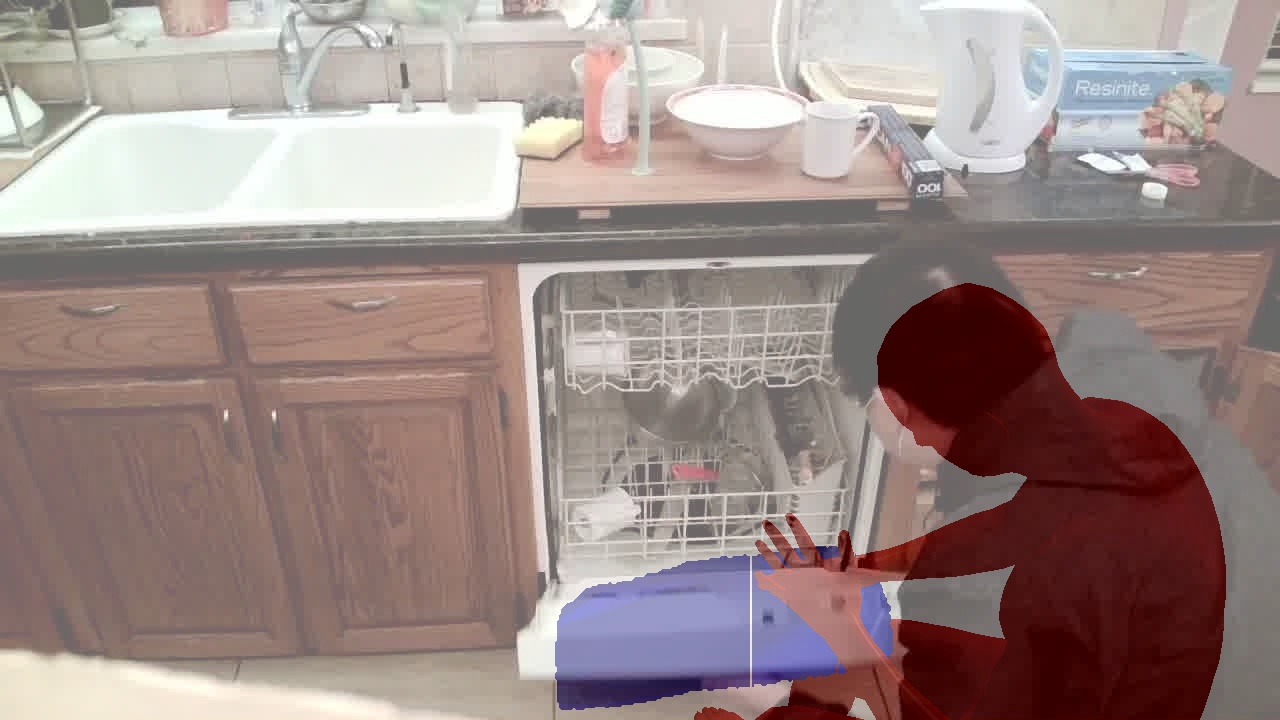} & 
\imgclip{0}{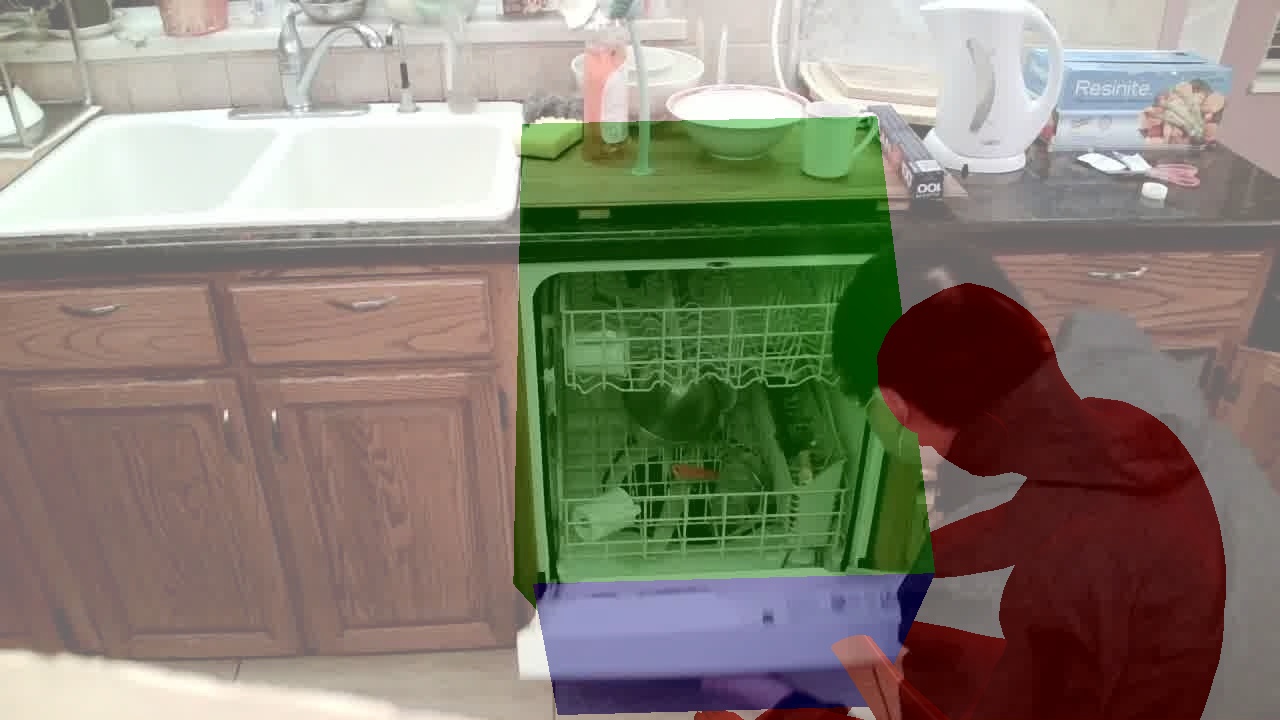} & 
\imgclip{0}{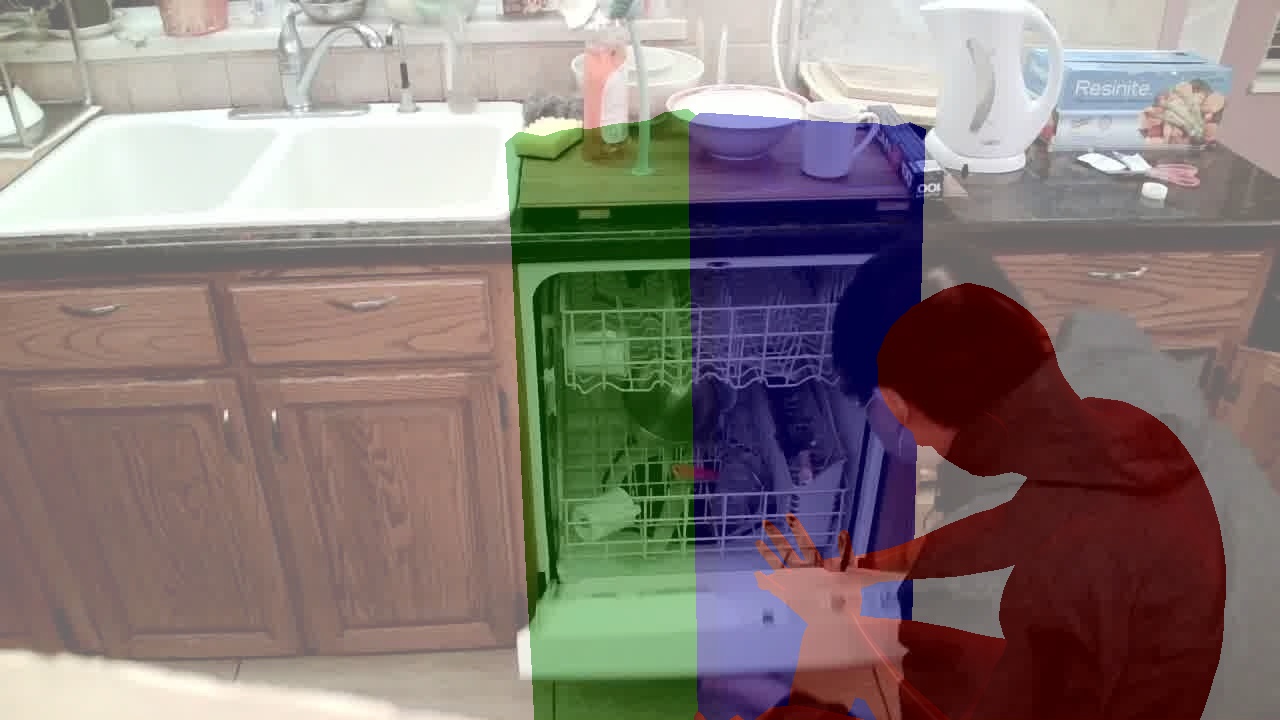} &
\imgclip{0}{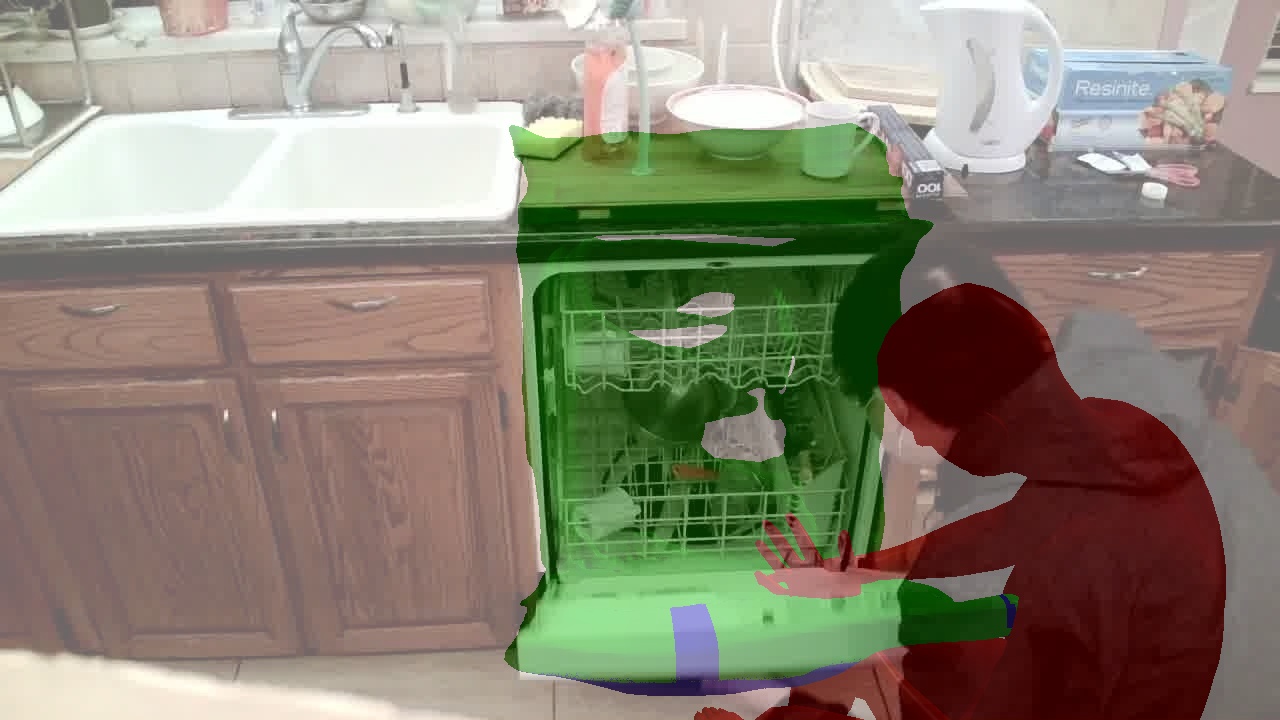} & 
\imgclip{0}{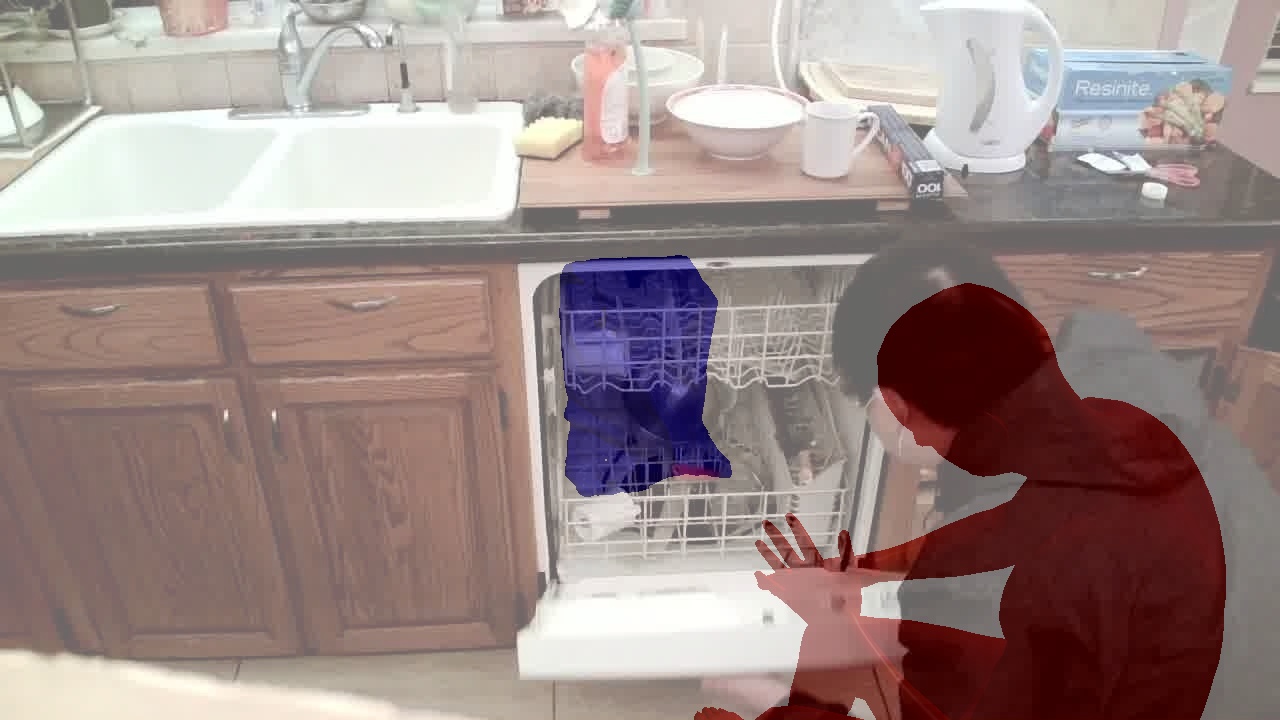}
\\

\imgclip{0}{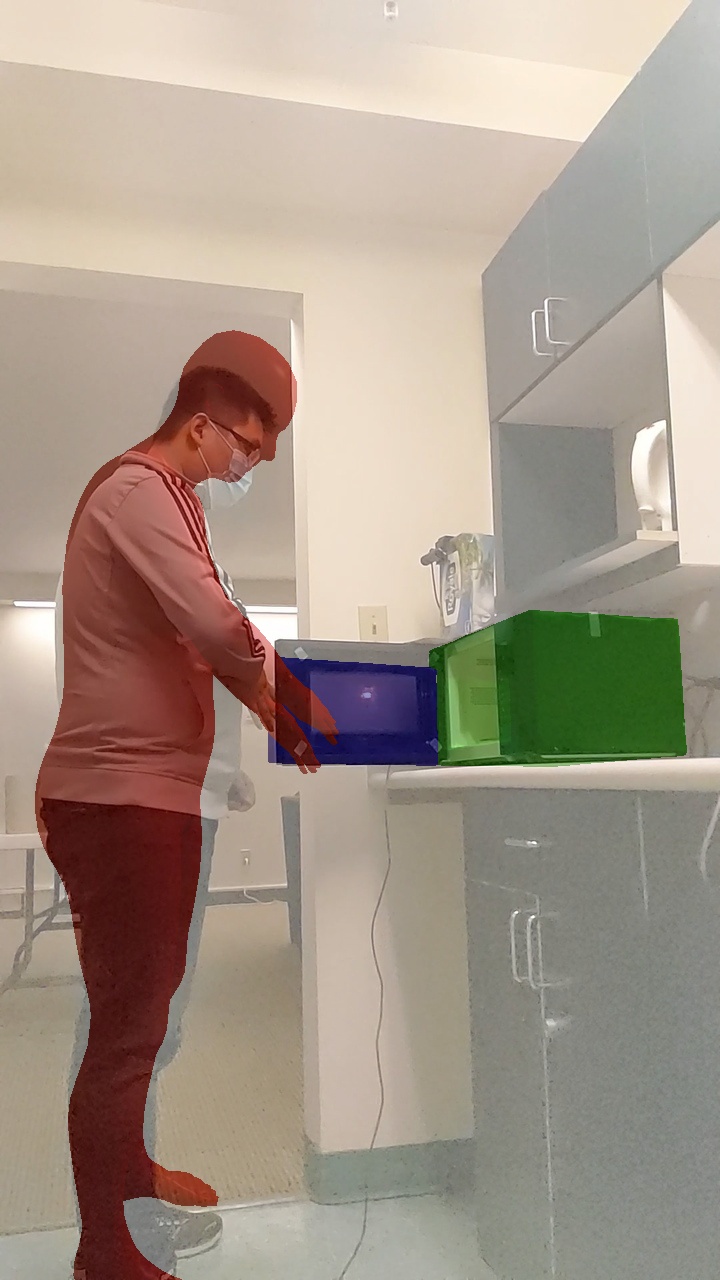} & 
\imgclip{0}{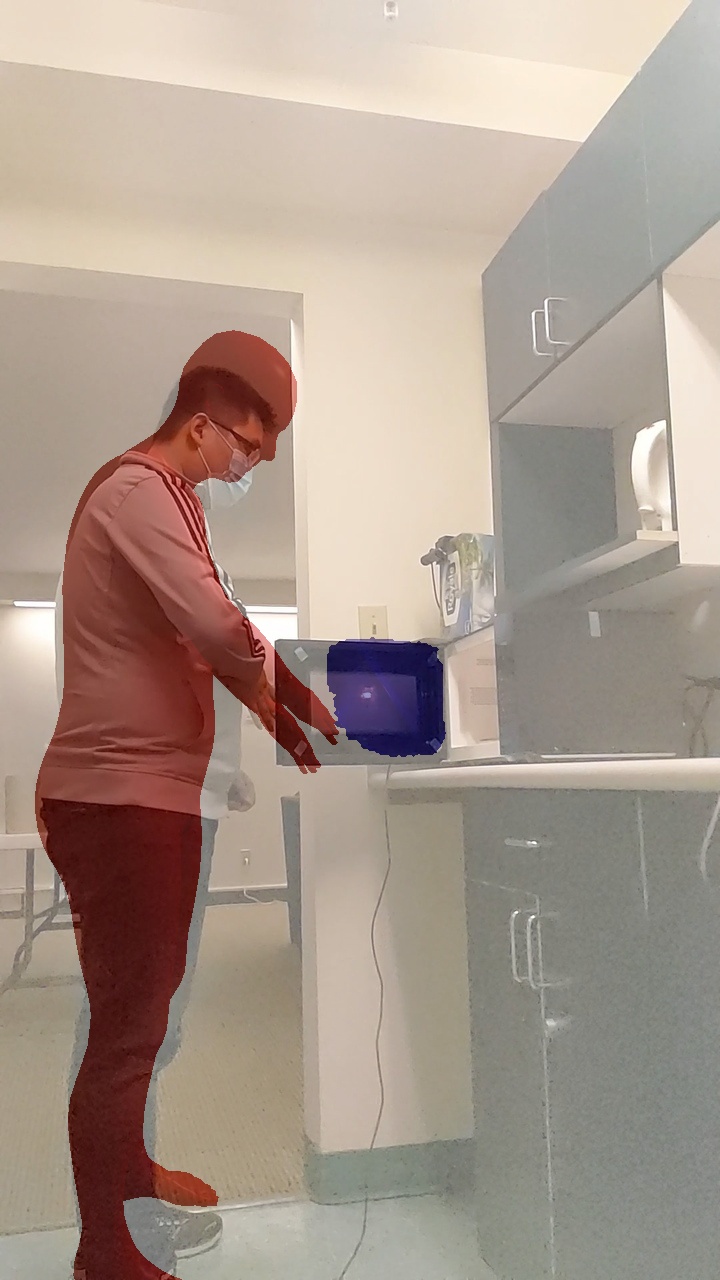} & 
\imgclip{0}{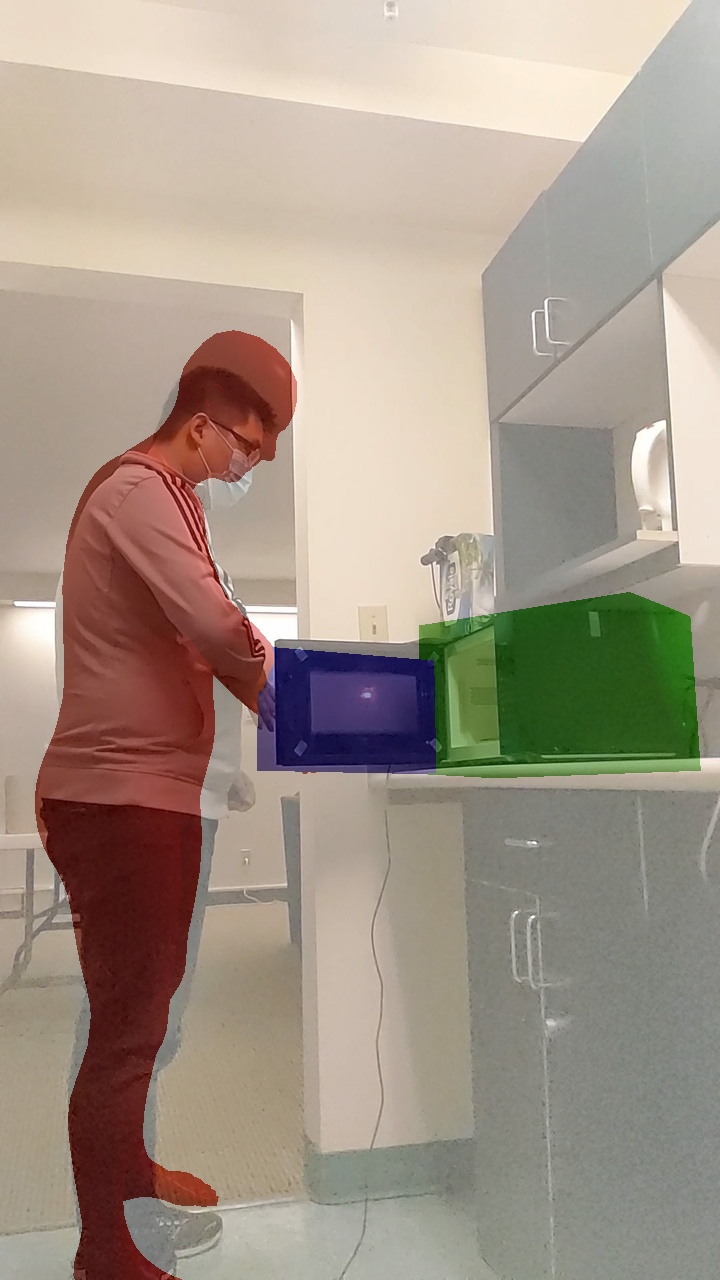} & 
\imgclip{0}{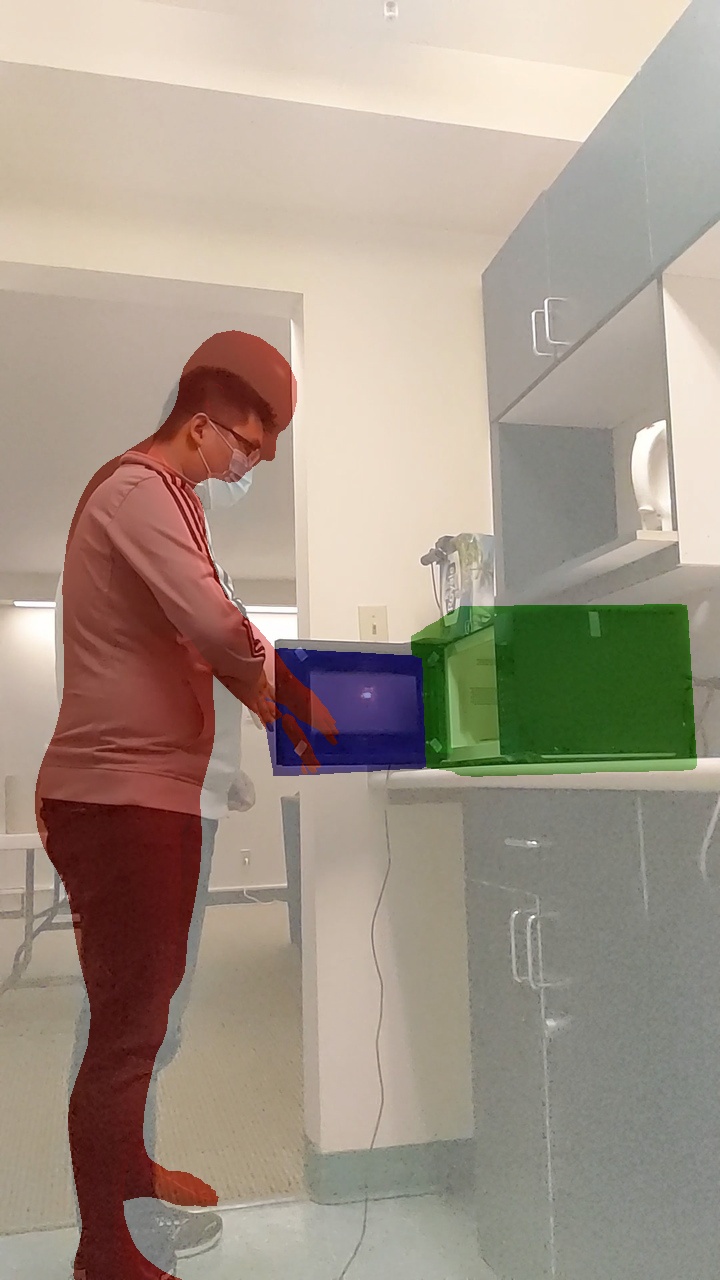} & 
\imgclip{0}{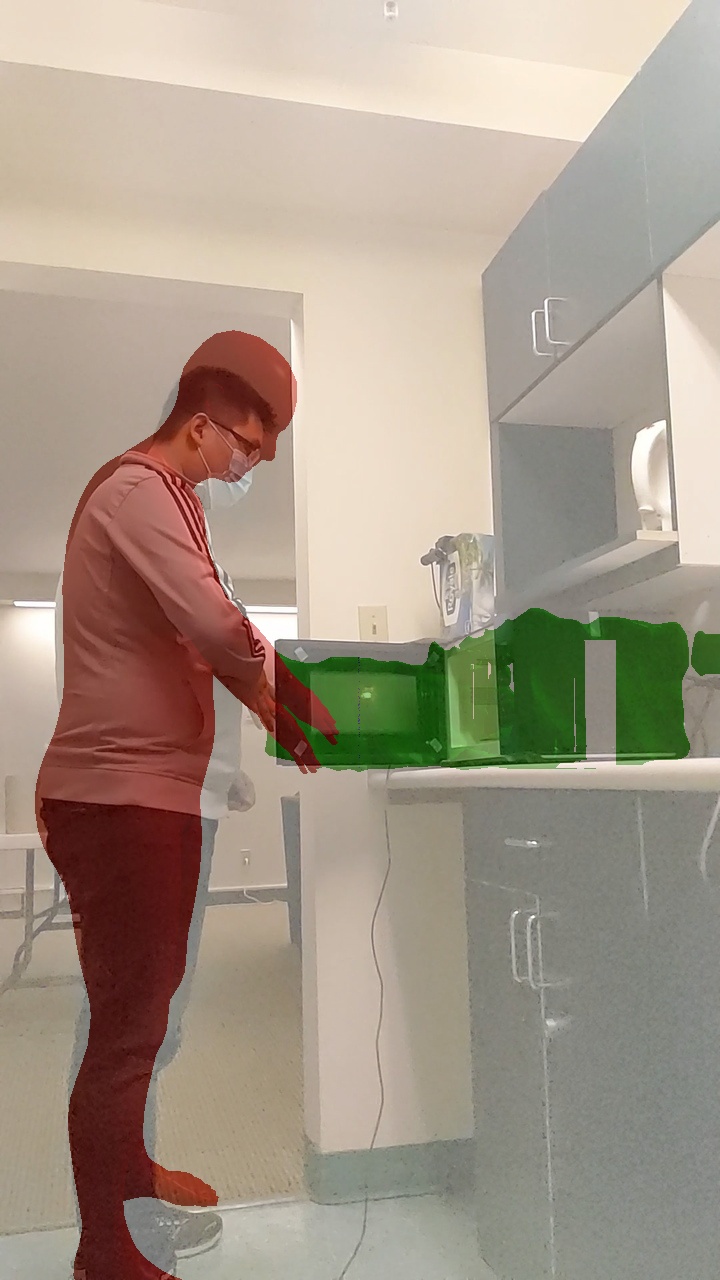} & 
\imgclip{0}{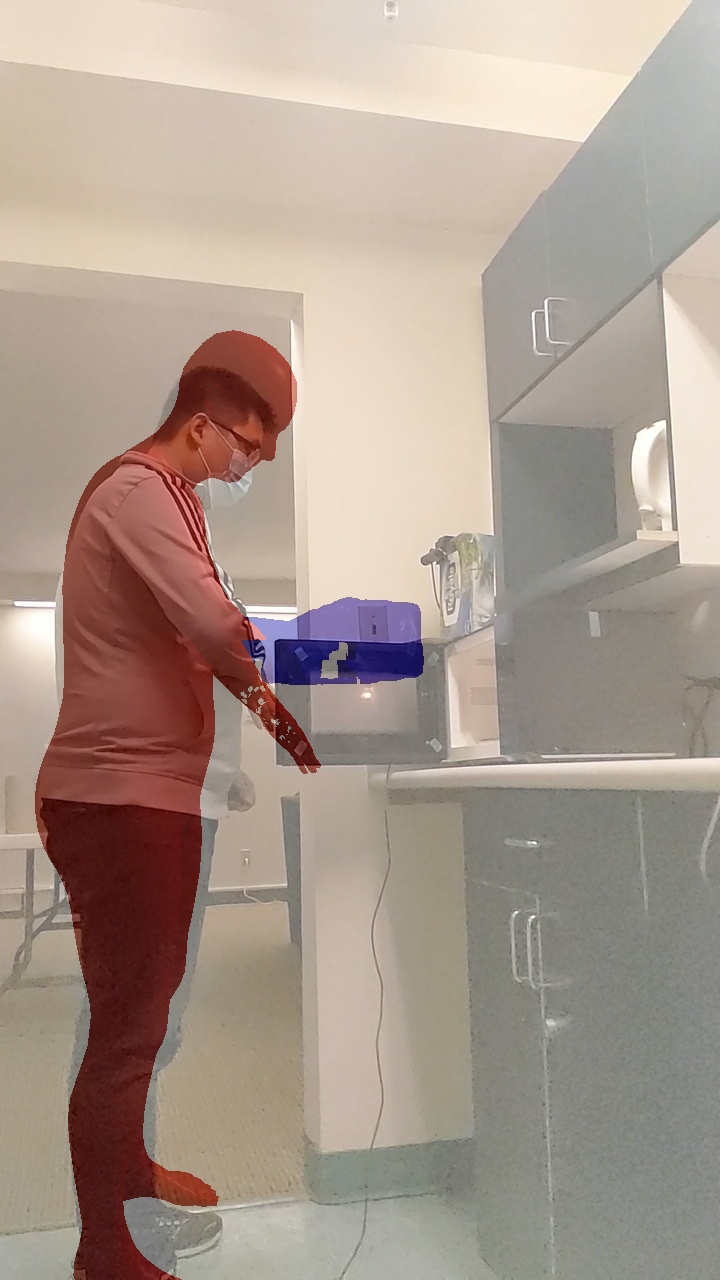}
\\

\imgclip{0}{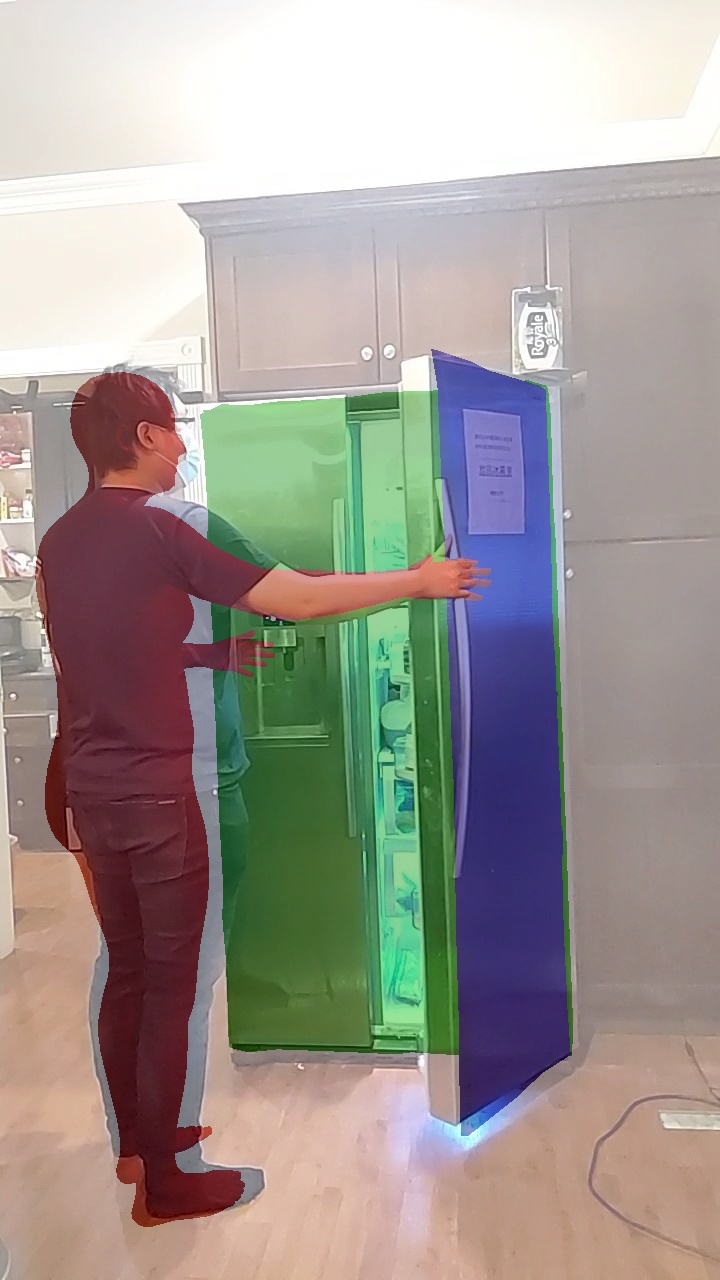} & 
\imgclip{0}{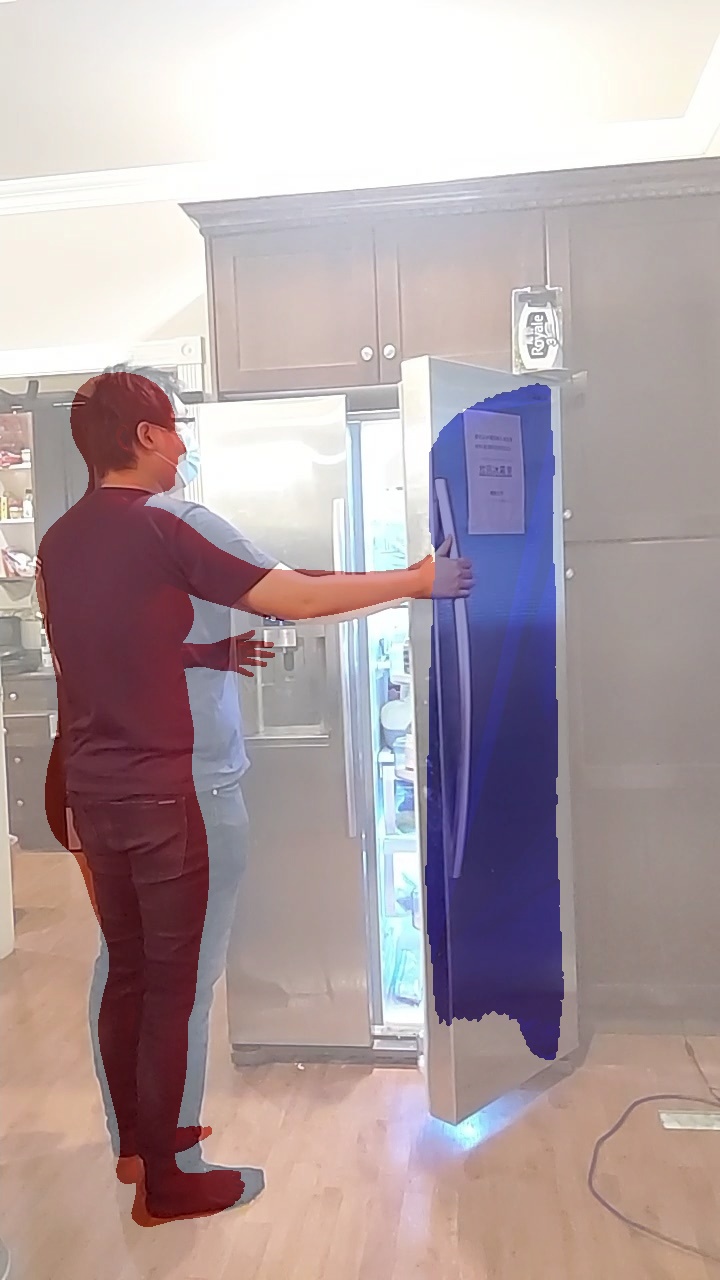} & 
\imgclip{0}{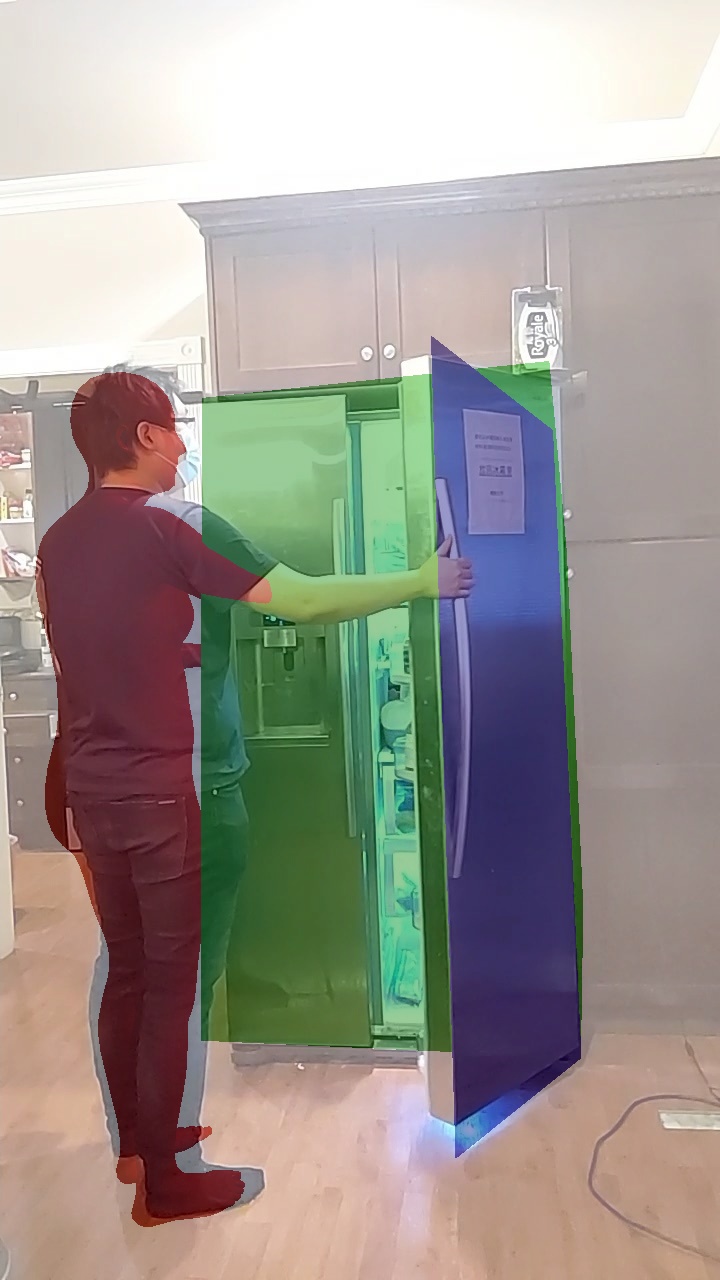} & 
\imgclip{0}{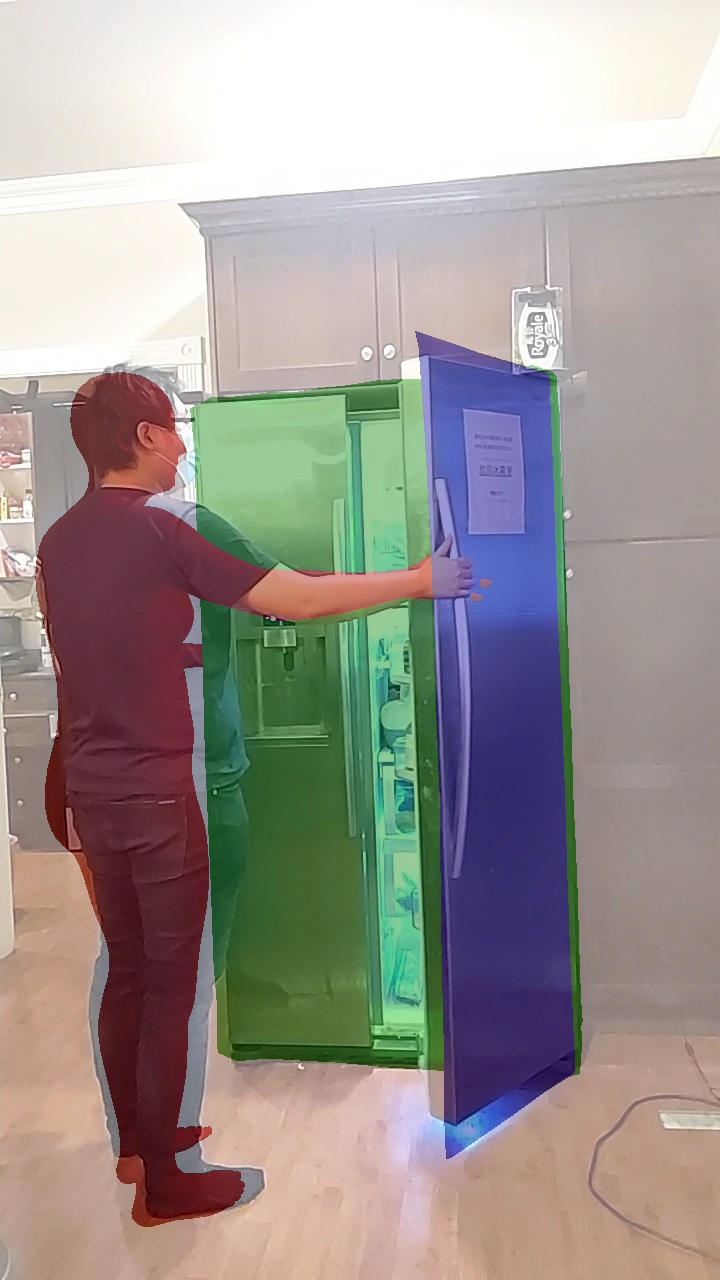} & 
\imgclip{0}{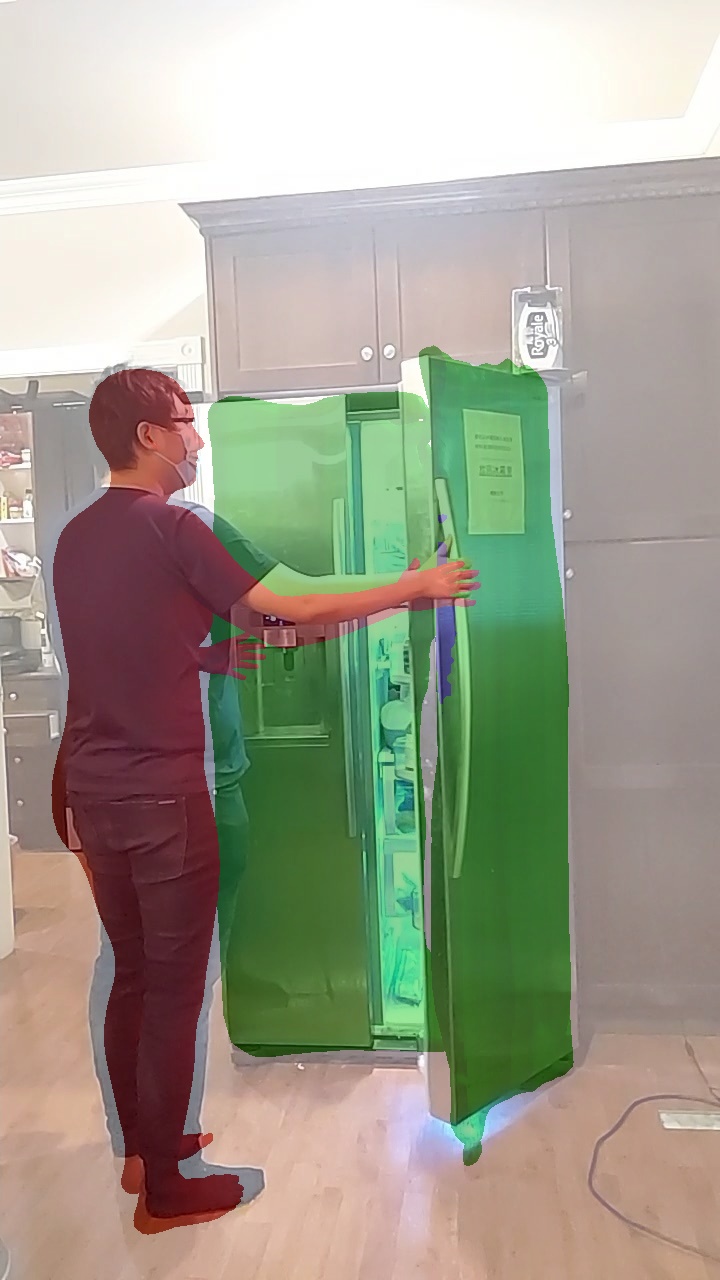} & 
\imgclip{0}{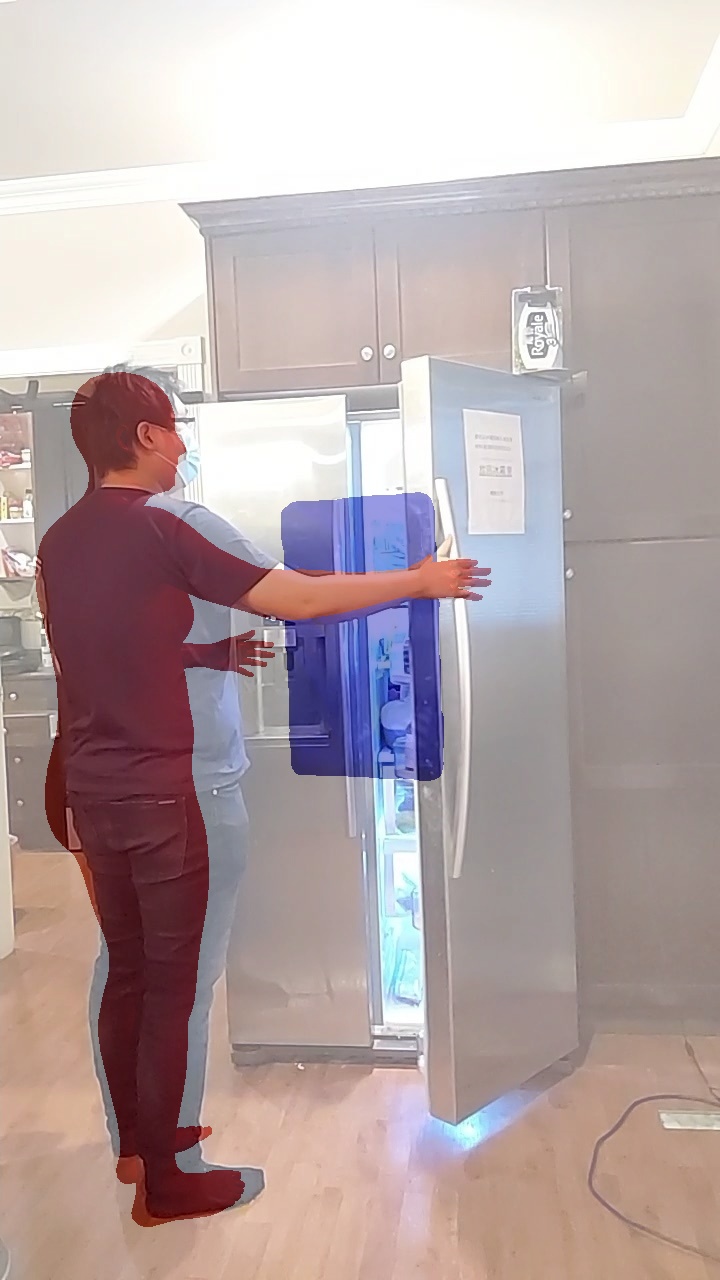} \\

\imgclip{0}{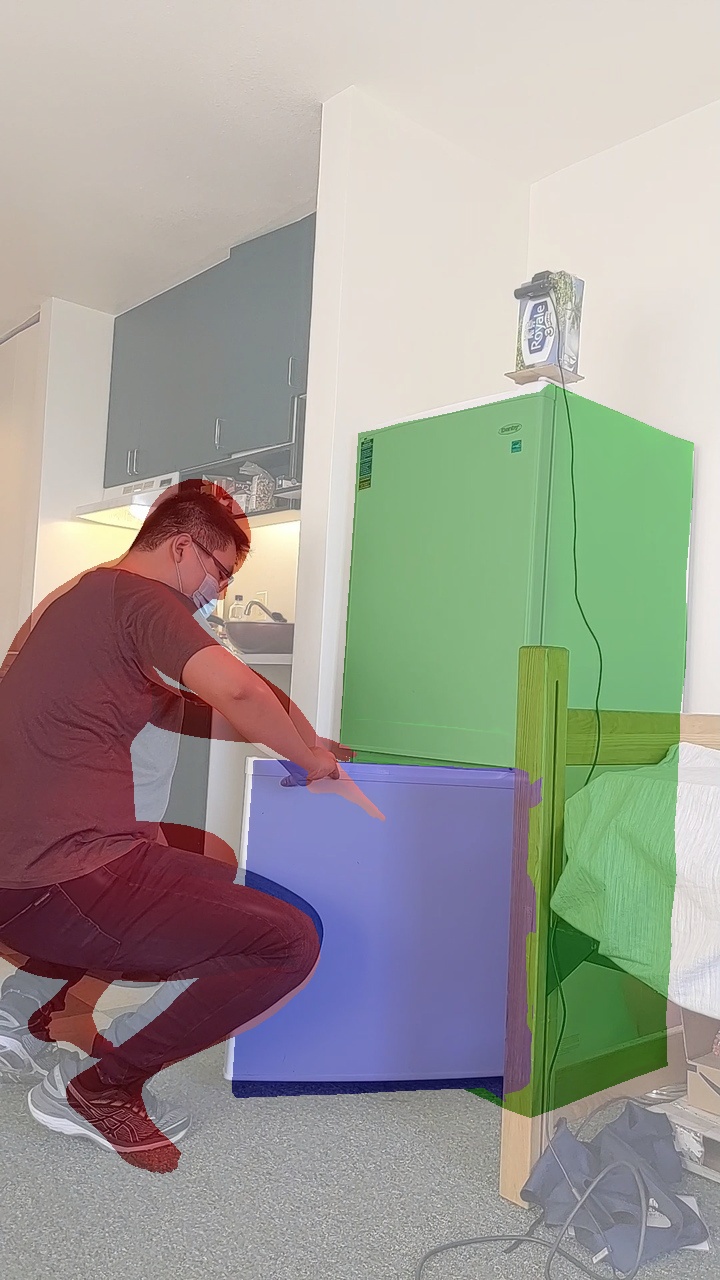} & 
\imgclip{0}{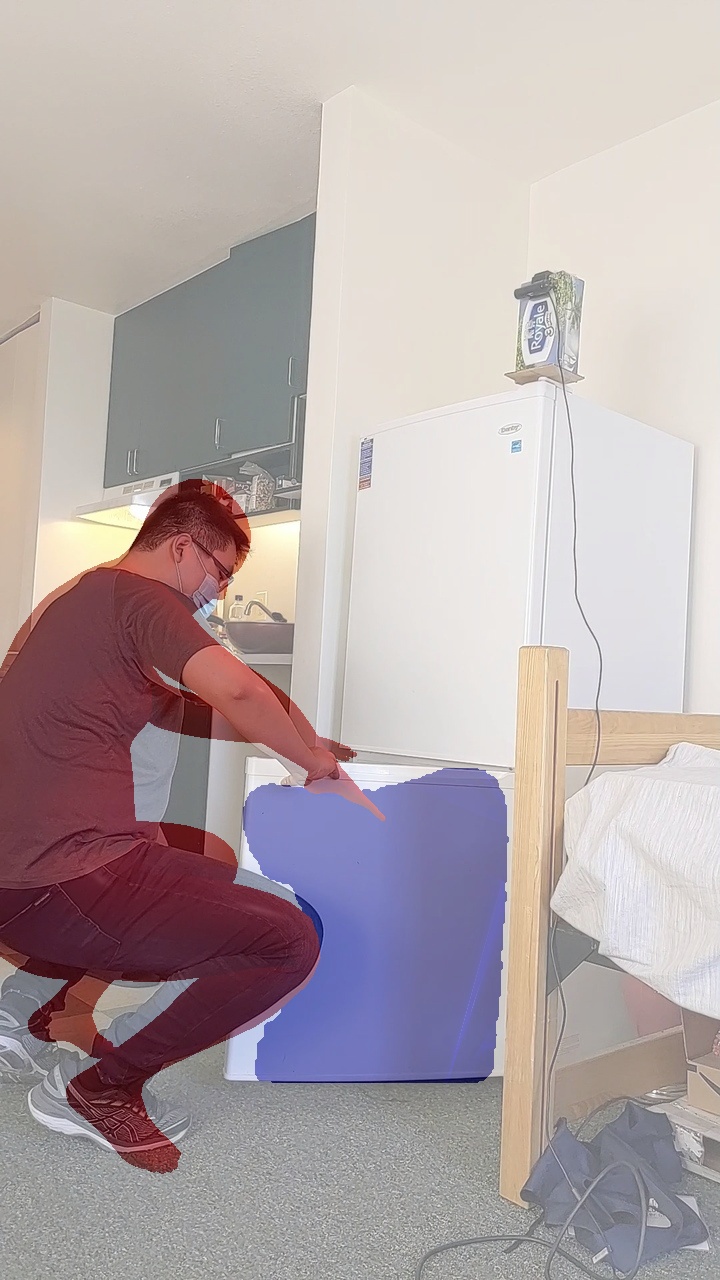} & 
\imgclip{0}{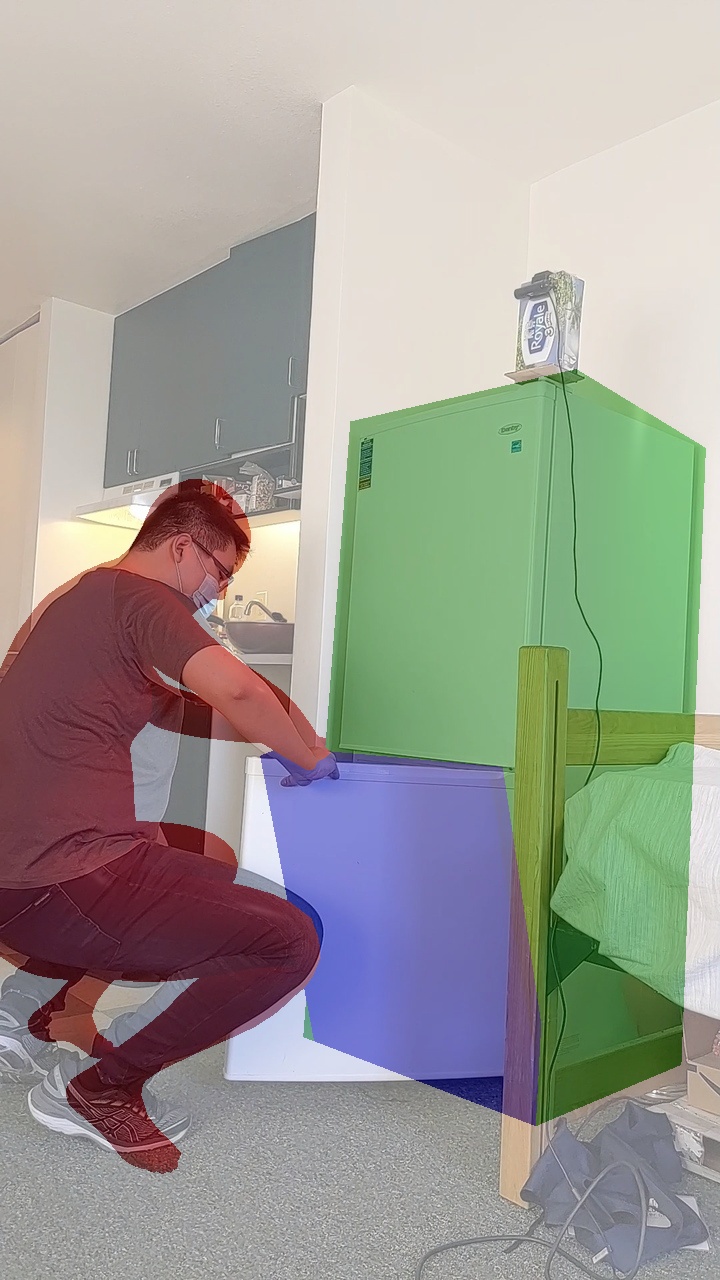} & 
\imgclip{0}{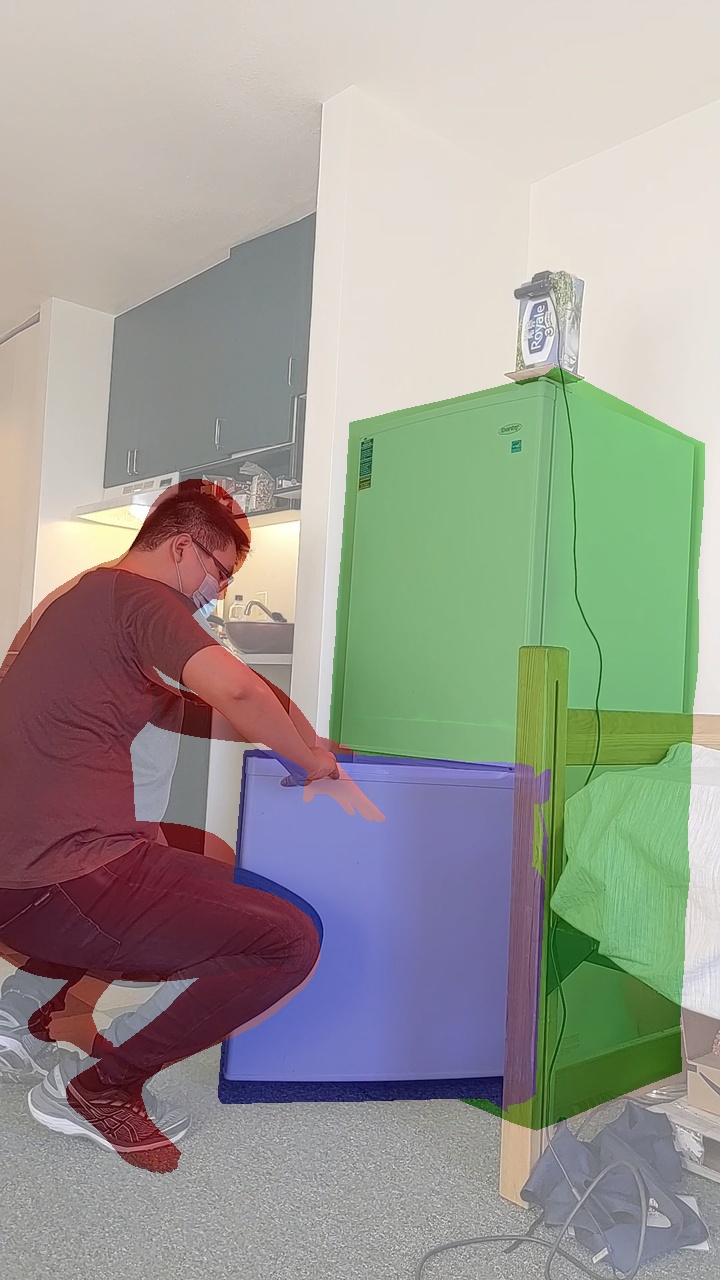} &
\imgclip{0}{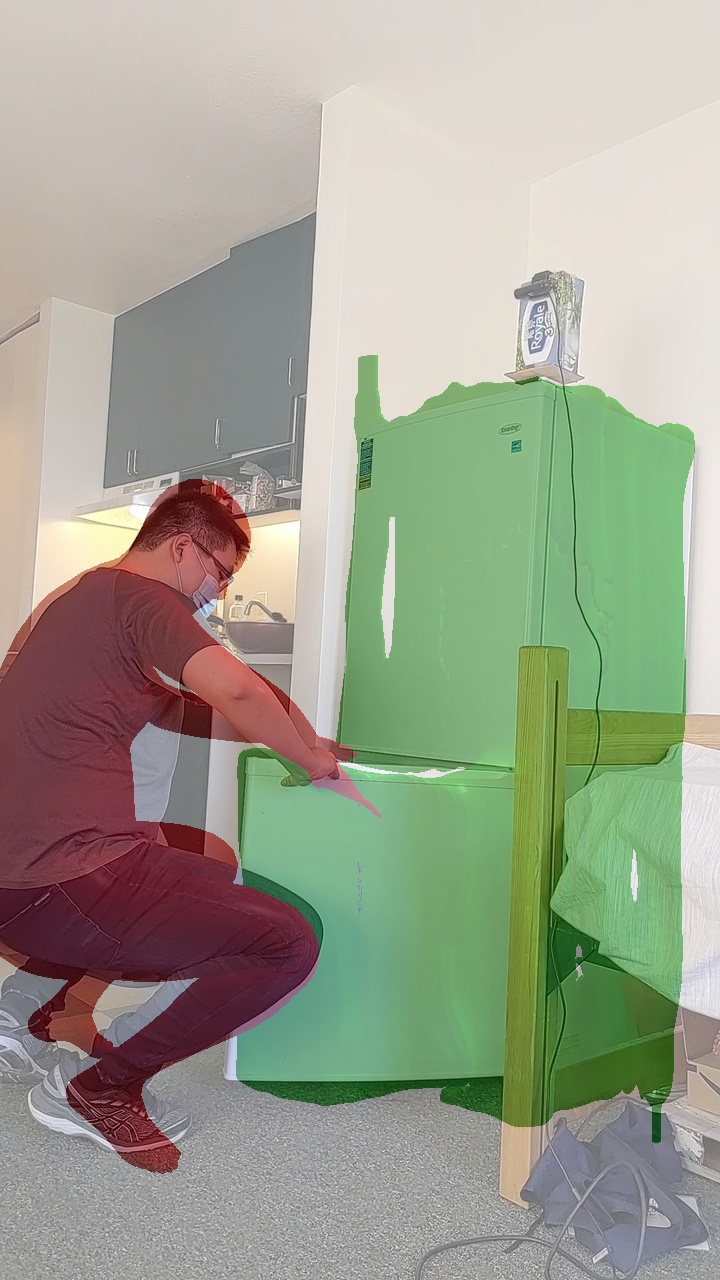} & 
\imgclip{0}{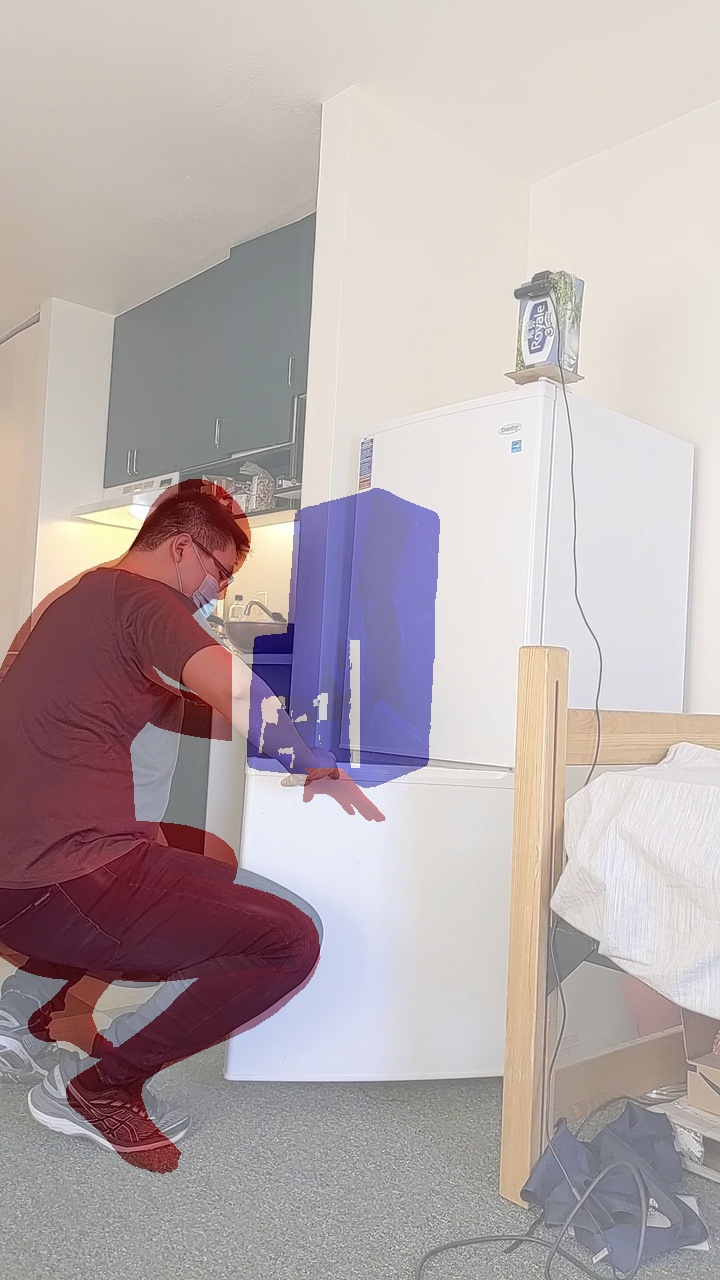} \\

\imgclip{0}{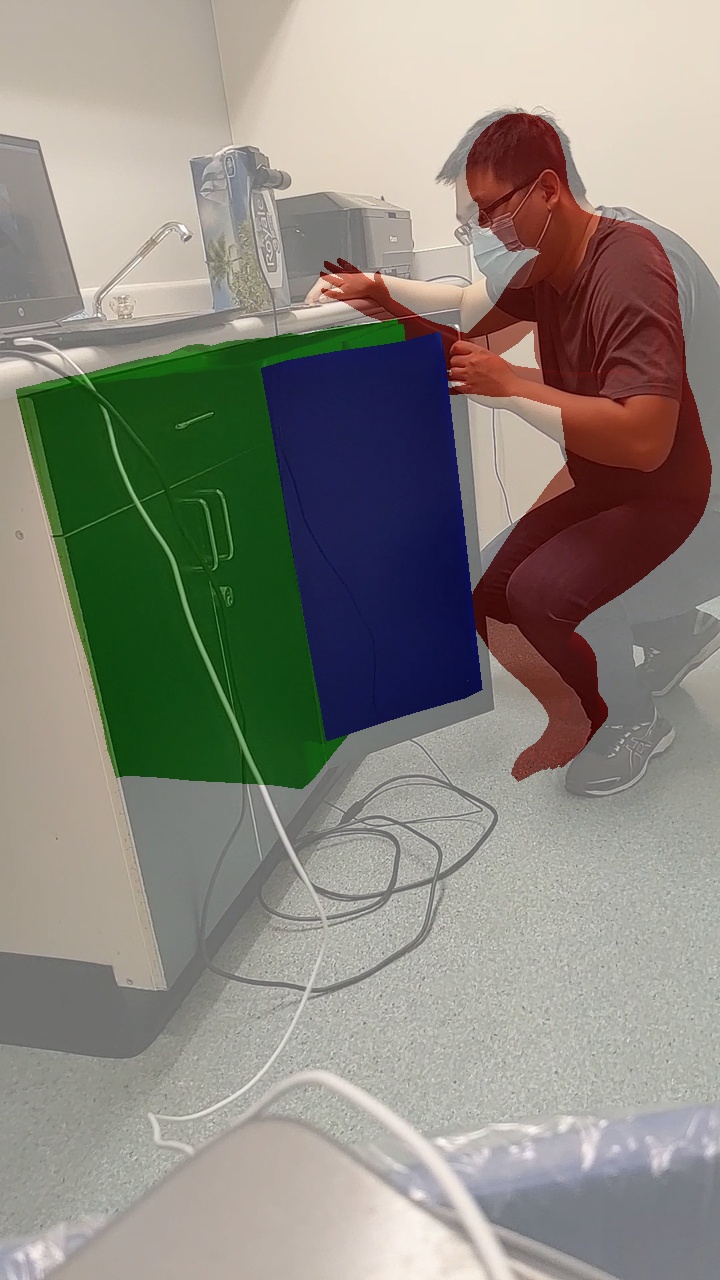} & 
\imgclip{0}{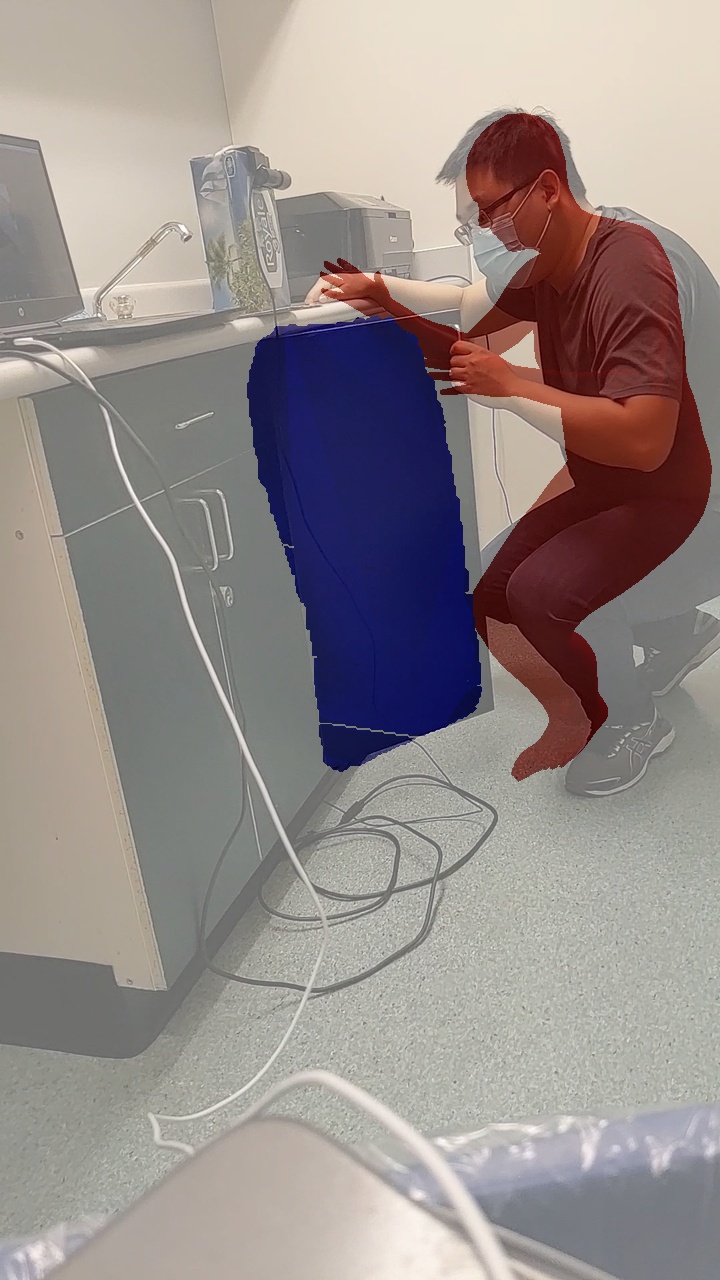} & 
\imgclip{0}{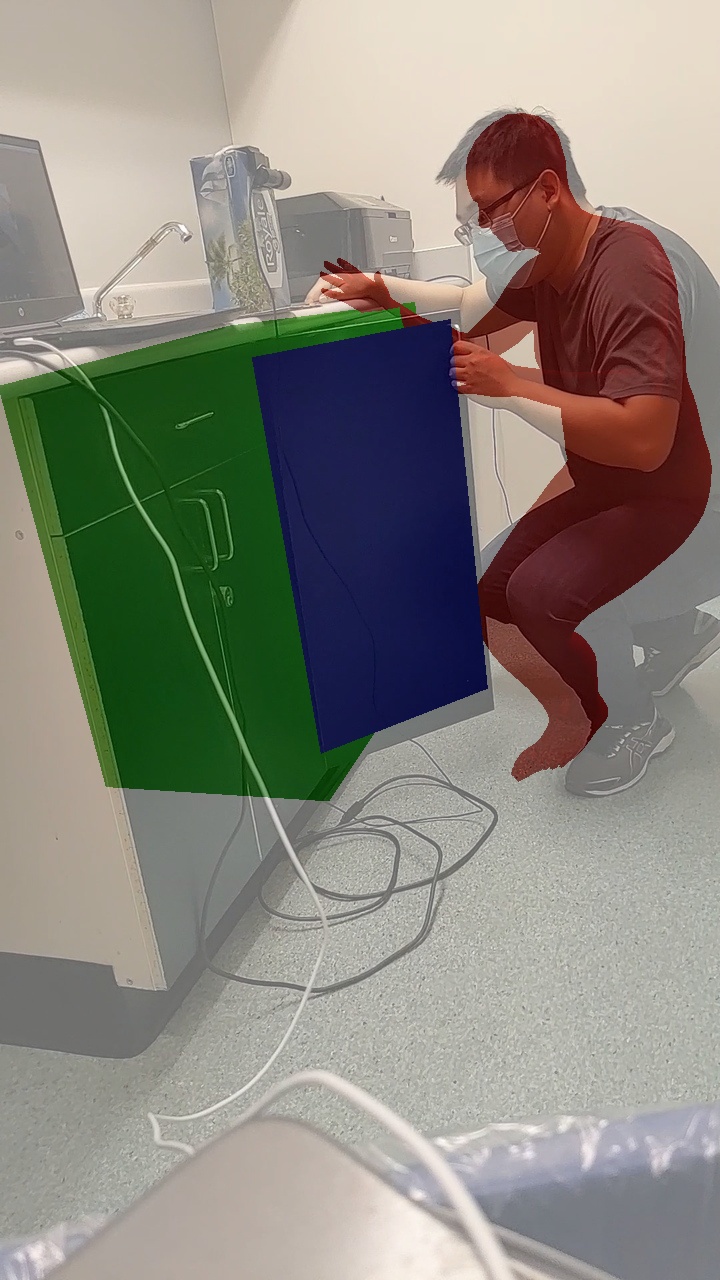} & 
\imgclip{0}{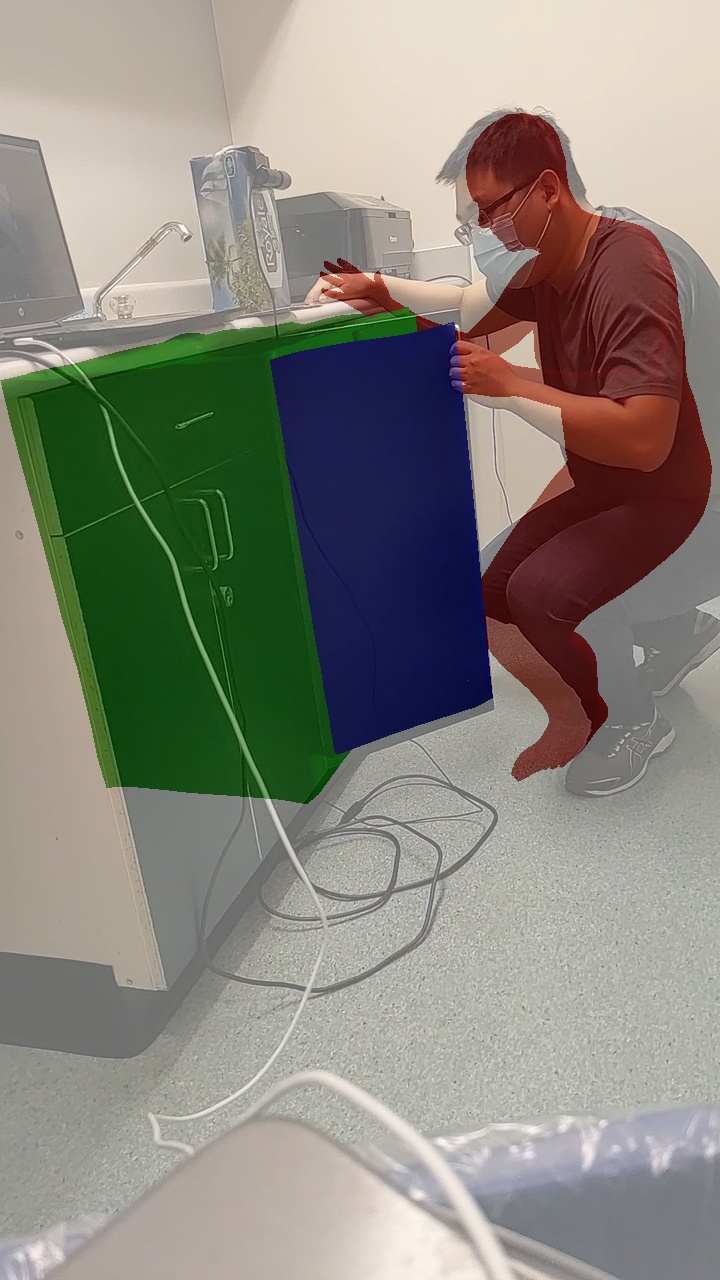} &
\imgclip{0}{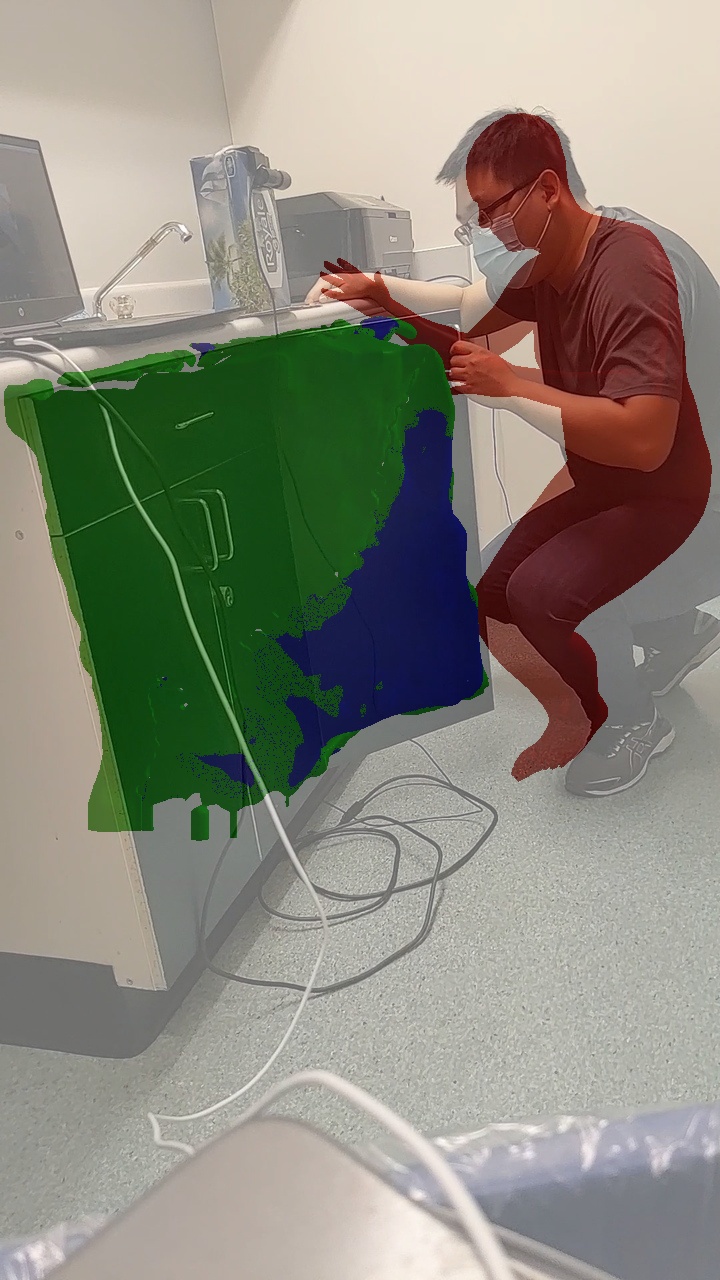} & 
\imgclip{0}{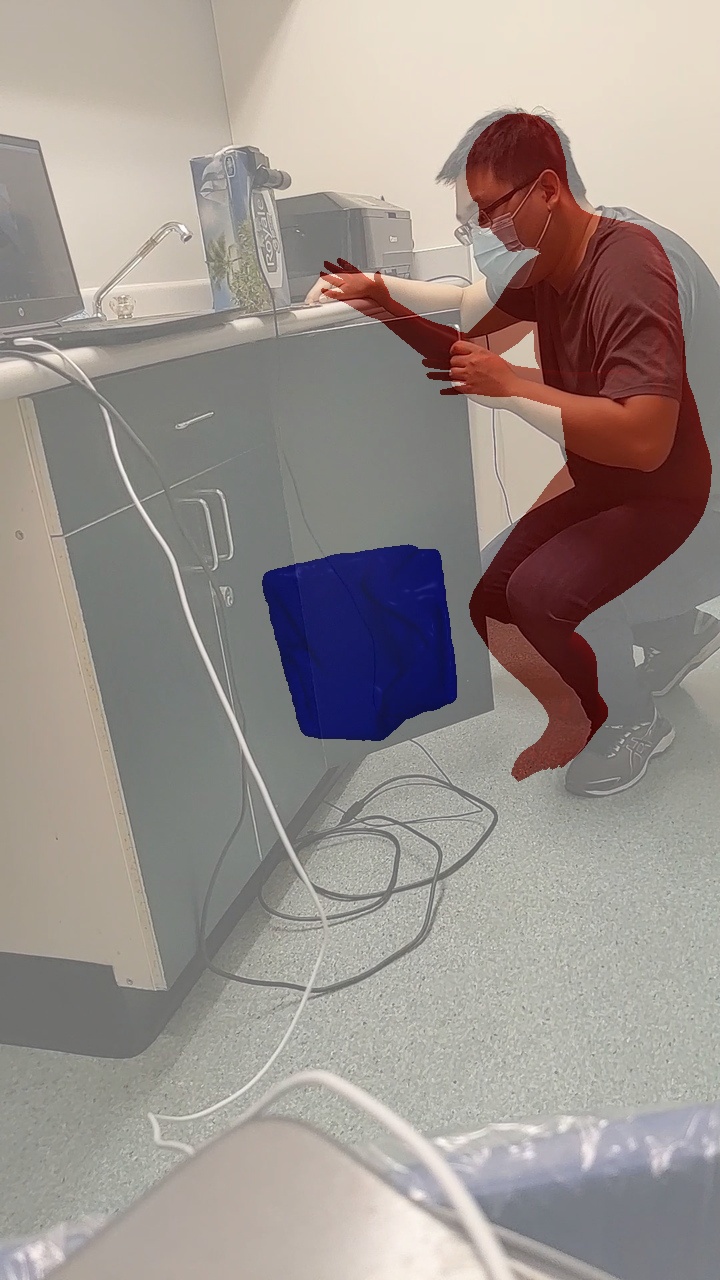}\\

\end{tabularx}
\caption{
Qualitative comparison of methods on several example videos.
Obtaining high accuracy results for articulated 3D human-object interaction is quite challenging.
All approaches (other than \internet) are given ground-truth object segmentation masks but still exhibit significant errors in object shape reconstruction (in particular \internet which only handles the moving part of the object), and in motion parameter estimation (in particular \cuboidopt which gets the motion axis wrong in the second to last row from the bottom). The \lasr approach often struggles to estimate a reliable pose for the reconstructed object.
}
\label{fig:qualitative}
\end{figure*}

We report error metrics for the five types of approaches in \Cref{tab:results-err-fullset}, 
and thresholded accuracies in \Cref{tab:results-acc-fullset}.
We also plot accuracies against error threshold for the accuracies in \Cref{fig:accuracy-plots}.
Finally, we show qualitative results in \Cref{fig:qualitative}.
Most of the methods in this comparison have access to privileged information: \lasr has access to the ground-truth (GT) object mask for each video frame, and \cuboidopt has access to the GT object and part masks. 

\mypara{How well does free-form mesh fitting work?}
\lasr is the most flexible in shape reconstruction, and gives comparable object CD as \cuboidopt.
However, it often has large errors in the pose of the object in camera coordinates.
To extract motion parameters from \lasr and \viser outputs, we use the predicted bone weights to obtain a part segmentation of the object. We then extract part boundary vertices and compute the dominant axis direction using PCA. The median of the boundary vertices is used as the origin.
We note that \viser outputs typically have a very different pose compared to GT and therefore struggles more on motion parameter estimation. On the other hand, \ditto performs well on reconstruction and pose parameter estimation. This may be attributed to the fact that \ditto uses GT point clouds.

\mypara{How much does knowing the ground-truth CAD model help?}
\dhoi is the most privileged method with access to the GT object mask and the GT CAD model that captures the shape of the object, the moving part and how it is attached to the static part.
There is limited need for reconstruction (though the scale factors $\sigma$ still need to be estimated). Instead, the problem is reduced to just pose and motion parameter estimation.
Unsurprisingly, \dhoi has the best overall performance (in terms of reconstruction, pose, and motion errors).
Reconstruction is still not perfect due to errors in estimating the scale parameters of the CAD model.
For a fairer comparison, we also consider fetching a random model (over all models vs using one with the GT category).
Performance deteriorates but is still fairly high for the \dhoi-Rand-Cat-CAD as there is a limited number of models to sample from (24 overall, and on average about two given the category).
\dhoi-Rand-CAD is comparable to \cuboidopt, better on some metrics (moving part CD, pose rotation and translation), while worse on others (motion prediction).

\mypara{How well can na\"{\i}ve cuboidal approximation of object shape work?}
We also evaluate \cuboidrand, a ``random'' baseline based on \cuboidopt with random initialization.
Note that this is stronger than a purely random baseline as \cuboidopt already has inductive biases due to the prior knowledge of the template structures that are built-in.
While \cuboidrand performs worse than the other methods in most of the error metrics, the pose error is actually similar to that of the optimized \cuboidopt and \dhoi-Rand-CAD, indicating that a simple approximation with two cuboids captures the articulated object fairly well.
From the accuracy results (\Cref{tab:results-acc-fullset}), the weakness of this random baseline becomes more obvious, as it has extremely low accuracy on object reconstruction, pose rotation/translation, and zero accuracy on motion prediction metrics.
The supplement provides additional ablations that investigate the impact of the loss terms in the \cuboidopt approach.

\mypara{What if we used a plane approximation and had no access to GT object masks?}
\internet is given the least information.
It is the only approach that does not assume access to the GT object mask.
However, it can only model the moving part, so it has the highest object-level CD, but relatively low CD for the moving part.
The pose error is also the highest of the methods, and motion error is also relatively high.
\internet does not predict the motion state (i.e. rotation angle) for every frame as its temporal optimization algorithm selects subsets of frames from the input.
We report the errors on the subset of the frames with predicted motion parameters.

\begin{figure}
\includegraphics[width=\linewidth]{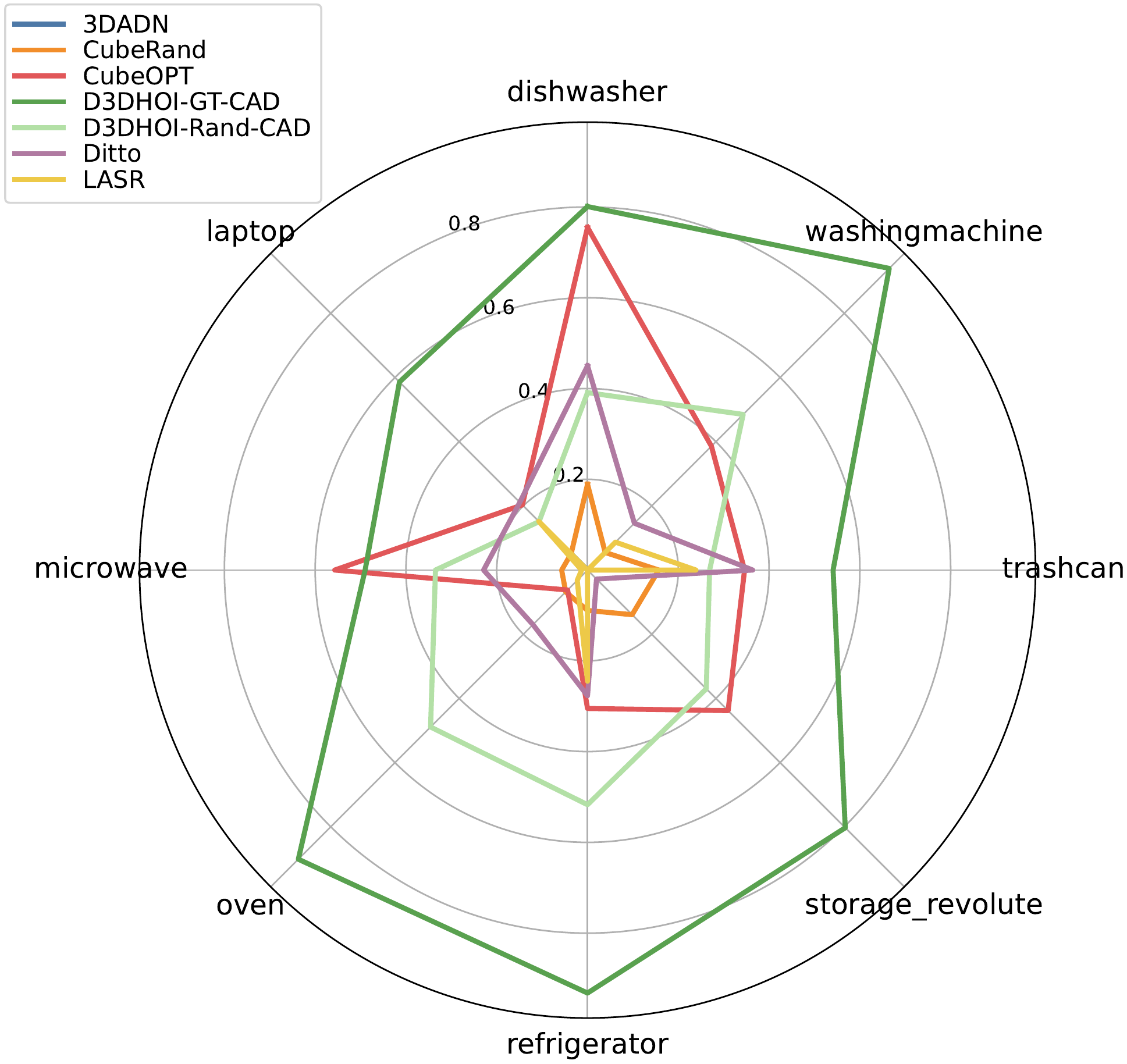}
\caption{Category-wise breakdown of the reconstruction accuracy metric (\accr) for all methods in our experiments.
The microwave, laptop, and trashcan are overall most challenging for most methods.
\internet is at zero in almost all categories except laptop and therefore not visible.}
\label{fig:cats}
\end{figure}

\mypara{How challenging are different object categories?}
\Cref{fig:cats} shows a category-wise breakdown of the reconstruction accuracy metric for each method.
Overall, laptops are most challenging (likely because the moving part is relatively small), while dishwashers are easiest (relatively large object and large moving part).
\cuboidopt works much better for microwaves than \dhoi (good fit with two-cuboid assumption).
Both methods do well on dishwashers, but less well on ovens (the bottom proofing drawer likely makes the articulation edge not as clear for ovens).

\mypara{What if we use predicted object masks for \cuboidopt?}
We also experiment with predicted object and part masks for \cuboidopt.    
Specifically, we compare using ground-truth (GT) vs predicted (Pred) object and part masks.
We use the predicted part segmentation from 3DADN to as the moving part mask.
For the object mask, we use Mask2Former~\cite{cheng2021maskformer} (Swin Large trained on COCO).
Since the pretrained Mask2Former on COCO does not include all the categories for D3DHOI, we only evaluate on 4 categories (laptop, microwave, oven, refrigerator) from the original dataset (170 videos).
We compare \cuboidopt-Pred with \internet for these 170 videos, and find that \cuboidopt achieves lower error and higher accuracy (see supplement).

\mypara{Discussion.}
Overall, the results indicate that this is a challenging problem with much space for improvement.
All methods, despite having access to GT information, have fairly high motion axis and direction errors and motion state errors.
Some predictions are very good, but there are many poor results even within a single video input (as is indicated by the relatively high standard error margins on the reported mean error values in \Cref{tab:results-err-fullset}).

While fairly simple and limited in terms of object shapes it can model \cuboidopt reconstructs the object without needing a CAD model dataset.
The built-in bias towards cuboidal part articulated objects helps it outperform \internet which similarly does not rely on a database of 3D CAD models, unlikely \dhoi.

\section{Conclusion}

In this paper we defined a canonicalized articulated 3D human-object interaction task: single-view reconstruction of objects from RGB videos of humans interacting with an articulated object.
We carried out a systematic benchmark of five classes of methods for this task.
Using a set of metrics to evaluate the predicted object reconstruction, pose, and motion we saw that all methods struggle to obtain high quality results even when provided with privileged information such as ground truth object masks and CAD models.
We see a number of opportunities for future work on improved methods for articulated 3D human-object interaction.

\mypara{Acknowledgements:}
This work was funded in part by a Canada CIFAR AI Chair, a Canada Research Chair and NSERC Discovery Grant, and enabled in part by support from \href{https://www.westgrid.ca/}{WestGrid} and \href{https://www.computecanada.ca/}{Compute Canada}.
We thank Xiang Xu, Shengyi Qian and Zhenyu Jiang for answering questions about D3DHOI, 3DADN and Ditto respectively.

{\small
\bibliographystyle{plainnat}
\setlength{\bibsep}{0pt}
\bibliography{main}
}

\clearpage
\appendix

This supplemental document provides the following additional content to support the main paper:

\hspace{1pt}\ref{sec:implementation_details} : implementation details for the \cuboidopt method.

\hspace{1pt}\ref{sec:ditto_details} : details of method based on \ditto.

\hspace{1pt}\ref{sec:additional_qual} : additional qualitative result examples.

\hspace{1pt}\ref{sec:additional_quant} : additional quantitative evaluations and ablations.

\section{\cuboidopt implementation details}
\label{sec:implementation_details}

\begin{figure}
\includegraphics[width=\linewidth]{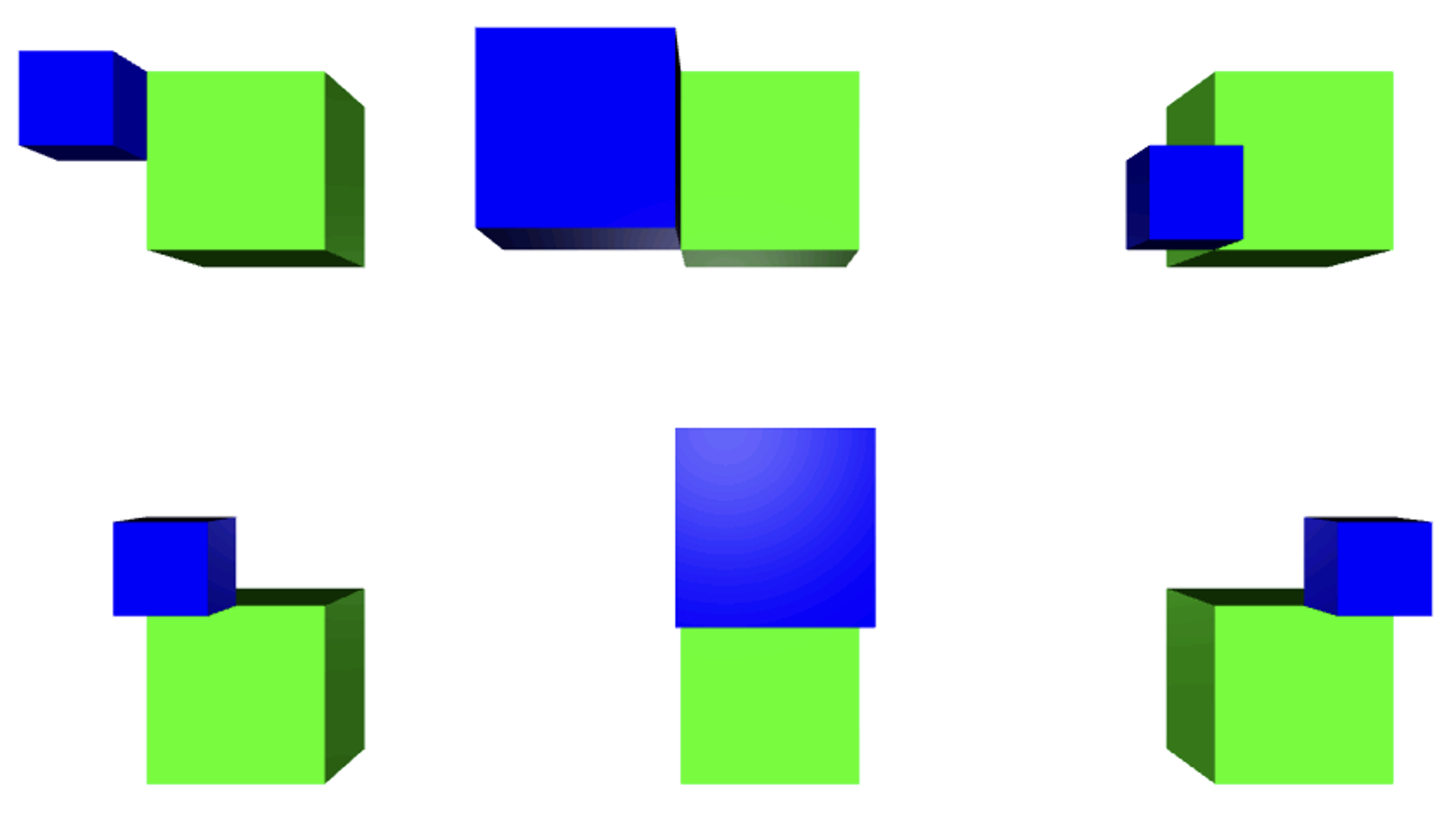}
\caption{Example \cuboidopt initialization templates for the left edge (top row) and for the top edge (bottom row).
For each edge, the moving part (blue cuboid) is attached at half length constrained to touch one end of the edge, full length spanning the edge, or half length touching the other end of the edge.}
\label{fig:inits}
\end{figure}

\mypara{Initializations.}
Estimating a cuboid abstraction from 2D images is an underconstrained problem.
Thus, we design a set of ``templates'' that serve as initializations.
We first initialize the base part to be in the center of the image.
Then for each of up to four edges that are visible from the camera, we attach the moving part to the edge in three possible configurations.
The moving part is scaled to be: i) the full length of the edge; or ii) half the length and starting from one end of the edge; or iii) half the length and starting from the other end.
\Cref{fig:inits} shows example templates for two edges.
This scheme leads to a total of 12 possible templates that are used as initializations.
We run \cuboidopt over all these initializations and select the best result.

\mypara{Objective Functions.}
We estimate the rotation, translation, scale and articulation parameters by optimizing an objective function which is a sum of a silhouette loss ($L_\text{sil}$) and a Dice loss ($L_\text{dice}$) for ensuring the ground truth and projected 3D object masks are the same, an overlap loss ($L_\text{over}$) to penalize interpenetration of the cuboids, and a human-object interaction loss ($L_\text{hoi}$).

\textbf{Silhouette loss $L_\text{sil}$:} penalizes the discrepancy between the ground truth object mask, $M_t^\text{gt}$ and the projected 3D object mask $M_t^\text{proj}$.
We project the object into a 2D image mask at each frame time $t$ to compute the loss as a mean of the discrepancies,
\begin{equation}
    L_\text{sil}(M^\text{gt}, M^\text{proj}) = \frac{1}{N} \sum_{t=1}^{N} ||M_t^\text{gt} - M_t^\text{proj}||^2_2
\end{equation}

We apply this loss over the full object, base part and moving part separately i.e. the loss is defined as,
\begin{equation}
    L_\text{sil} =  L_\text{sil}^\text{obj} + L_\text{sil}^\text{base} + L_\text{sil}^\text{move}
\end{equation}

\textbf{Dice loss $L_\text{dice}$:} maximizes intersection-over-union (IoU) between ground truth object mask $M_t^\text{gt}$ and projected 3D object mask $M_t^\text{proj}$.
The loss is computed over all frames as,
\begin{equation}
    L_\text{dice}(M^\text{gt}, M^\text{proj}) = \frac{1}{N} \sum_{t=1}^{N} 1 - \text{IoU}(M_t^\text{gt}, M_t^\text{proj})
\end{equation}

Similarly to the silhouette loss, we also apply the Dice loss over full object, base part and moving part separately,
\begin{equation}
    L_\text{dice} =  L_\text{dice}^\text{obj} + L_\text{dice}^\text{base} + L_\text{dice}^\text{move}
\end{equation}

\textbf{Overlap loss $L_\text{over}$:} penalizes overlap between the two cuboids $C_b$ and $C_m$.
This is used to discourage degenerate solutions where both the cuboids have significant overlap to the point where only one cuboid is visible in the projection.
We adopt a mesh-mesh overlap term from prior work to detect colliding mesh triangles and penalize the depth of the penetration of the collisions~\cite{karras2012maximizing, tzionas2016capturing}.

\textbf{Human-object interaction loss $L_\text{hoi}$:}
consists of a ``depth loss'' to encourage the depth of the human and the object to be similar, and a ``contact curve loss'' to encourage the moving part to move along with the human contact point.
The former is implemented by minimizing the difference between the average $z$-component of the human and object mesh vertices.
The latter is adapted from \citet{xu2021d3d}'s contact curve loss to constrain the moving part of the object to follow a similar curve as the human hand.

The $L_\text{hoi}$ loss is composed of two terms: $L_\text{depth}$ and $L_\text{contact}$.

For the $L_\text{depth}$ term, since we know that the human and object are always in front of the camera, we use the $z$ coordinates of the human and object mesh vertices as a proxy for depth from the camera.
We define this term as the difference between the mean $z$ coordinate of the vertices of the human and object meshes.
$$
L_\text{depth} = \max(0, |\frac{1}{N}\sum_i^N H_i^z - \frac{j}{M} \sum_j^M S_j^z| - \lambda)
$$
where $H_i^z$ is the $z$-component of the $i$-th vertex of the human mesh and similarly, $S_j^z$ is the $z$-component of the $j$-th vertex of the object mesh and $\lambda$ is a threshold parameter which we set to $0.1$ in our experiments.

For the $L_\text{contact}$ term, we adapt the contact curve loss from~\citet{xu2021d3d} which constrains the moving part of the object to follow a similar curve as the human hand.
We select the centroid of the moving part at each timestep $t$, $v_c(t)$, as our contact point on the object and the centroid of a hand of the human at each timestep, $h_c(t)$, as the contact point on the human.
Note that the human pose estimates are ``jittery'' and therefore it is difficult to get a consistent contact point estimate.
This is why we resort to the centroid of the moving part as our contact point to mitigate some of the pose estimation noise.
Also, we are more concerned with the trajectory of the motion than accurate contact point estimates.
Following \citet{xu2021d3d}, we allow for a rigid transformation of $v_c(t)$ so that the contact loss focuses only on the shape of the curve.
The contact curve loss is then defined as,
$$
L_\text{contact} = \sum_{t=1}^{t=N} (R_c v_c(t) + T_c - h_c(t))^2
$$
Here, $R_c$ and $T_c$ are rotation and translation parameters that we optimize for the contact curve.

\textbf{Final objective function:} a sum of all the loss terms described above,
\begin{equation}
    L_\text{final} = L_\text{sil} + L_\text{dice} + L_\text{over} + L_\text{hoi} %
\label{eq:final}
\end{equation}

\begin{figure*}
\setkeys{Gin}{width=\linewidth}
\footnotesize
\begin{tabularx}{\textwidth}{@{} Y Y Y Y Y Y @{}}
GT & \internet & \cuboidopt & \dhoi-GT-CAD & \ditto & \lasr \\

\imgclip{0}{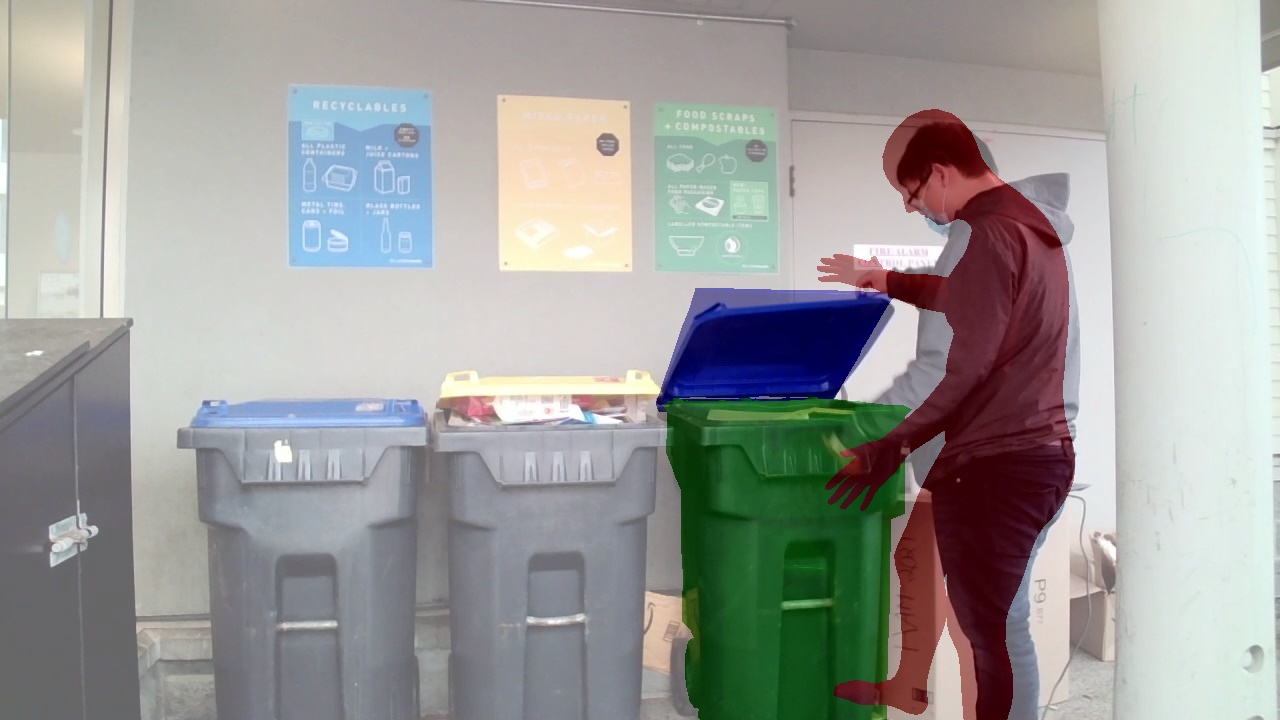} & 
\imgclip{0}{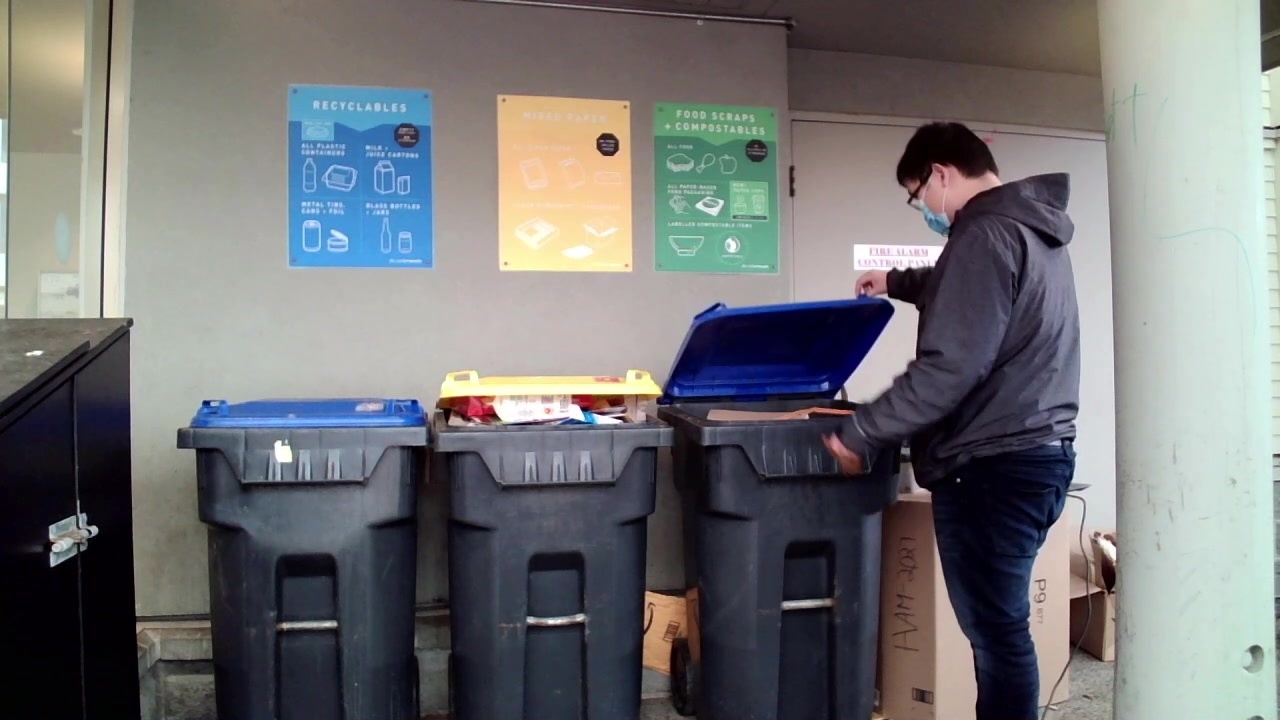} & 
\imgclip{0}{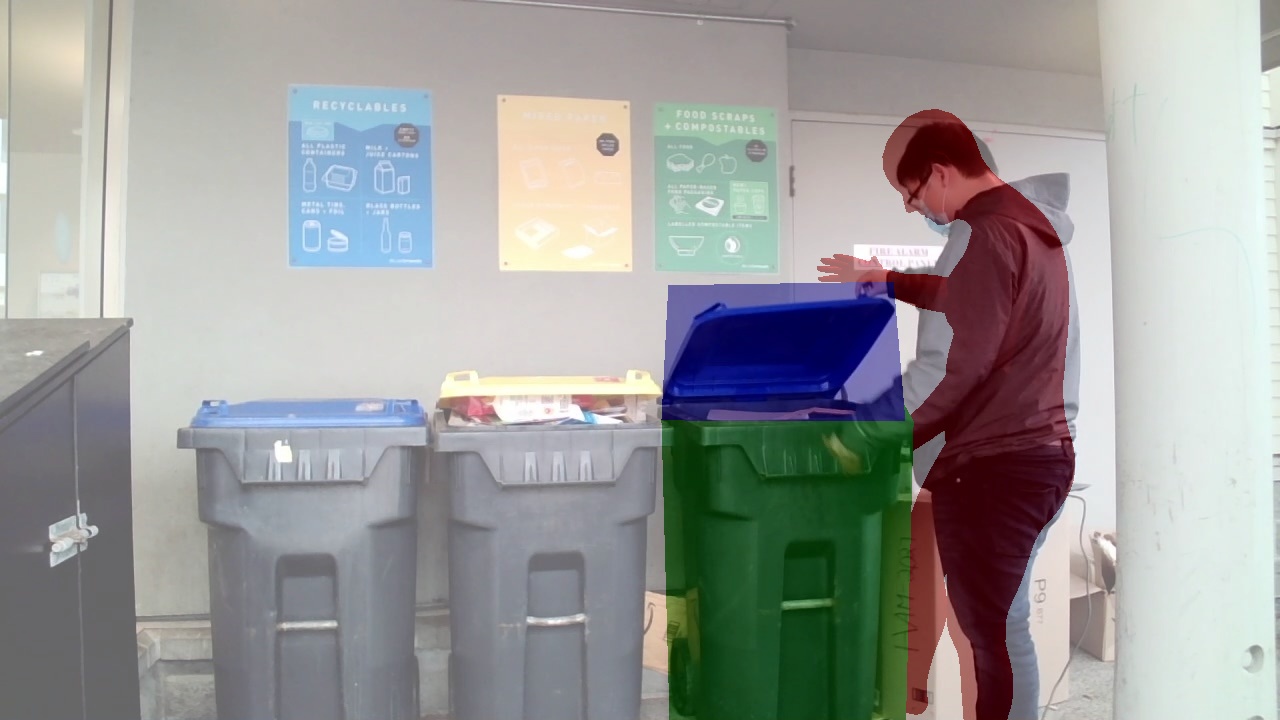} & 
\imgclip{0}{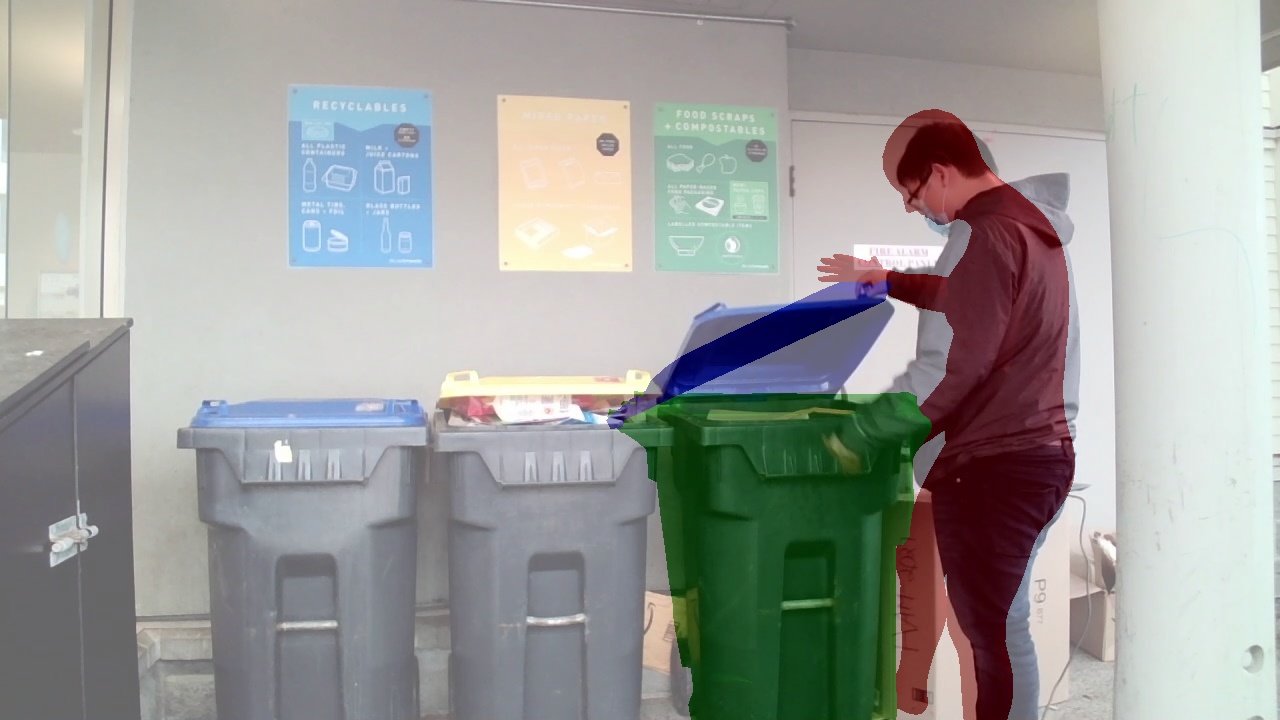}  &
\imgclip{0}{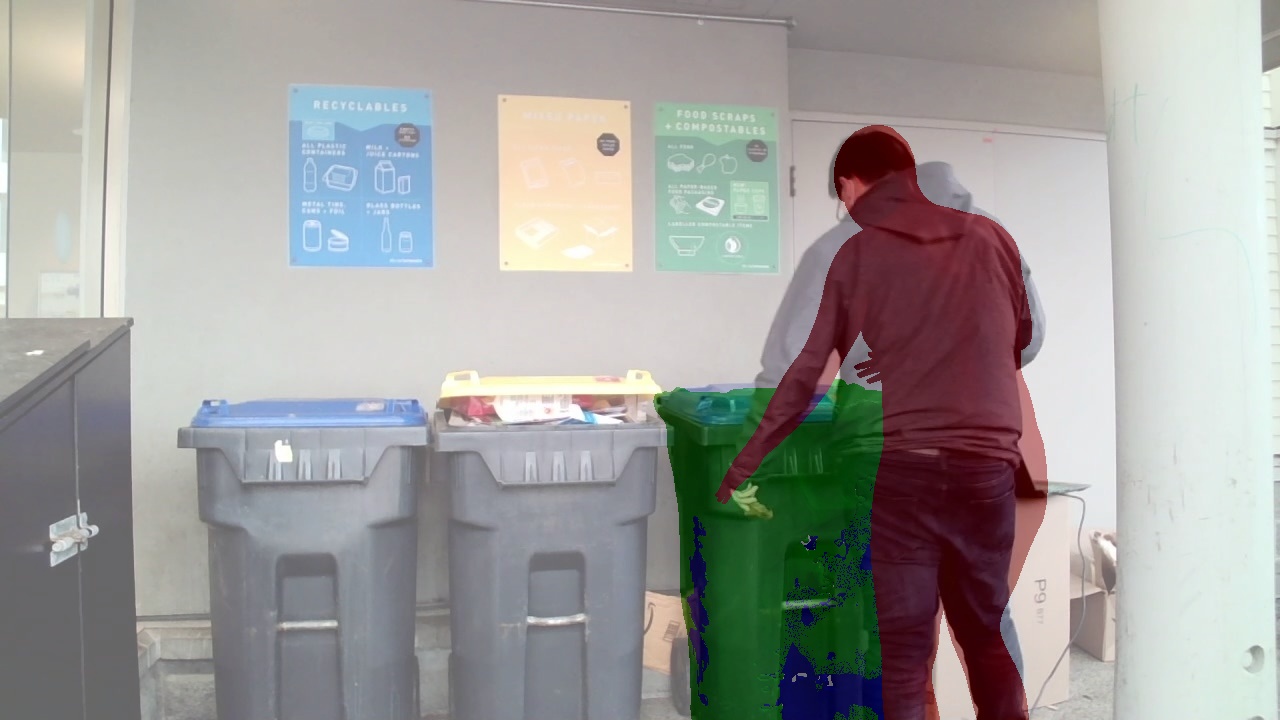} & 
\imgclip{0}{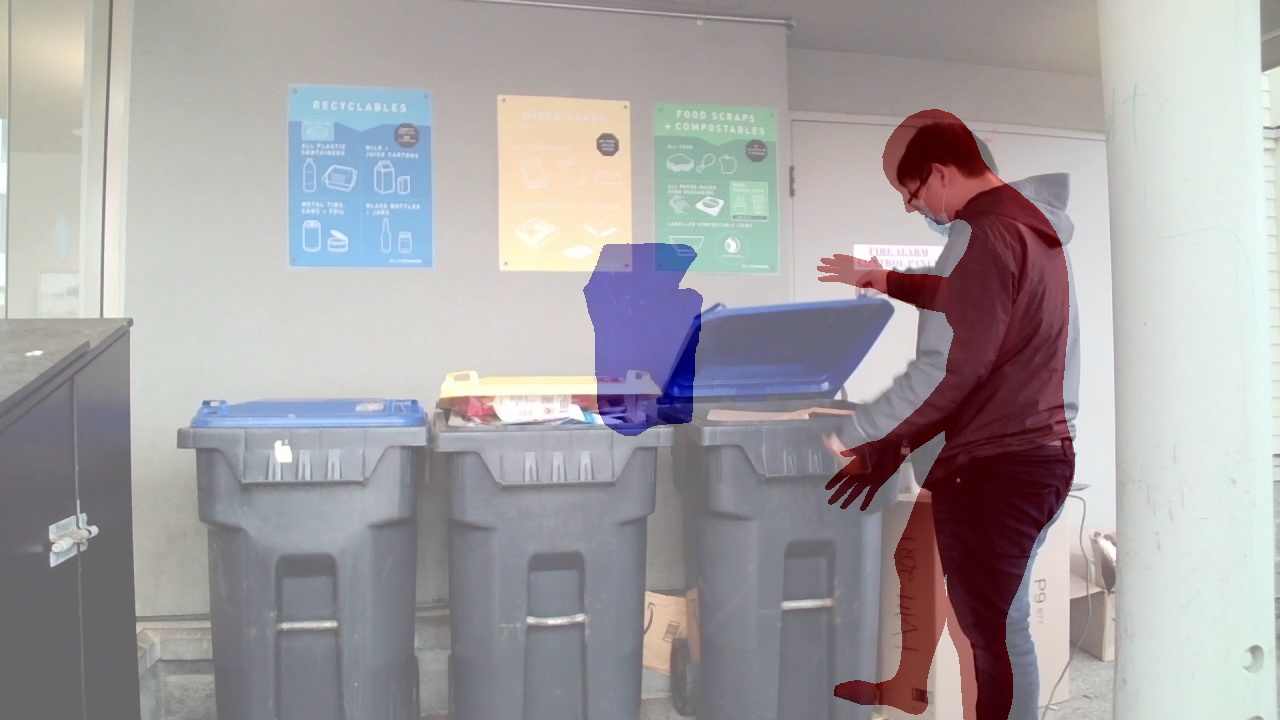}\\

\imgclip{0}{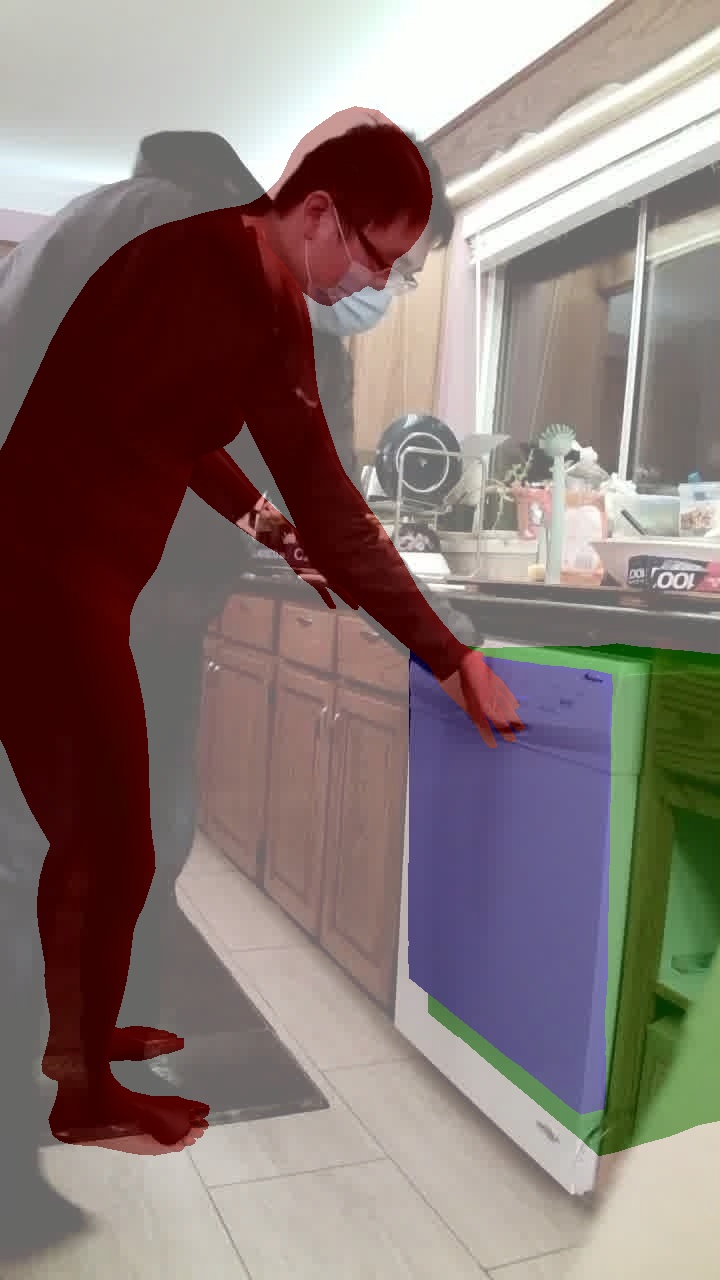} & 
\imgclip{0}{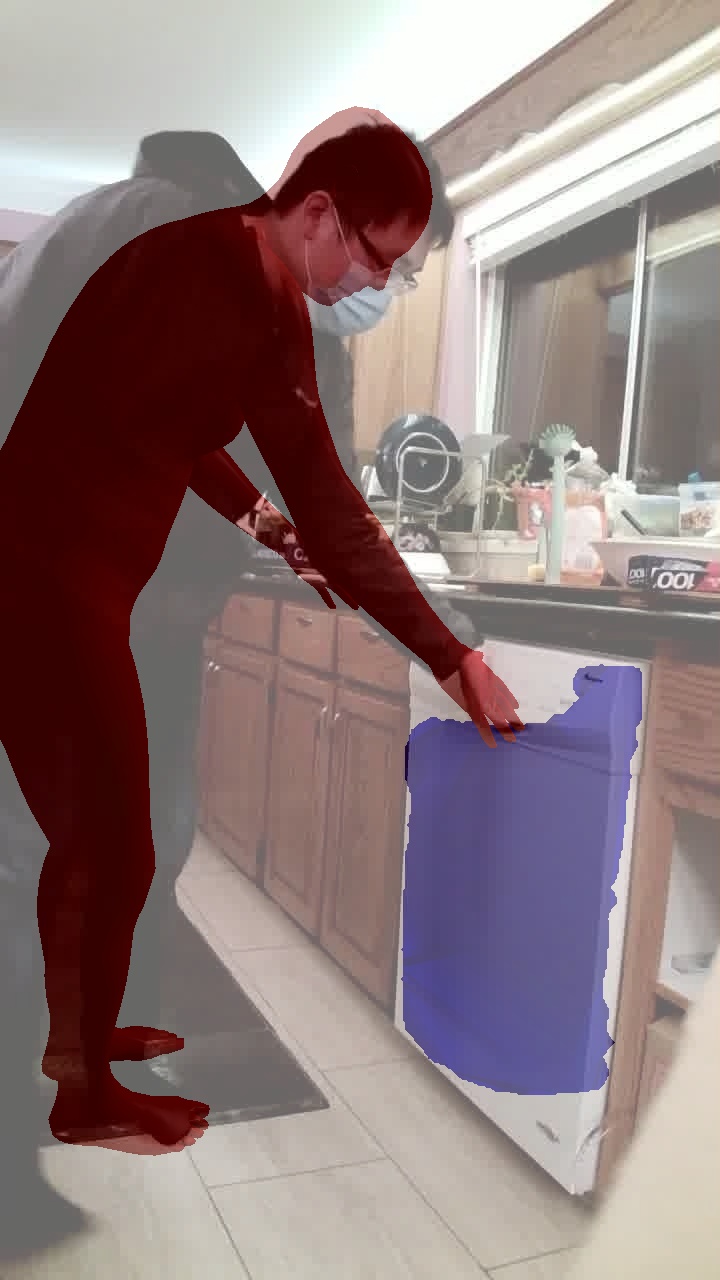} & 
\imgclip{0}{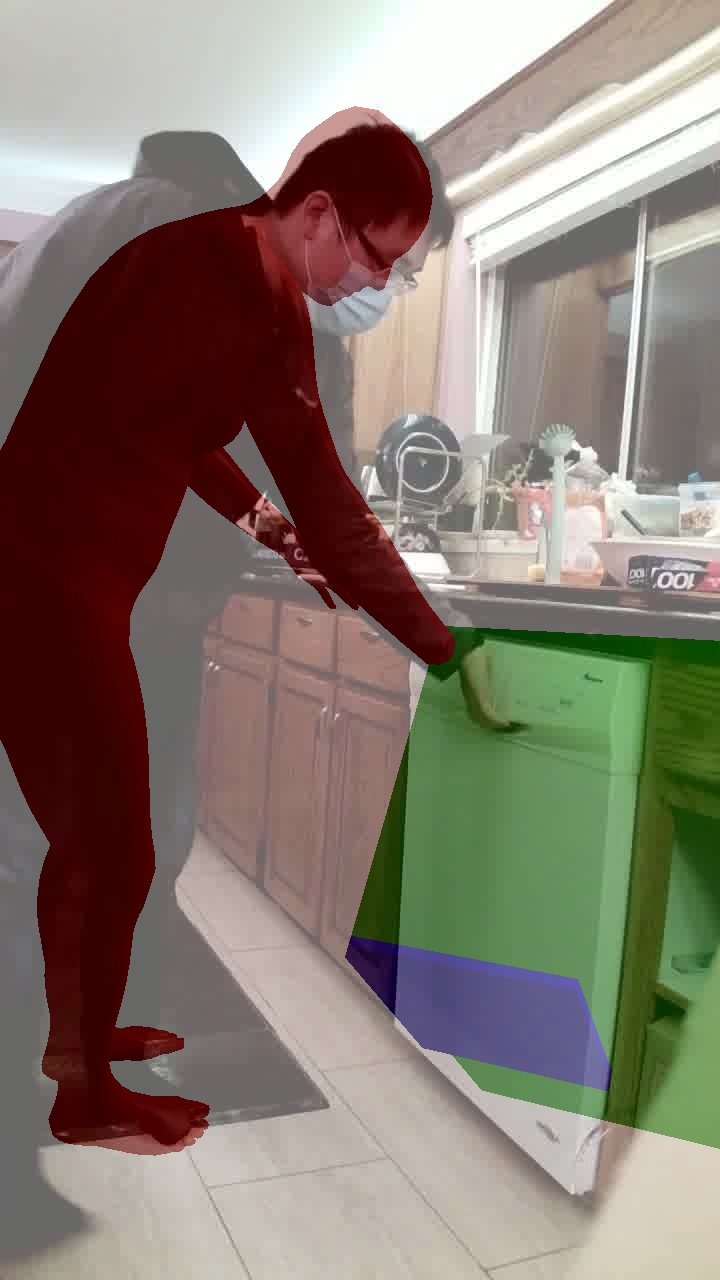} & 
\imgclip{0}{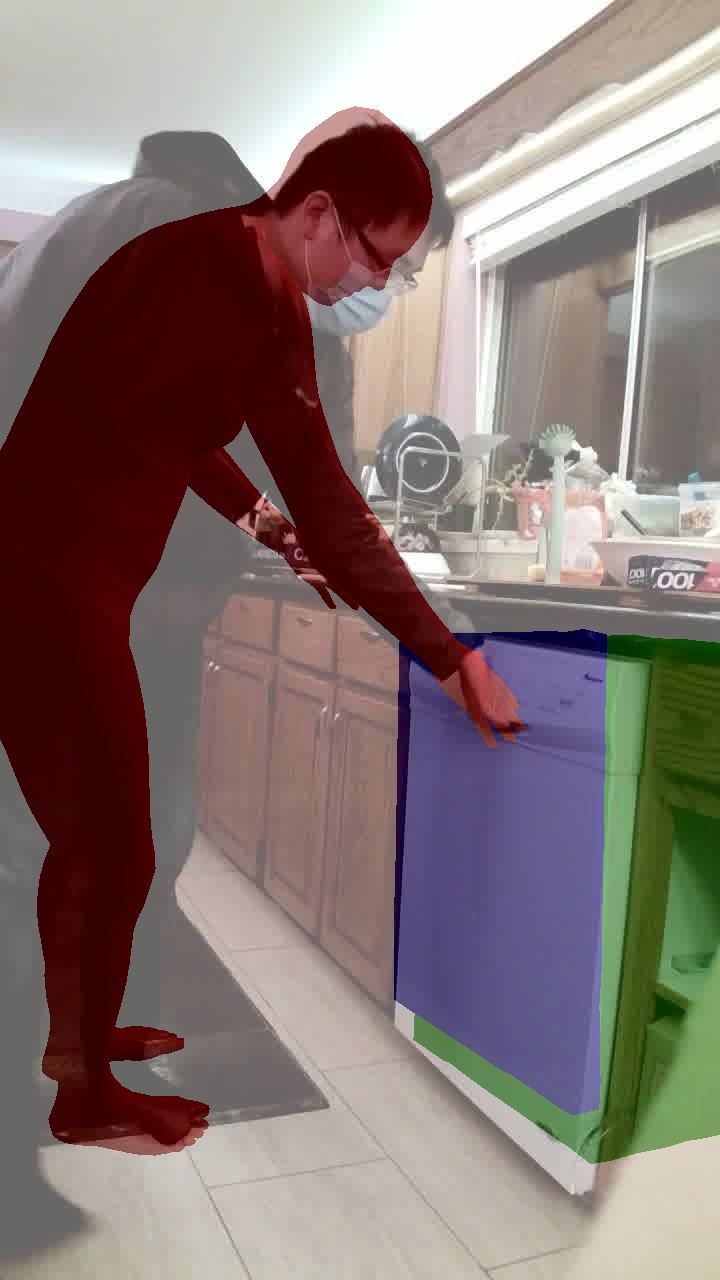}  &
\imgclip{0}{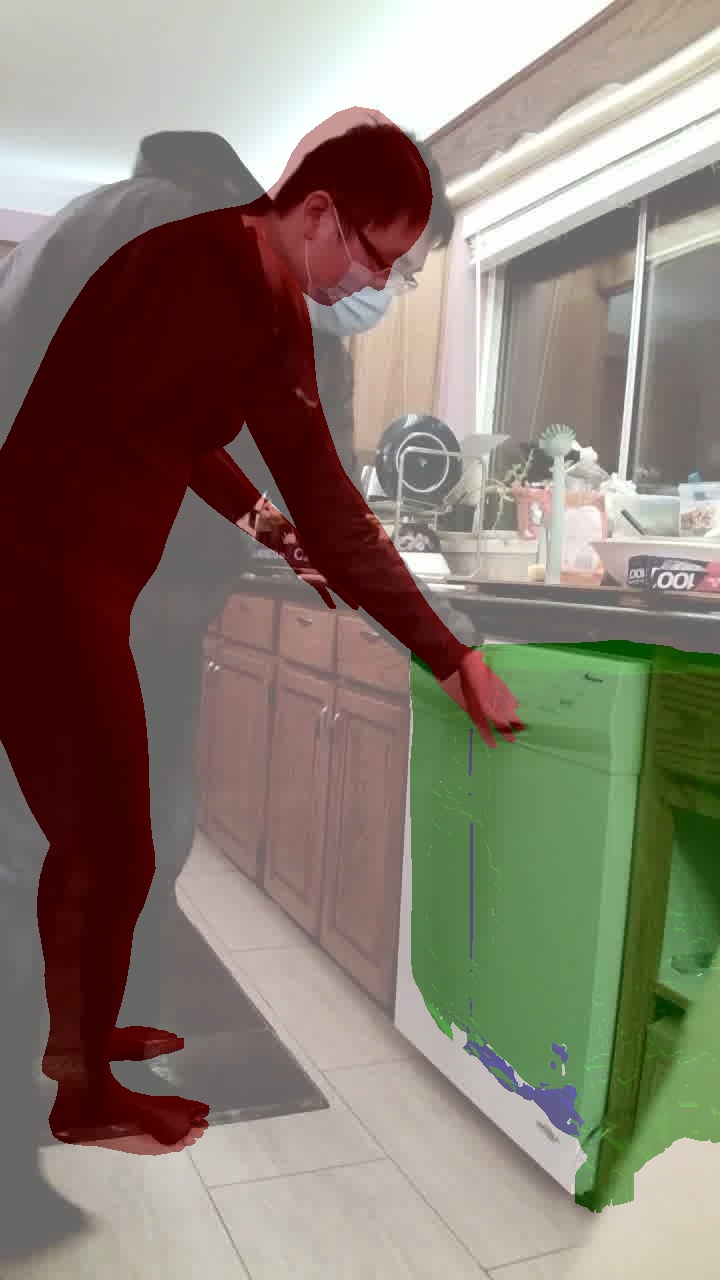} & 
\imgclip{0}{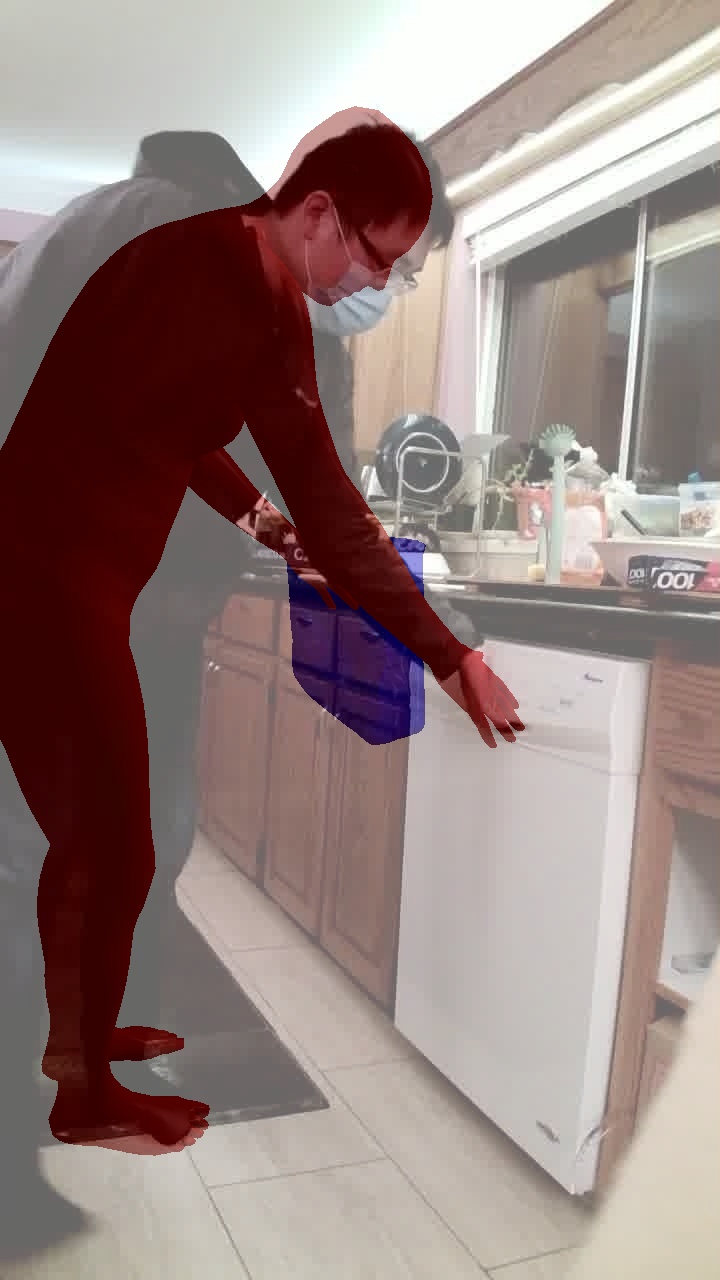}\\

\imgclip{0}{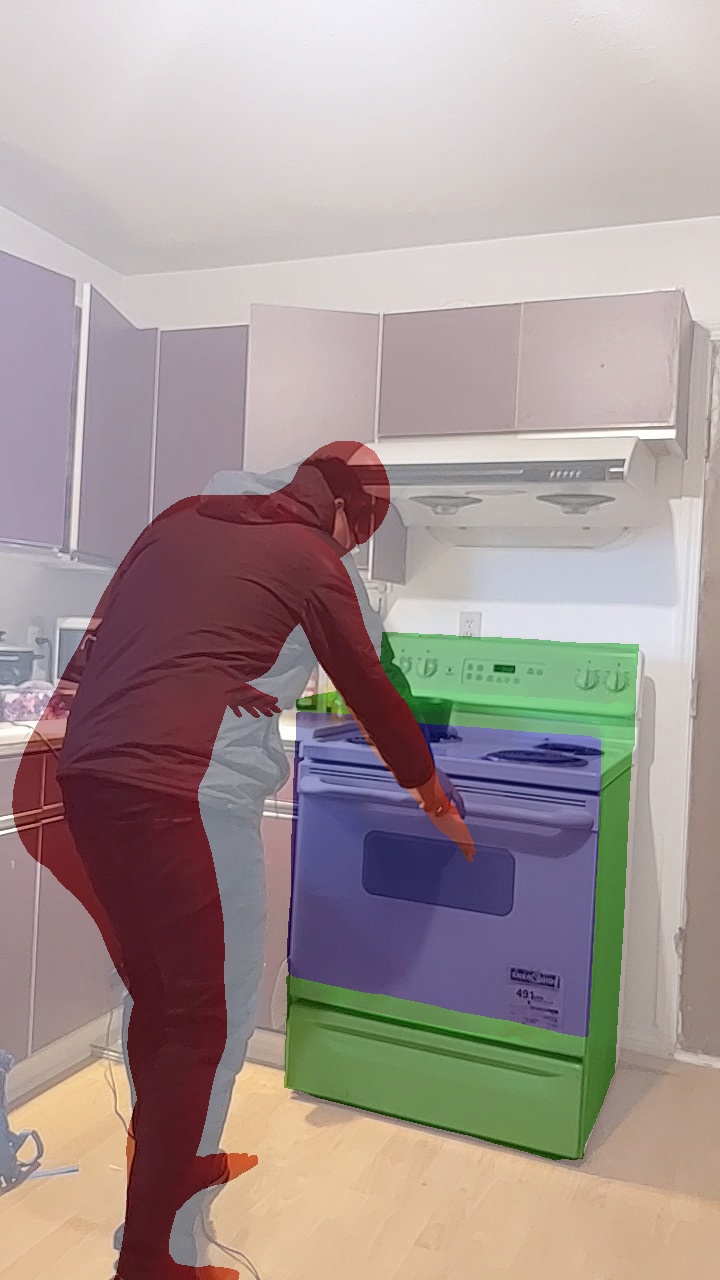} & 
\imgclip{0}{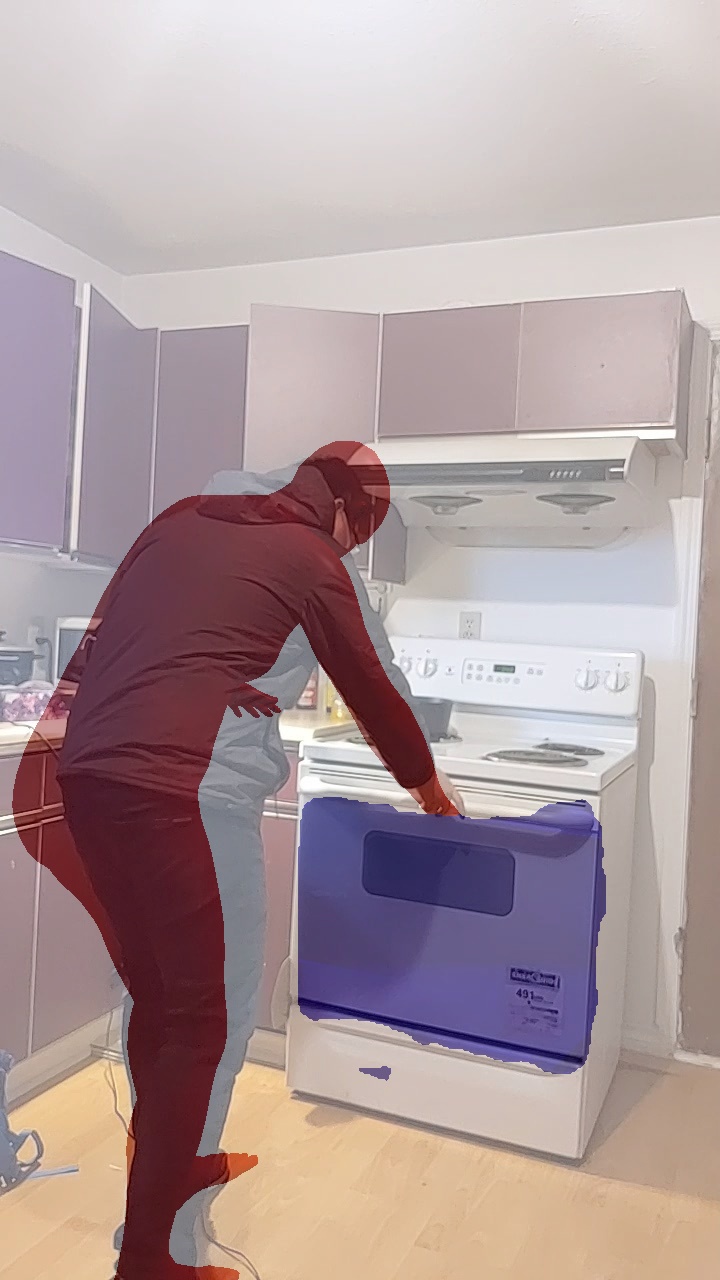} & 
\imgclip{0}{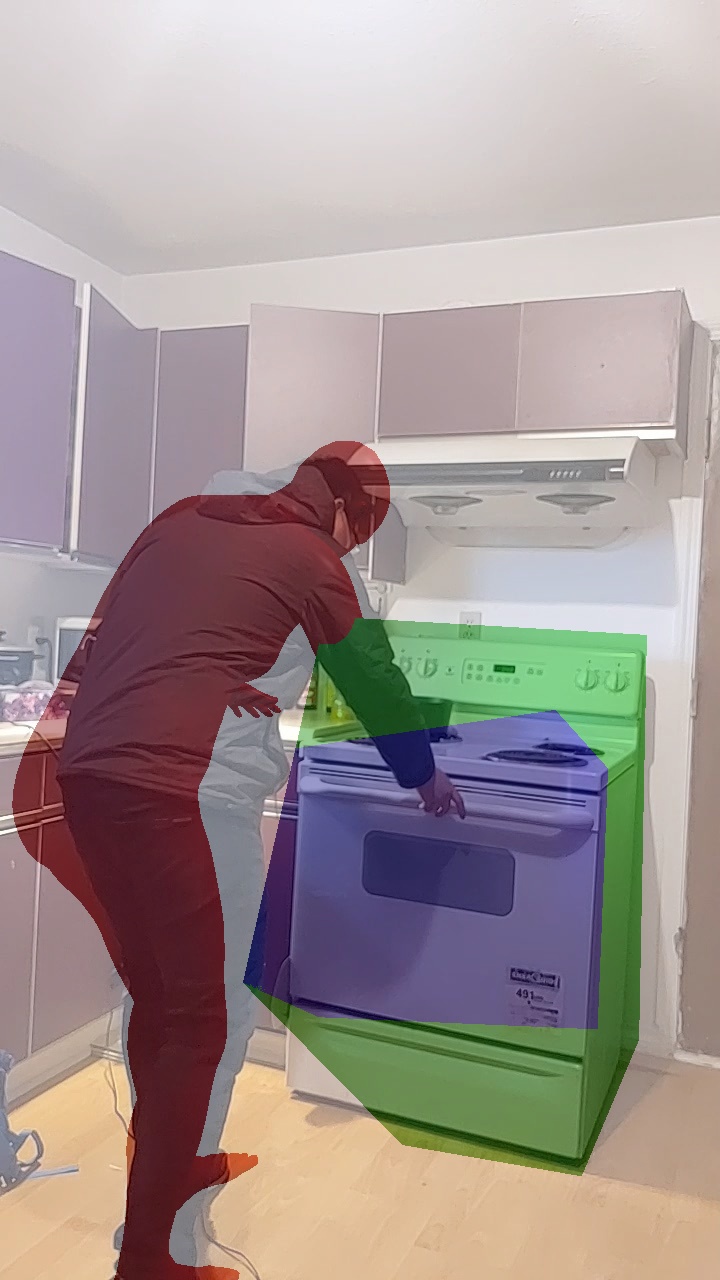} & 
\imgclip{0}{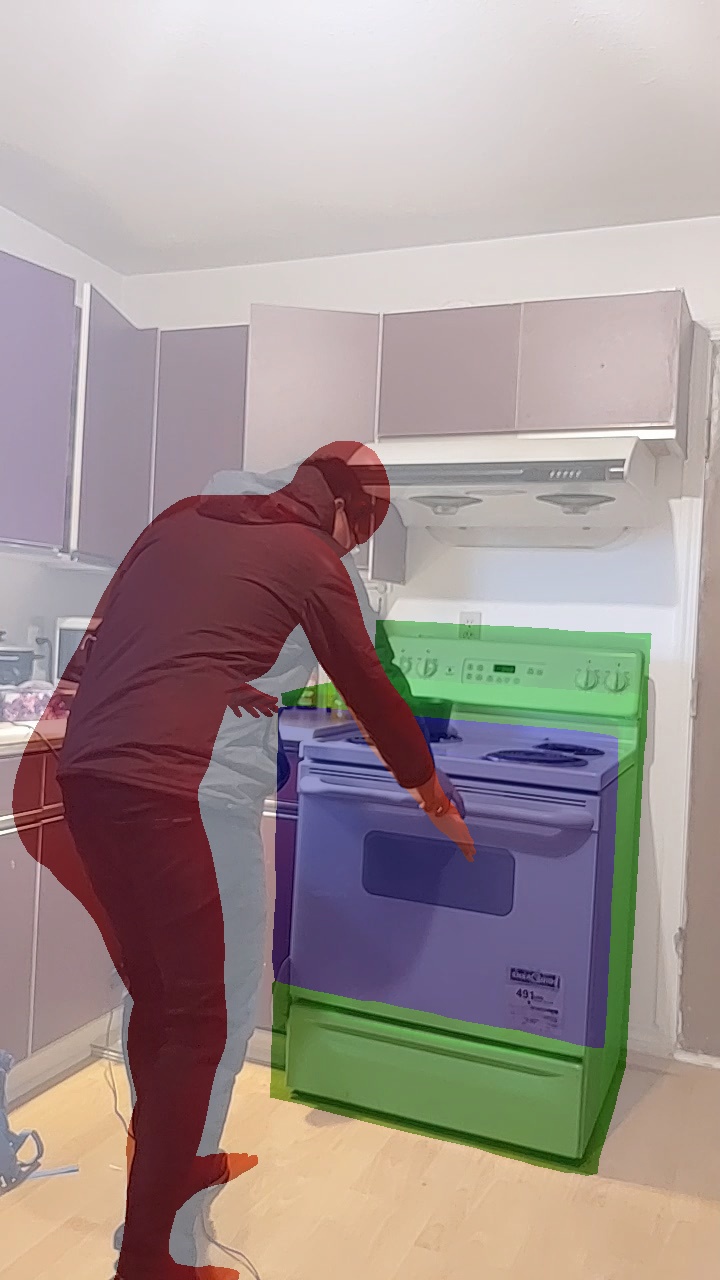}  &
\imgclip{0}{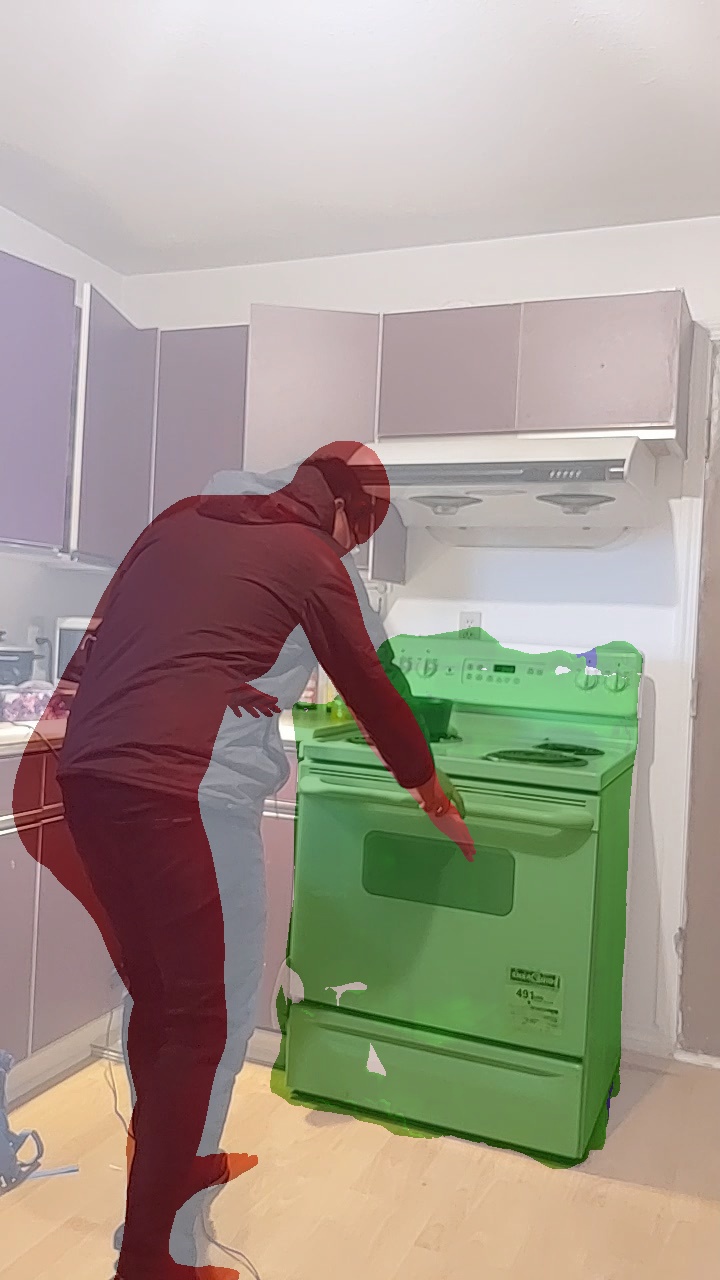} & 
\imgclip{0}{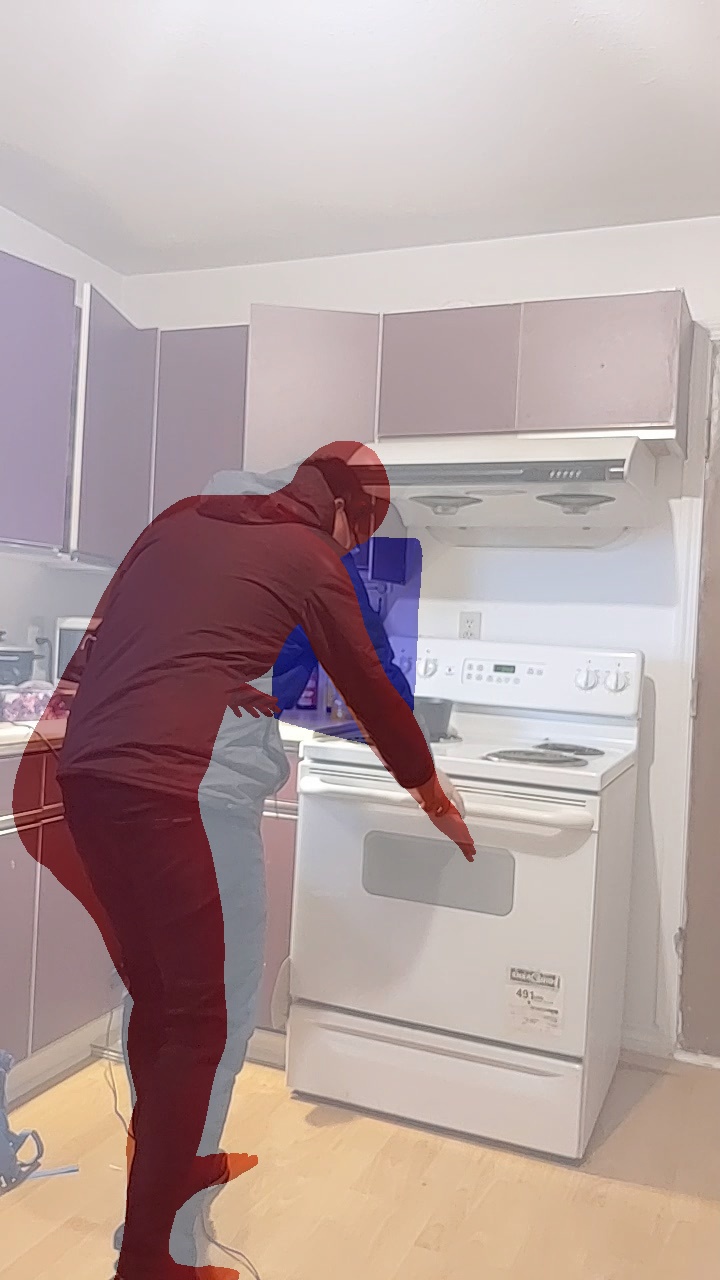}\\

\imgclip{0}{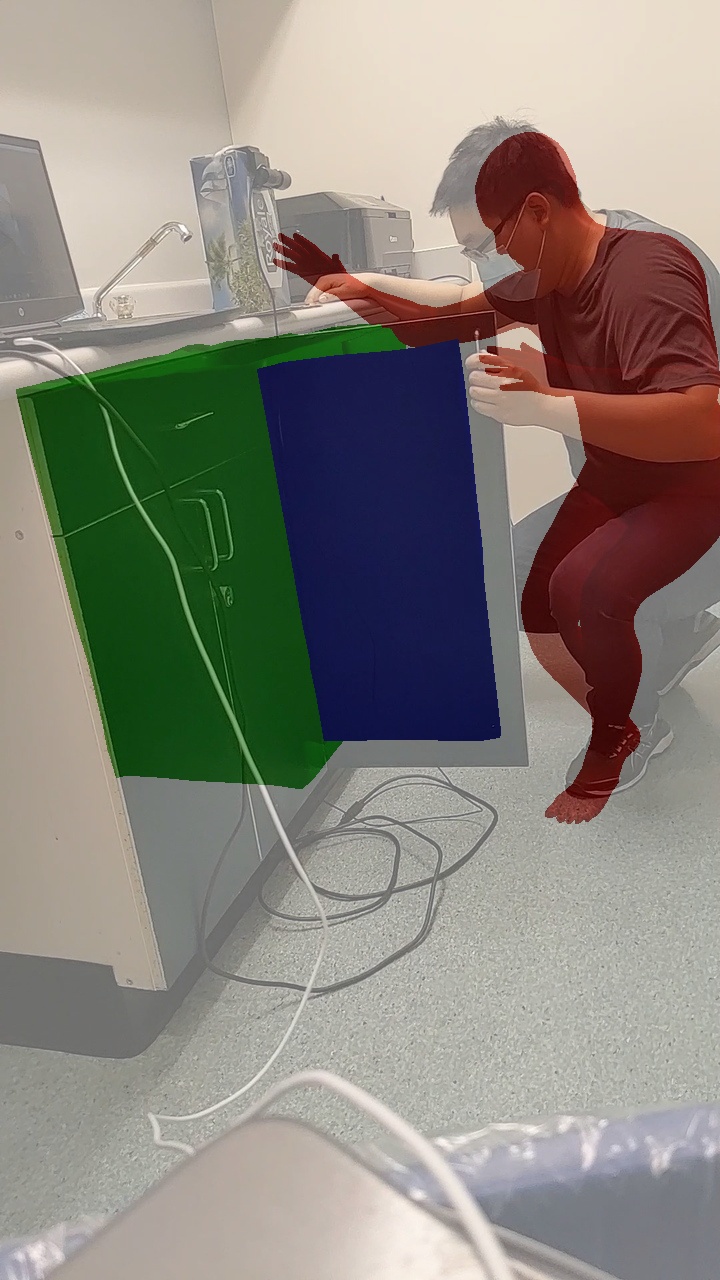} & 
\imgclip{0}{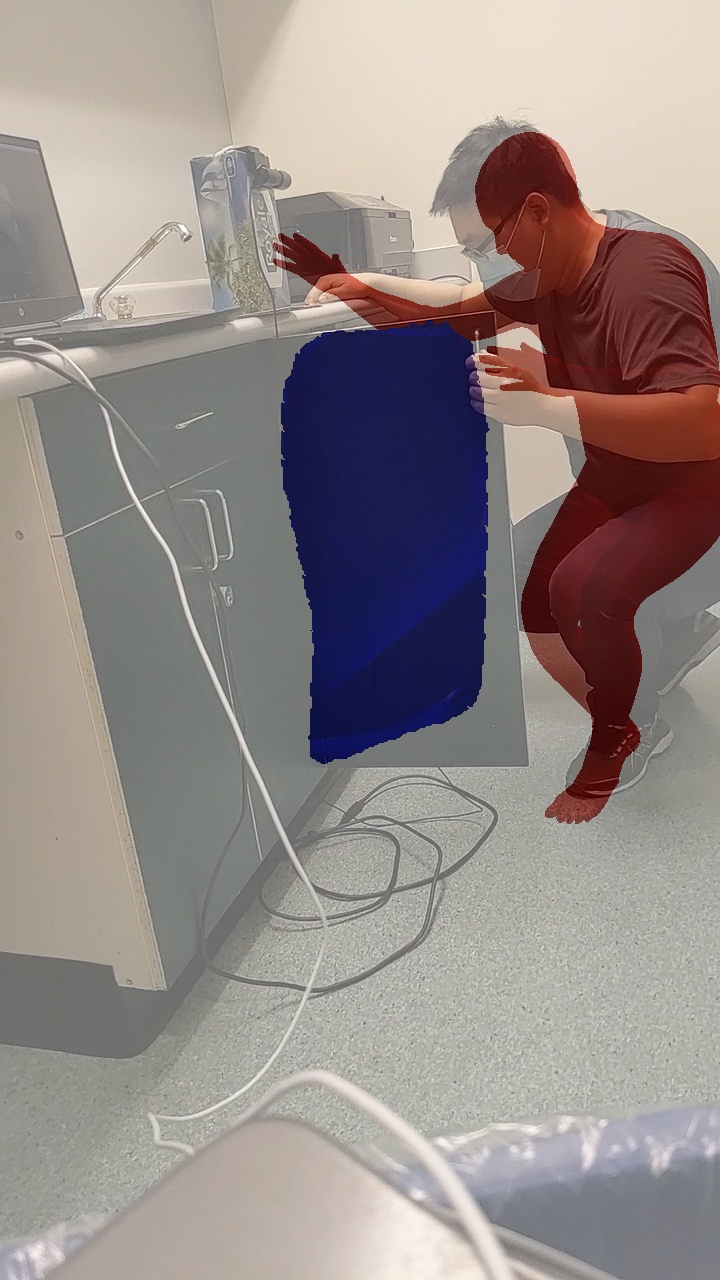} & 
\imgclip{0}{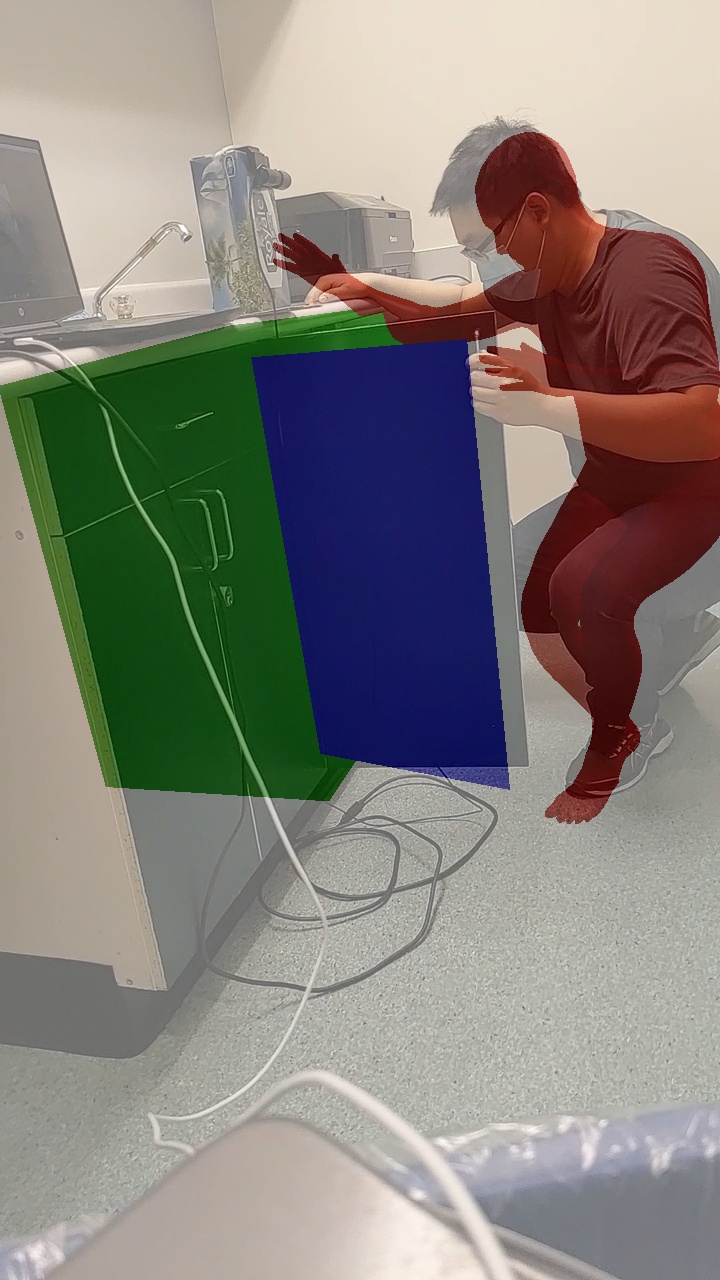} & 
\imgclip{0}{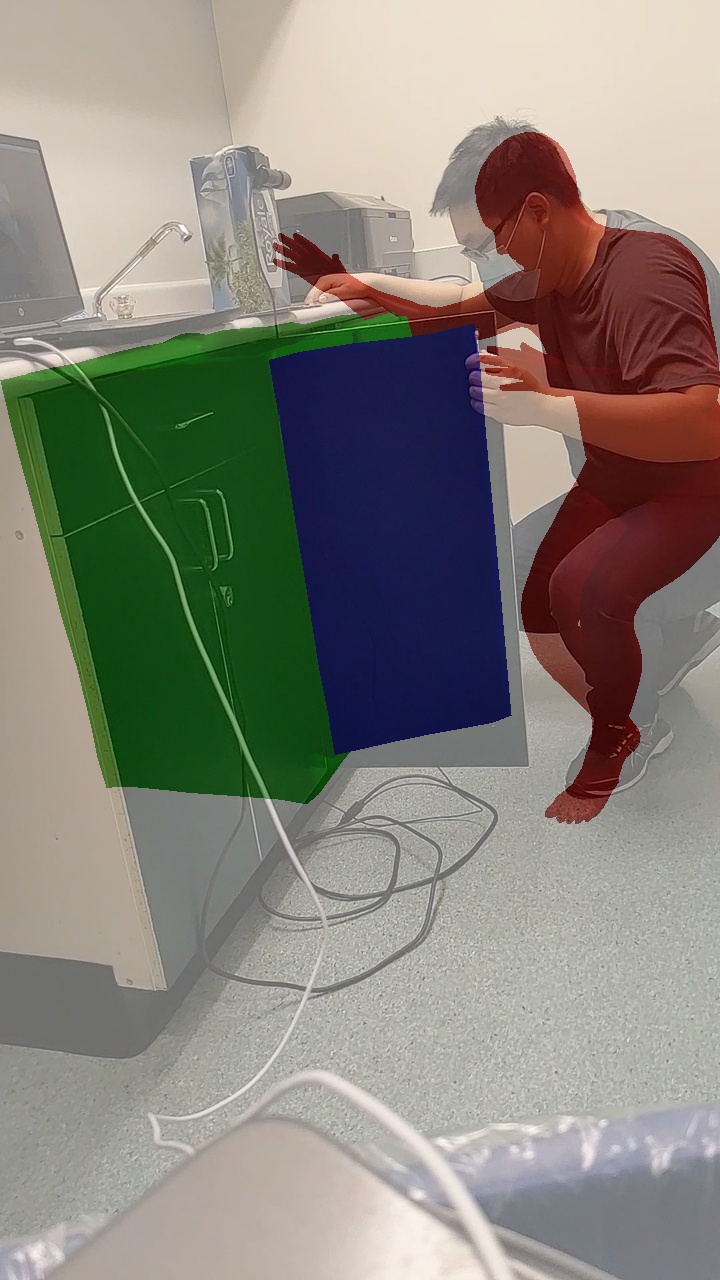} & 
\imgclip{0}{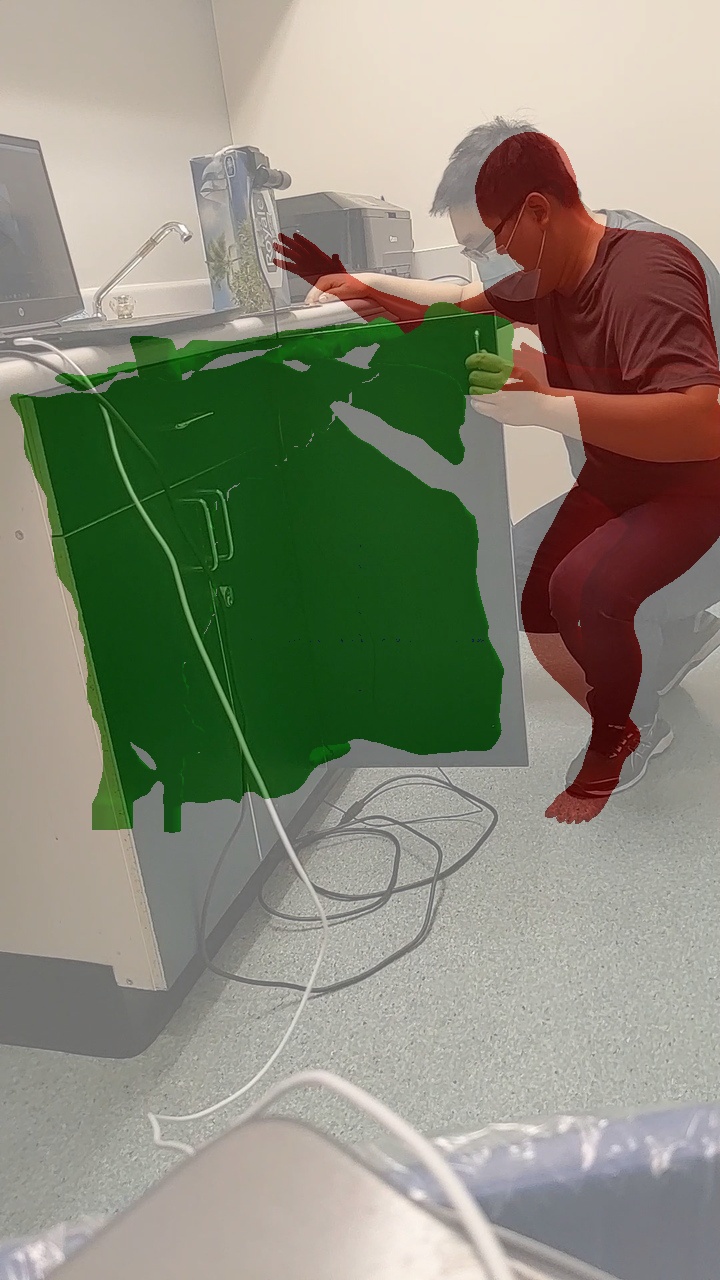} & 
\imgclip{0}{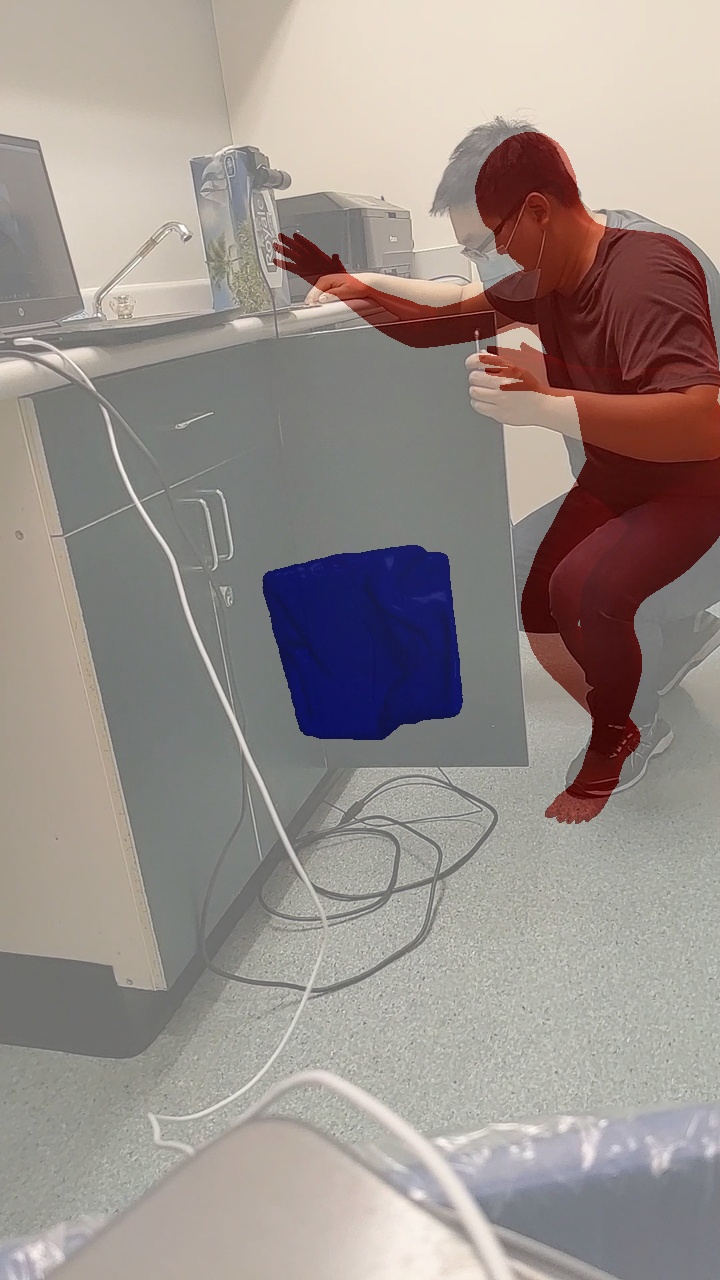}\\

\imgclip{0}{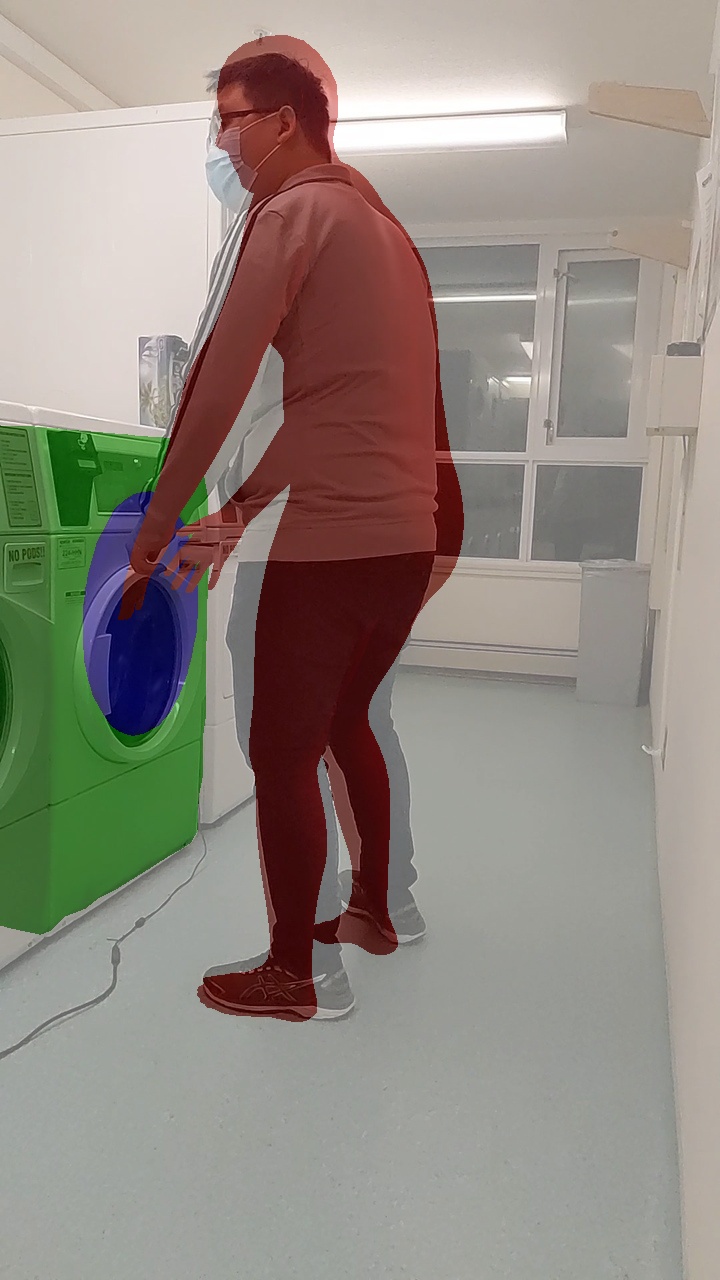} & 
\imgclip{0}{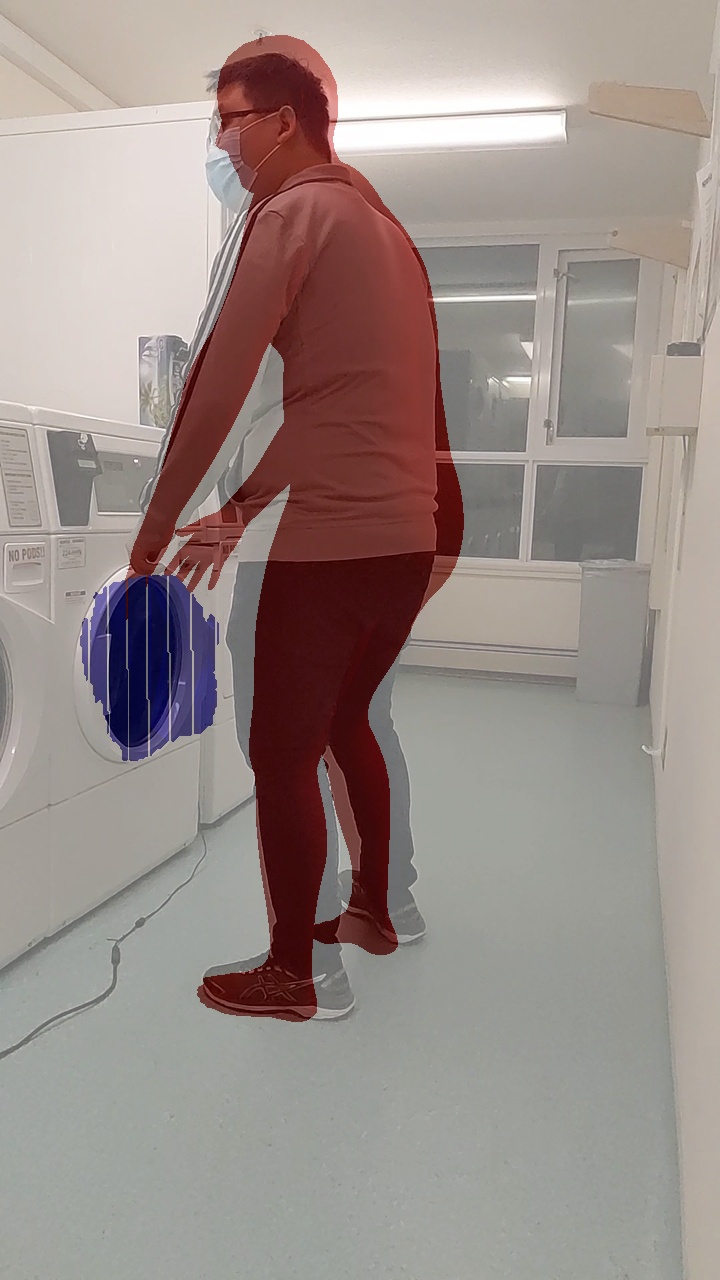} & 
\imgclip{0}{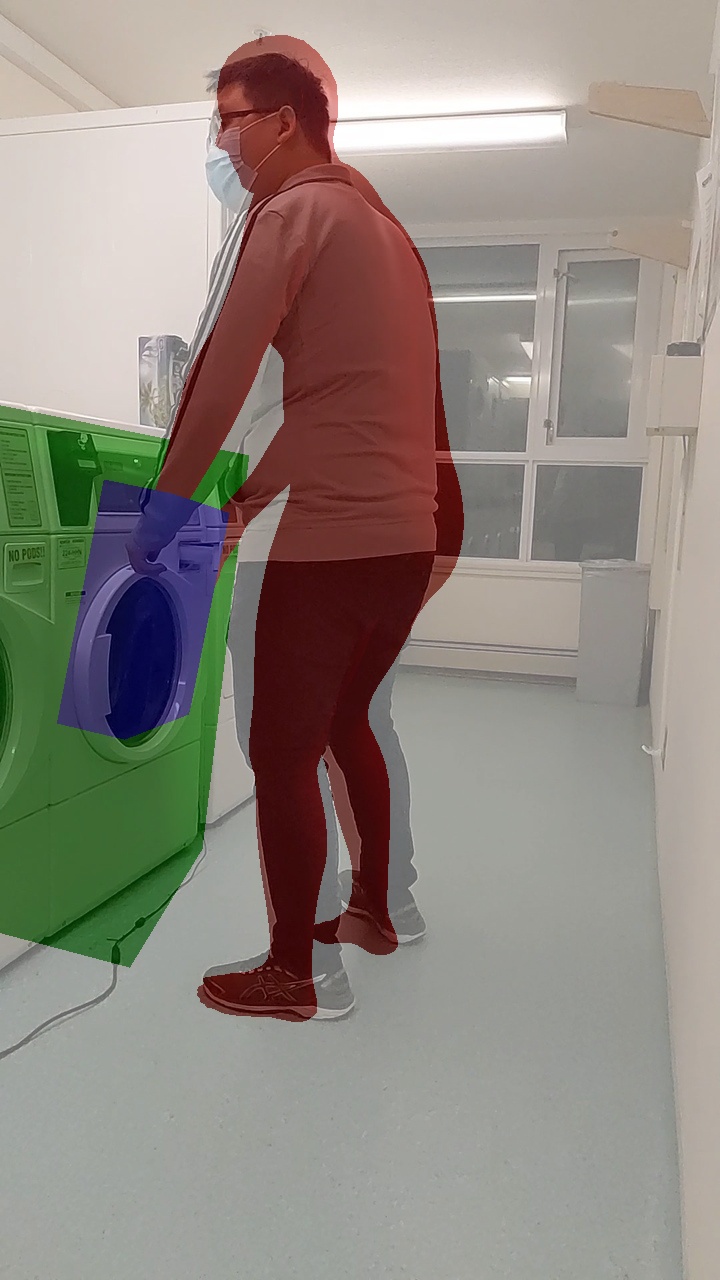} & 
\imgclip{0}{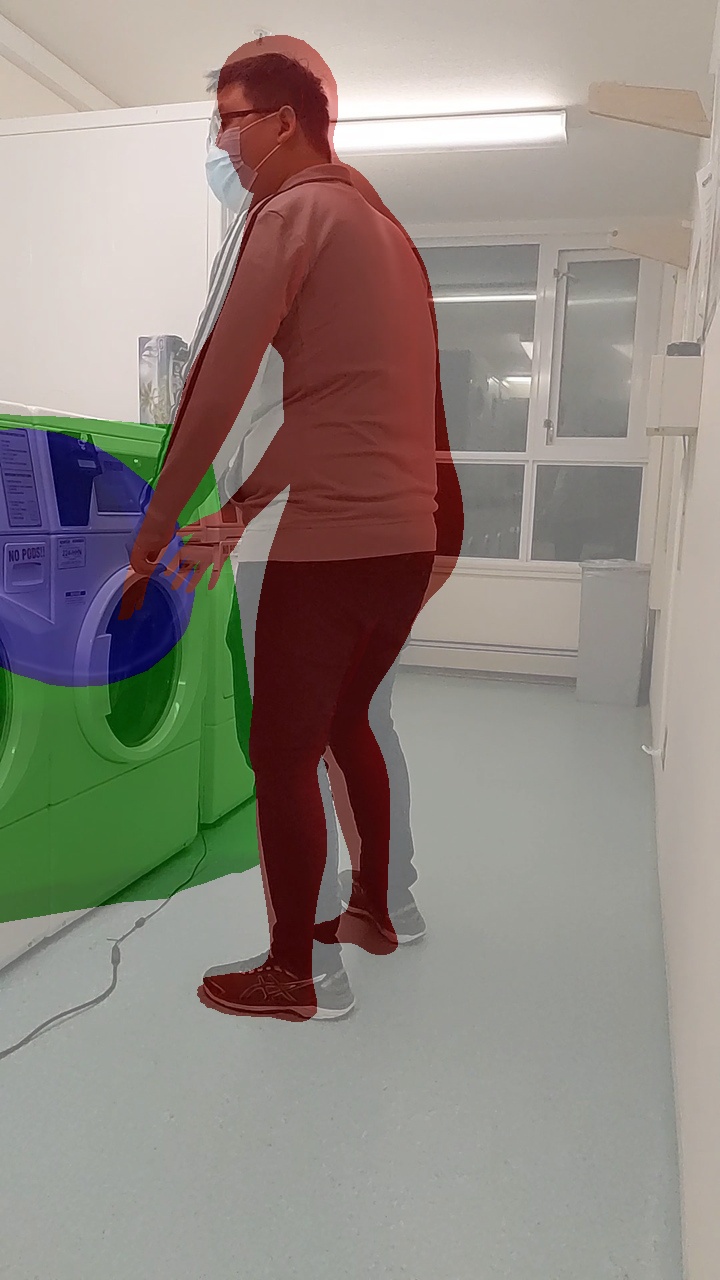} &
\imgclip{0}{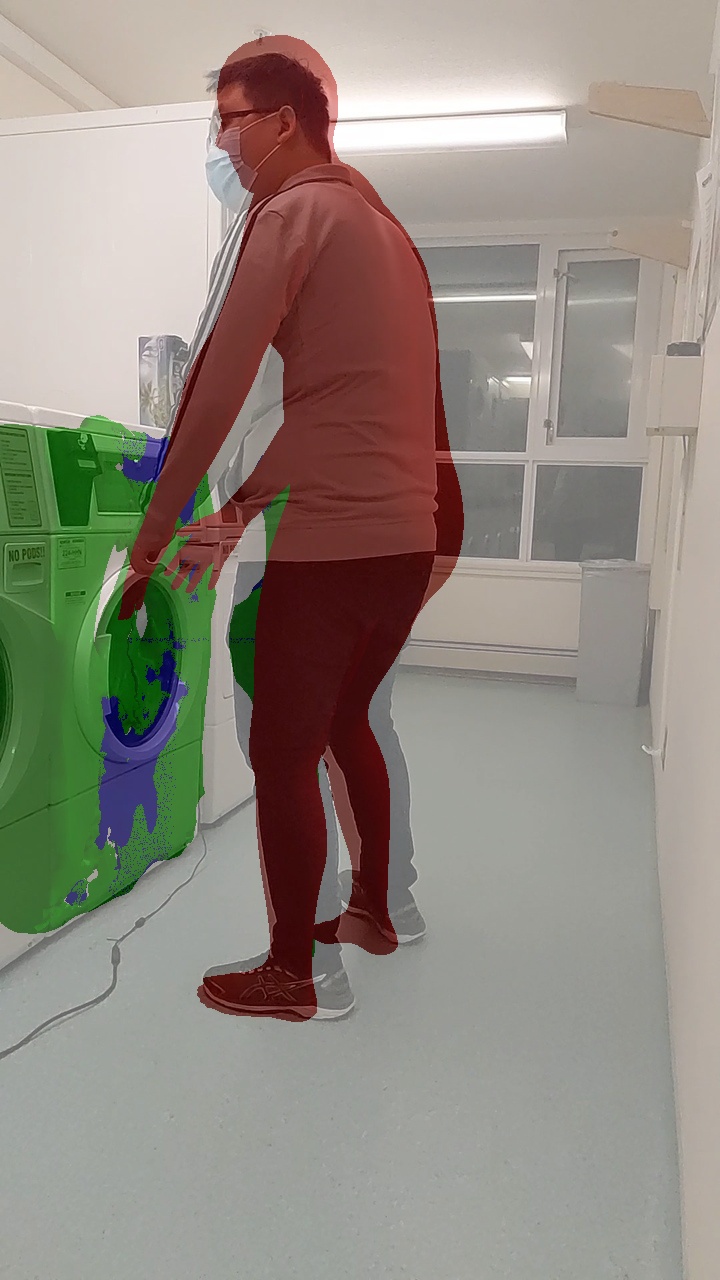} & 
\imgclip{0}{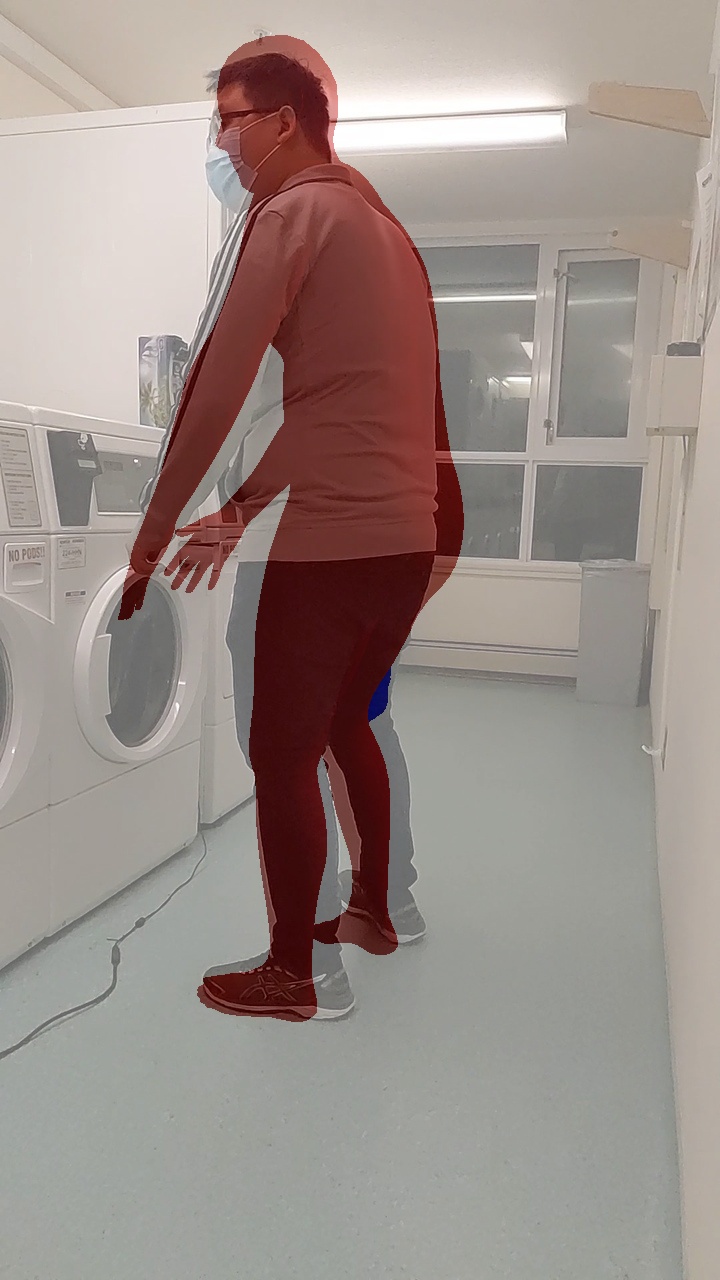}  \\

\end{tabularx}
\caption{
Additional qualitative comparisons.
The results show that reconstructing articulated 3D human-object interaction is very challenging.
All approaches exhibit significant errors in object shape reconstruction (in particular \internet which only handles the moving part of the object), and in motion parameter estimation (in particular \cuboidopt which gets the motion axis wrong in the second and third rows from the top).
Even \dhoi which has access to ground truth CAD models struggles to estimate the correct orientation in the first and last rows.
}
\label{fig:qualitative_supp}
\end{figure*}

\section{Details of method based on \ditto}
\label{sec:ditto_details}

There are a number of key differences in how we use \ditto~\cite{jiang2022ditto} compared to the original paper.
These differences may result in poorer performance when applied to our scenario. 
1) In the \ditto paper, the input point clouds are obtained by fusing multi-view depth images, but may still suffer from incompleteness due to self-occlusion (e.g., part of the object is not visible such as a drawer of a cabinet in the closed state).
Note that this is different from the incompleteness of the single-view point clouds with which we work.
2) The pretrained model is trained with the world coordinates matching the canonical object coordinate.
Thus, there is an implicit assumption that points are in canonical object coordinate.
Note that this differs from our use case, where we do not have the canonical object coordinates.
3) We apply the \ditto model on a full video sequence vs just two point clouds.
For some frames, the articulation state w.r.t. the reference closed state may be too similar, causing the network to perform badly.
4) Our videos contain a human interacting with the object, which can result in additional occlusions.

\section{Additional qualitative examples}
\label{sec:additional_qual}

\Cref{fig:qualitative_supp} shows further examples of the reconstruction quality achieved by the methods we benchmarked.
As we saw in the main paper, all approaches make significant errors in reconstruction and motion parameter estimation.
Due to the severely under-constrained nature of this problem, even methods with ground truth information struggle to reconstruct the objects reliably.
Specifically, \cuboidopt often makes error in motion axis and motion state parameter estimation.
\dhoi which has access to ground truth cad models also struggles to estimate the pose in many scenarios (trash bin in first row, and cabinet in second to last row).

\begin{table*}[!t]
\caption{\cuboidopt ablations of ground-truth vs predicted object and part masks, and ablations of human loss terms.
Values report mean error for the corresponding estimate.
}
\resizebox{\linewidth}{!}{
\begin{tabular}{@{} ll rr rrr rr rr @{}}
\toprule
& & \multicolumn{2}{c}{Reconstruction Error $\downarrow$} & \multicolumn{3}{c}{Pose Error $\downarrow$}
& \multicolumn{4}{c}{Motion Error $\downarrow$}\\
\cmidrule(l{0pt}r{2pt}){3-4} \cmidrule(l{0pt}r{2pt}){5-7}  \cmidrule(l{0pt}r{2pt}){8-11}  
Mask & Ablation & CD (Object) & CD (Moving) & Rotation & Translation & Scale &  Origin & Axis & Direction & State  \\
\midrule
\multirow{4}{*}{GT} & all losses & $1.83 \pm 0.15$ & $\best{0.67} \pm {0.08}$ & $44.19 \pm 2.09$ & $2.64 \pm 0.12$ &$\best{0.31} \pm 0.02$ & $\best{0.72} \pm 0.04$ & $24.76 \pm 2.38$ & $\textbf{95.86} \pm 5.52$ &$132.01 \pm 5.89$ \\
 & no $L_\text{depth}$ & $\best{1.06} \pm 0.07$ & $0.73 \pm 0.09$ & $44.22 \pm 2.39$ & $\textbf{2.58} \pm 0.10$ &$0.36 \pm 0.01$ & $0.95 \pm 0.05$ & $\best{19.44} \pm 2.14$ & $101.99 \pm 5.75$ &$130.95 \pm 6.64$ \\
& no $L_\text{contact}$ & $2.02 \pm 0.15$ & $0.68 \pm 0.07$ & $46.08 \pm 2.25$ & $2.69 \pm 0.13$ &$0.35 \pm 0.02$ & $0.77 \pm 0.04$ & $24.27 \pm 2.39$ & $101.32 \pm 5.51$ &$147.60 \pm 6.26$ \\ 
& no $L_\text{hoi}$ & $1.14 \pm 0.08$ & $0.85 \pm 0.10$ & $\best{43.07} \pm 2.29$ & $2.58 \pm 0.10$ &$0.38 \pm 0.02$ & $0.99 \pm 0.05$ & $25.95 \pm 2.42$ & $96.74 \pm 5.46$ &$\best{121.11} \pm 6.78$ \\

\midrule
\multirow{2}{*}{Pred}  & all losses  & $2.15 \pm 0.15$ & $\textbf{0.89} \pm 0.09$ & $\textbf{46.87} \pm 2.07$ & $2.67 \pm 0.11$ &$\textbf{0.29} \pm 0.02$ & $\textbf{0.69} \pm 0.04$ & $\textbf{27.46} \pm 2.39$ & $\textbf{91.30} \pm 5.36$ &$\textbf{126.88} \pm 5.56$ \\ 
& no $L_\text{hoi}$ & $\textbf{1.75} \pm 0.12$ & $1.09 \pm 0.10$ & $47.56 \pm 2.12$ & $\textbf{2.67} \pm 0.09$ &$0.39 \pm 0.02$ & $1.15 \pm 0.05$ & $40.10 \pm 2.72$ & $94.42 \pm 4.68$ &$129.95 \pm 5.34$ \\

\midrule
--- & \cuboidrand & $ 1.40 \pm 0.17 $ & $ 1.91 \pm 0.01 $ & $ 70.48 \pm 6.34 $ & $ 2.10 \pm 0.21 $ & $ 0.42 \pm 0.02 $ & $ 3.60 \pm 0.00 $ & $ 53.25 \pm 0.00 $ & $ 122.76 \pm 0.00 $ & $ 176.66 \pm 40.33 $ \\

\bottomrule
\end{tabular}
}
\label{tab:cubopt-err-ablation}
\end{table*}

\begin{table*}[!t]
\caption{\cuboidopt ablation for ground-truth vs predicted object and part masks, and ablations of human loss terms.
Values are accuracies computed at specified error thresholds.
The rotation threshold is 10 degree, the translation threshold is 0.5 and the scale threshold is 0.3.}
\resizebox{\linewidth}{!}{
\begin{tabular}{@{} ll rrr rrrr rrrr rrr @{}}
\toprule
& & \multicolumn{3}{c}{Reconstruction $\%$} & \multicolumn{4}{c}{Pose $\%$} & \multicolumn{4}{c}{Motion $\%$} & \multicolumn{3}{c}{Overall $\%$} \\
\cmidrule(l{0pt}r{2pt}){3-5} \cmidrule(l{0pt}r{2pt}){6-9} \cmidrule(l{0pt}r{2pt}){10-13} \cmidrule(l{0pt}r{2pt}){14-16} 
Mask & Ablation & Object@0.5 & Moving@0.5 & \accr & Rot@10 & Trans@0.5 & Scale@0.3 & \accp & O@0.5 & OA@10 & OAD@10 & {\accm}@10 & \accrp & RPOA & \accrpm \\
\midrule
\multirow{4}{*}{GT} & all losses & $31.8$ & $65.1$ & $26.4$ & $18.0$ & $2.1$ &$\textbf{56.5}$ & $1.4$ & $\textbf{42.9}$ & $\textbf{21.2}$ & $15.9$ &$8.6$ & $1.4$ & $1.0$ & $0.6$ \\
& no $L_\text{depth}$ & $33.2$ & $65.6$ & $26.9$ & $\textbf{20.8}$ & $3.4$ &$38.7$ & $1.3$ & $35.3$ & $19.4$ & $\textbf{17.6}$ &$\textbf{10.6}$ & $1.2$ & $1.2$ & $\textbf{1.2}$ \\
& no $L_\text{contact}$ & $32.4$ & $\textbf{66.4}$ & $\textbf{28.5}$ & $19.3$ & $\textbf{3.6}$ &$47.0$ & $\textbf{2.2}$ & $39.4$ & $18.2$ & $14.7$ &$8.7$ & $\textbf{2.2}$ & $\textbf{1.8}$ & $0.4$ \\ 
& no $L_\text{hoi}$ & $\textbf{35.8}$ & $62.6$ & $25.9$ & $18.1$ & $1.6$ &$36.6$ & $0.8$ & $27.6$ & $8.8$ & $5.9$ &$2.9$ & $0.6$ & $0.6$ & $0.4$ \\
\midrule
\multirow{2}{*}{Pred} & all losses & $21.1$ & $\textbf{58.4}$ & $\textbf{17.6}$ & $\textbf{14.6}$ & $\textbf{4.5}$ &$\textbf{58.3}$ & $\textbf{2.8}$ & $\textbf{41.8}$ & $\textbf{21.2}$ & $\textbf{15.3}$ &$\textbf{4.4}$ & $\textbf{2.4}$ & $\textbf{2.2}$ & $\textbf{0.8}$ \\ 
& no $L_\text{hoi}$ & $\textbf{21.8}$ & $51.5$ & $15.4$ & $12.3$ & $2.2$ &$38.3$ & $0.5$ & $12.9$ & $4.7$ & $1.8$ &$0.5$ & $0.3$ & $0.3$ & $0.1$ \\

\midrule
--- & \cuboidrand & $ 7.2 $ & $ 9.8 $ & $ 0.8 $ & $ 0.4 $ & $ 0.7 $ & $ 32.4 $ & $ 0.0 $ & $ 0.0 $ & $ 0.0 $ & $ 0.0 $ & $ 0.0 $ & $ 0.0 $ & $ 0.0 $ & $ 0.0 $ \\ 

\bottomrule
\end{tabular}
}
\label{tab:cubopt-acc-ablation}
\end{table*}

\begin{table*}[!t]
\caption{Comparison of \cuboidopt using predicted masks vs \internet. We use Mask2Former to obtain object masks, and \internet to obtain part masks.  Since the pretrained Mask2Former on COCO only overlaps in 4 categories (laptop, oven, microwave, refrigerator) with our objects, we evaluate on the subset of videos for these objects  (12 videos).  \cuboidopt-Pred outperforms \internet across all metrics.
}
\resizebox{\linewidth}{!}{
\begin{tabular}{@{} l rr rrr rr rr @{}}
\toprule
& \multicolumn{2}{c}{Reconstruction Error $\downarrow$} & \multicolumn{3}{c}{Pose Error $\downarrow$}
& \multicolumn{4}{c}{Motion Error $\downarrow$}\\
\cmidrule(l{0pt}r{2pt}){2-3} \cmidrule(l{0pt}r{2pt}){4-6}  \cmidrule(l{0pt}r{2pt}){7-10}  
Mask &  CD (Object) & CD (Moving) & Rotation & Translation & Scale &  Origin & Axis & Direction & State  \\
\midrule

\cuboidopt-Pred & $\textbf{2.153} \pm 0.148$ & $0.895 \pm 0.091$ & $\textbf{46.872} \pm 2.065$ & $2.670 \pm 0.112$ &$\textbf{0.294} \pm 0.015$ & $\textbf{0.693} \pm 0.037$ & $27.457 \pm 2.386$ & $\textbf{91.304} \pm 5.357$ &$126.884 \pm 5.561$ \\
\midrule
\internet & $4.041 \pm 0.109$ & $\textbf{0.316} \pm 0.038$ & $47.709 \pm 1.844$ & $\textbf{2.009} \pm 0.065$ &$0.512 \pm 0.015$ & $0.846 \pm 0.046$ & $\textbf{23.549} \pm 2.102$ & $111.416 \pm 5.443$ &$\textbf{71.509} \pm 19.692$ \\

\bottomrule
\end{tabular}
}
\label{tab:pred-err-fullset}
\end{table*}

\begin{table*}[!t]
\caption{Comparison of \cuboidopt using predicted masks vs \internet. We use Mask2Former to obtain object masks, and \internet to obtain part masks.  Since the pretrained Mask2Former on COCO only overlaps in 4 categories (laptop, oven, microwave, refrigerator) with our objects, we evaluate on the subset of videos for these objects  (12 videos).
\cuboidopt-Pred outperforms \internet across all metrics.
The rotation threshold is 10 degree, the translation threshold is 0.5, and the scale threshold is 0.3.}
\resizebox{\linewidth}{!}{
\begin{tabular}{@{} l rrr rrrr rrrr rrr @{}}
\toprule
& \multicolumn{3}{c}{Reconstruction $\%$} & \multicolumn{4}{c}{Pose $\%$} & \multicolumn{4}{c}{Motion $\%$} & \multicolumn{3}{c}{Overall $\%$} \\
\cmidrule(l{0pt}r{2pt}){2-4} \cmidrule(l{0pt}r{2pt}){5-8} \cmidrule(l{0pt}r{2pt}){9-12} \cmidrule(l{0pt}r{2pt}){13-15} 
Mask & Object@0.5 & Moving@0.5 & \accr & Rot@10 & Trans@0.5 & Scale@0.3 & \accp & O@0.5 & OA@10 & OAD@10 & {\accm}@10 & \accrp & RPOA & \accrpm \\
\midrule
\cuboidopt-Pred & $\textbf{21.1}$ & $\textbf{58.4}$ & $\textbf{17.6}$ & $\textbf{14.6}$ & $\textbf{4.5}$ &$\textbf{58.3}$ & $\textbf{2.8}$ & $\textbf{41.8}$ & $\textbf{21.2}$ & $\textbf{15.3}$ &$\textbf{4.4}$ & $\textbf{2.4}$ & $\textbf{2.2}$ & $\textbf{0.8}$ \\ 
\midrule
\internet  & $1.3$ & $45.0$ & $1.3$ & $3.1$ & $0.9$ &$9.0$ & $0.0$ & $17.5$ & $9.7$ & $5.8$ &$0.0$ & $0.0$ & $0.0$ & $0.0$ \\
\bottomrule
\end{tabular}
}
\label{tab:pred-acc-fullset}
\end{table*}

\section{Additional quantitative evaluation}
\label{sec:additional_quant}

\mypara{\cuboidopt ablations.}
We experiment with different versions of the \cuboidopt baseline (see \Cref{tab:cubopt-err-ablation} and \Cref{tab:cubopt-acc-ablation}).
Here, we report numbers for different configurations of the losses used in \cuboidopt.
We start with all the losses and then ablate the components of $L_\text{hoi}$.
We note that $L_\text{contact}$ helps achieve much better reconstruction and pose results and $L_\text{depth}$ leads to better performance on Scale and Origin estimates.

We also investigate the use of ground-truth (GT) and predicted (Pred) object and part masks.
For predicted masks, we use Mask2Former~\cite{cheng2021maskformer} (Swin Large, trained on COCO) to predict the object masks.
Since the pretrained Mask2Former on COCO does not include all the categories for the D3D-HOI dataset we used, we only evaluate on 4 categories (laptop, microwave, oven, refrigerator).
For the moving part masks, we use the \internet mask estimates without any temporal optimization.
\Cref{tab:cubopt-err-ablation} shows that \cuboidopt performs much worse on predicted masks than ground truth masks.
It performs worse for reconstruction and pose error, however it performs close to the ground truth masks baseline on the motion estimates.
We also note that \cuboidopt with predicted masks performs similar to random initialization on some metrics showing that there is much room for improvement.

\mypara{\cuboidopt comparison with \internet.}
\internet~\cite{qian2022understanding} works by first detecting moving part planes in 2D images and then performing temporal optimization to predict motion parameters including axis and state.
We note that the temporal optimization step in \internet is analogous to \cuboidopt optimization.
Therefore, we compare the results of \internet and \cuboidopt with predicted moving part masks. The results are shown in \Cref{tab:pred-err-fullset} and \Cref{tab:pred-acc-fullset}.
We note that \cuboidopt generally outperforms \internet including across almost all metrics in \Cref{tab:pred-acc-fullset}.
This shows that an optimization approach with a simple inductive bias of cuboidal abstraction for articulated parts is quite effective, and can compete and outperform a learning-based approach that required large volumes of supervised data.

\end{document}